\documentclass[runningheads]{llncs}
\usepackage{graphicx}
\usepackage{subfigure}
\usepackage{cite}
\usepackage{wrapfig,lipsum,booktabs}

\usepackage{comment}
\usepackage{amsmath,amssymb} % define this before the line numbering.
\usepackage{color}
\usepackage{booktabs}
\usepackage{amsmath}
\usepackage{bbm}
\usepackage{multirow}
\usepackage{tabularx}
\usepackage{makecell}
\usepackage{xcolor,pifont}
\usepackage{pgfplots}
\usepackage{array}

\usepackage{algorithmic}
\usepackage{color}
\pgfplotsset{compat=newest}

\usepackage[toc,page]{appendix}

\definecolor{tb_color_1}{RGB}{245,124,0}
\definecolor{tb_color_2}{RGB}{0,167,247}

\usepackage{array}

\newcommand{\xmark}{\ding{55}}%
\newcommand{\cmark}{\ding{51}}%

\definecolor{ForestGreen}{RGB}{34,139,34}
\newcommand{\markgood}[1]{{\color{ForestGreen}#1}}

\usepackage[accsupp]{axessibility}  % Improves PDF readability for those with disabilities.

\usepackage{hyperref}
\hypersetup{
    colorlinks = true,
    linkbordercolor = {white}
}

\makeatletter
\def\hlinewd#1{%
\noalign{\ifnum0=`}\fi\hrule \@height #1 \futurelet
\reserved@a\@xhline}
\usepackage[ruled,vlined]{algorithm2e}
\definecolor{commentcolor}{rgb}{0.294,0.604,0.608}   % define comment color
\definecolor{pccolor}{rgb}{0.858, 0.188, 0.478}

\usepackage{pifont}% http://ctan.org/pkg/pifont

\begin{document}

\pagestyle{headings}
\mainmatter
\def\ECCVSubNumber{3758}  % Insert your submission number here

\title{ConMatch: Semi-Supervised Learning with Confidence-Guided Consistency Regularization} % Replace with your title

\begin{comment}/
\titlerunning{ECCV-22 submission ID \ECCVSubNumber} 
\authorrunning{ECCV-22 submission ID \ECCVSubNumber} 
\author{Anonymous ECCV submission}
\institute{Paper ID \ECCVSubNumber}
\end{comment}
%******************
\renewcommand*{\@fnsymbol}[1]{\ensuremath{\ifcase#1\or \dagger\or *\or \ddagger\or
   \mathsection\or \mathparagraph\or \|\or **\or \dagger\dagger
   \or \ddagger\ddagger \else\@ctrerr\fi}}
\newcommand*\samethanks[1][\value{footnote}]{\footnotemark[#1]}

% \begin{comment}
\titlerunning{ConMatch}
\author{Jiwon Kim \inst{1,2}\thanks{Work done in Korea University}\thanks{Equal contribution} \and
Youngjo Min\inst{1}\samethanks\quad \and
Daehwan Kim \inst{3}\samethanks\quad \and
Gyuseong Lee\inst{1} \and\\
Junyoung Seo\inst{1} \and
Kwangrok Ryoo\inst{1} \and
Seungryong Kim \inst{1}}
\authorrunning{J. Kim et al.}
\institute{Korea University, Seoul, Korea \\
\email{\{naancoco@korea.ac.kr, 1320harry@korea.ac.kr, jpl358@korea.ac.kr, se780@korea.ac.kr, kwangrok21@korea.ac.kr, seungryong\_kim@korea.ac.kr\}} \and
NAVER AI Lab, Seongnam, Korea \\ \and
Samsung Electro-Mechanics, Suwon, Korea\\
\email{daehwan85.kim@samsung.com}} 

\maketitle

\begin{abstract}
We present a novel semi-supervised learning framework that intelligently leverages the consistency regularization between the model's predictions from two strongly-augmented views of an image, weighted by a confidence of pseudo-label, dubbed ConMatch. While the latest semi-supervised learning methods use weakly- and strongly-augmented views of an image to define a directional consistency loss, how to define such direction for the consistency regularization between two strongly-augmented views remains unexplored. To account for this, we present novel confidence measures for pseudo-labels from strongly-augmented views by means of weakly-augmented view as an anchor in non-parametric and parametric approaches. Especially, in parametric approach, we present, for the first time, to learn the confidence of pseudo-label within the networks, which is learned with backbone model in an end-to-end manner. In addition, we also present a stage-wise training to boost the convergence of training. When incorporated in existing semi-supervised learners, ConMatch consistently boosts the performance. We conduct experiments to demonstrate the effectiveness of our ConMatch over the latest methods and provide extensive ablation studies. Code  has been made publicly available at ~\url{https://github.com/JiwonCocoder/ConMatch}

\begin{comment}
\sr{(WILL REWRITE!!!)
This paper presents a novel semi-supervised learning framework, called ConMatch, that capably boosts the performance of various semi-supervised learning (SSL) methods as plug in play way. The latest existing SSL methods are applying the constant fixed high threshold for all samples to select samples to be used for training, thus the method results in slow convergence, erasing too many correct pseudo-labeled samples and using too many incorrect pseudo-labeled samples. To address these this problem, we propose ConMatch that gives guidance of supervision through the measured confidence of the pseudo label. Confidence-guided self-supervised learning framework expedite optimization by using numbers of unlabeled data from the beginning of training unlike existing methods, through using a weak-augmentation branch as an anchor and estimating the confidence of each pseudo label from the strong branch to exploit a self-supervision between two strong-augmentation branches. Confidence can be obtained either parametrically or non-parametrically, and both methods surpass the previous state-of-the-arts in various benchmarks, demonstrating the potential for improvement that confidence measurement in SSL can lead. With the proposed contributions, both FixMatch and FlexMatch yields significant performance gain over the original model and produces leading results on the various CIFAR10, CIFAR100, SVHN benchmark for semi-supervised learning.}
\end{comment}

\end{abstract}

\section{Introduction}
Semi-supervised learning has emerged as an attractive solution to mitigate the reliance on large labeled data, which is often laborious to obtain, and intelligently leverage a large amount of unlabeled data, to the point of being
deployed in many computer vision applications, especially image classification~\cite{yalniz2019billion,xie2020self,pham2021meta}. Generally, this task have adopted pseudo-labeling~\cite{grandvalet2004semi,lee2013pseudo, shi2018transductive,xie2019unsupervised, arazo2020pseudo, zoph2020rethinking, pham2021meta} or consistency regularization~\cite{laine2016temporal, tarvainen2017mean, french2017self, miyato2018virtual,ke2019dual, xie2020self}. Some methods~\cite{berthelot2019mixmatch, berthelot2019remixmatch, xie2020unsupervised, sohn2020fixmatch, zhang2021flexmatch, xu2021dash, rizve2021defense} proposed to integrate both approaches in a unified framework, which is often called holistic approach. 
As one of pioneering works, FixMatch~\cite{sohn2020fixmatch} first generates a pseudo-label from the model's prediction on the weakly-augmented instance and then encourages the prediction from the strongly-augmented instance to follow the pseudo-label. Their success inspired many variants that use, e.g., curriculum learning~\cite{zhang2021flexmatch,xu2021dash}.

On the other hand, concurrent to the race for better semi-supervised learning methods~\cite{sohn2020fixmatch,zhang2021flexmatch,xu2021dash}, substantial progress has been
made in self-supervised representation learning, especially with contrastive learning~\cite{bachman2019learning, chen2020simple, he2020momentum, grill2020bootstrap, caron2020unsupervised,chen2021exploring}, aiming at learning a task-agnostic feature representation without any supervision, which can be well transferred to the downstream tasks. Formally, they encourage the features extracted from two differently-augmented images to be pulled against each other, which injects some invariance or robustness into the models. Not surprisingly, semi-supervised learning frameworks can definitely benefit from self-supervised representation learning~\cite{li2021comatch, kim2021selfmatch, lucas2021barely} in that good representation from the feature encoder yields better performance with semi-supervised learning, and thus, some methods~\cite{li2021comatch,kim2021selfmatch} attempt to combine the aforementioned two paradigms to boost the performance by achieving the better feature encoder. 
% For instance, SelfMatch~\cite{kim2021selfmatch} first trained the model with self-supervised pre-training and fine-tuned it with semi-supervised learning, and CoMatch~\cite{li2021comatch} directly combined the two paradigms in a unified framework. 

Extending techniques presented in existing self-supervised representation learning~\cite{bachman2019learning, chen2020simple, he2020momentum, grill2020bootstrap, caron2020unsupervised,chen2021exploring}, which only focus on learning feature encoder, to further consider the model's prediction itself would be an appealing solution to effectively combine the two paradigms, which allows for boosting not only feature encoder but also classifier. However, compared to feature representation learning~\cite{bachman2019learning, chen2020simple, he2020momentum, grill2020bootstrap, caron2020unsupervised,chen2021exploring}, the consistency between the model's predictions from two different augmentations should be defined by considering which direction is better to achieve not only invariance but also high accuracy in image classification. Without this, simply pulling the model's predictions as done in~\cite{bachman2019learning, chen2020simple, he2020momentum, grill2020bootstrap, caron2020unsupervised,chen2021exploring} may hinder the classifier output, thereby decreasing the accuracy. 

In this paper, we present a novel framework for semi-supervised learning, dubbed ConMatch, that intelligently leverages the confidence-guided consistency regularization between the model's predictions from two strongly-augmented images. Built upon conventional frameworks~\cite{sohn2020fixmatch,zhang2021flexmatch}, we consider two strongly-augmented images and one weakly-augmented image, and define the consistency between the model's predictions from two strongly-augmented images, while still using an unsupervised loss between the model's predictions from one of the strongly-augmented images and the weakly-augmented image, as done in~\cite{sohn2020fixmatch,zhang2021flexmatch}. Since defining the direction of consistency regularization between two strongly-augmented images is of prime importance, rather than selecting in a deterministic manner, we present a probabilistic technique by measuring the confidence of pseudo-labels from each strongly-augmented image, and weighting the consistency loss with this confidence. 
To measure the confidence of pseudo-labels, we present two techniques, including non-parametric and parametric approaches. With this confidence-guided consistency regularization, our framework dramatically boosts the performance of existing semi-supervised learners~\cite{sohn2020fixmatch,zhang2021flexmatch}. In addition, we also present a stage-wise training scheme to boost the convergence of training. 
Our framework is a plug-and-play module, and thus various semi-supervised learners~\cite{berthelot2019remixmatch, xie2020unsupervised, sohn2020fixmatch,zhang2021flexmatch, xu2021dash, li2021comatch, kim2021selfmatch, lucas2021barely} can benefit from our framework. We briefly summarize our method with other highly relevant works in semi-supervised learning in Table~\ref{table:method-comparison}. Experimental results and ablation studies show that the proposed framework not only boosts the convergence but also achieves the state-of-the-art performance on most standard benchmarks~\cite{krizhevsky2009learning, netzer2011reading,coates2011analysis}. 

\begin{table}[t]
\caption{
\textbf{Comparison of our ConMatch to other relevant works which have a form of consistency regularization combining pseudo-labeling~\cite{berthelot2019mixmatch, sohn2020fixmatch, zhang2021flexmatch, xu2021dash, kim2021selfmatch, li2021comatch, lucas2021barely}}
}
%$\textbf{{F}}$ means a feature vector, $\textbf{{L}}$ means an output probability, and $\textbf{{C}}$ is a learned confidence from the network.
\label{table:method-comparison}\vspace{-25pt}
\begin{center}
\resizebox{\textwidth}{!}{
\begin{tabular}{l|ccccccc|c}
\toprule
% \makecell{\\Algorithm} & \makecell{Pseudo\\Labeling}  &\makecell{Self-Sup.\\Loss}  & \makecell{Consistency\\S-S} & \makecell{Learned\\Confidence} & \makecell{Confidence\\Guidance} \\
& MixMatch  &FixMatch & FlexMatch& Dash &SelfMatch &CoMatch & LESS & ConMatch\\
&~\cite{berthelot2019mixmatch} & ~\cite{sohn2020fixmatch} & ~\cite{zhang2021flexmatch} & ~\cite{xu2021dash} & ~\cite{kim2021selfmatch} & ~\cite{li2021comatch} &~\cite{lucas2021barely} & (Ours) \\
\midrule
Using pseudo-labeling    &  \xmark    & \markgood{\cmark} & \markgood{\cmark}  & \markgood{\cmark} & \markgood{\cmark} & \markgood{\cmark} & \markgood{\cmark} &\markgood{\cmark} \\
Using two strong branches      &  \xmark    & \xmark & \xmark  & \xmark & \markgood{\cmark} & \markgood{\cmark} & \xmark & \markgood{\cmark} \\
%Self-sup.target    &  \xmark    & \xmark & \xmark  & \xmark & \textbf{{F}} & \textbf{{F}} & \textbf{{F}}& \textcolor{red}{\textbf{{L}}} \\
Learning confidence measure & \xmark & \xmark & \xmark &\xmark & \xmark &\xmark &\xmark& \markgood{\cmark}\\
Using stage-wise training & \xmark & \xmark & \xmark &\xmark & \markgood{\cmark} &\xmark &\xmark& \markgood{\cmark}\\
%Learned confidence &  \xmark    & \xmark & \xmark  & \xmark & \xmark & \xmark & \xmark & \redcheck \\
%Confidence-guided loss  &  \xmark    & \xmark & \xmark  & \xmark & \xmark & \xmark & \xmark & \redcheck \\
\bottomrule
\end{tabular}
}
\end{center}
\vspace{-10pt}
\end{table}

\section{Related Works}

\subsubsection{Semi-supervised Learning.}
Semi-supervised learning has been an effective paradigm for leveraging an abundance of unlabeled data along with limited labeled data. For this task, various methods such as pseudo-labeling~\cite{lee2013pseudo,grandvalet2004semi} and consistency regularization~\cite{sajjadi2016regularization, laine2016temporal, tarvainen2017mean} have been proposed. In pseudo-labeling~\cite{lee2013pseudo}, a model uses unlabeled samples with high confidence as training targets, which reduces the density of data points at the decision boundary~\cite{grandvalet2004semi, sajjadi2016mutual}. Consistency regularization has been first introduced by $\pi$-model~\cite{sajjadi2016regularization}, which is further improved by numerous following works~\cite{laine2016temporal, tarvainen2017mean, french2017self, miyato2018virtual,ke2019dual, xie2020self}. In the consistency regularization, the model should minimize the distance between the model's predictions when fed perturbed versions of the input~\cite{laine2016temporal, tarvainen2017mean,miyato2018virtual,park2018adversarial,ke2019dual,xie2019unsupervised,xie2020self} or the model~\cite{laine2016temporal, tarvainen2017mean, park2018adversarial, ke2019dual, xie2020self,zhang2020wcp}. 
Very recently, advanced consistency regularization methods~\cite{berthelot2019remixmatch, xie2020unsupervised, sohn2020fixmatch} have been introduced by combining with pseudo-labeling. These methods show high accuracy, comparable to supervised learning in a fully-labeled setting, e.g., ICT~\cite{verma2019interpolation}, MixMatch~\cite{berthelot2019mixmatch}, UDA~\cite{xie2020unsupervised}, ReMixMatch~\cite{berthelot2019remixmatch}, and FixMatch~\cite{sohn2020fixmatch}. 
The aforementioned methods can be highly boosted by simultaneously considering the techniques proposed in recent self-supervised representation learning methods~\cite{bachman2019learning, chen2020simple, he2020momentum, grill2020bootstrap, caron2020unsupervised}.
\vspace{-10pt}

\subsubsection{Self-supervised Representation Learning.}
Self-supervised representation learning has recently attracted much attention~\cite{donahue2014decaf, zhang2016colorful, noroozi2016unsupervised, gidaris2018unsupervised, bachman2019learning, chen2020simple, he2020momentum, grill2020bootstrap, caron2020unsupervised} due to its competitive performance. Specifically, contrastive learning~\cite{bachman2019learning, chen2020simple, he2020momentum, grill2020bootstrap, caron2020unsupervised} becomes a dominant framework. It formally maximizes the agreement between different augmented views of the same image~\cite{donahue2014decaf, zhang2016colorful, noroozi2016unsupervised, gidaris2018unsupervised}. 
%Contrastive learning can be simply instantiated through Siamese networks~\cite{bromley1993signature}, which is effective for comparing entities by sharing weight-sharing neural networks applied on two or more inputs but vulnerable to model collapse. 
Most previous methods benefit from a large amount of negative pairs to preclude constant outputs and avoid a collapse problem~\cite{chen2020simple}. An alternative to approximate the loss is to use cluster-based approach by discriminating between groups of images with similar features~\cite{caron2020unsupervised}. 
Some methods~\cite{grill2020bootstrap, chen2021exploring} mitigated to use negative samples by using a momentum encoder~\cite{grill2020bootstrap} and a stop-gradient technique~\cite{chen2021exploring}. The aforementioned methods applied the consistency loss at the feature-level, unlike recent semi-supervised learning methods~\cite{sohn2020fixmatch,zhang2021flexmatch} that consider the consistency loss in the logit-level, which may not be optimal to be incorporated with semi-supervised learners. Formulating the consistency loss in the logit-level as self-supervision is challenging because a direction between two augmented views should be determined. Without this, simply pulling the model's predictions as done in~\cite{bachman2019learning, chen2020simple, he2020momentum, grill2020bootstrap, caron2020unsupervised,chen2021exploring} may hinder the classifier output, thereby decreasing the accuracy. 
\vspace{-10pt}

\begin{figure}[t!] 
\centering
    % \renewcommand{\thesubfigure}{}
% 	\subfigure[ \makecell{(a) Semi-Sup.\\(FixMatch)}]
	\subfigure[ Semi-sup.]{\includegraphics[width=0.33\linewidth]{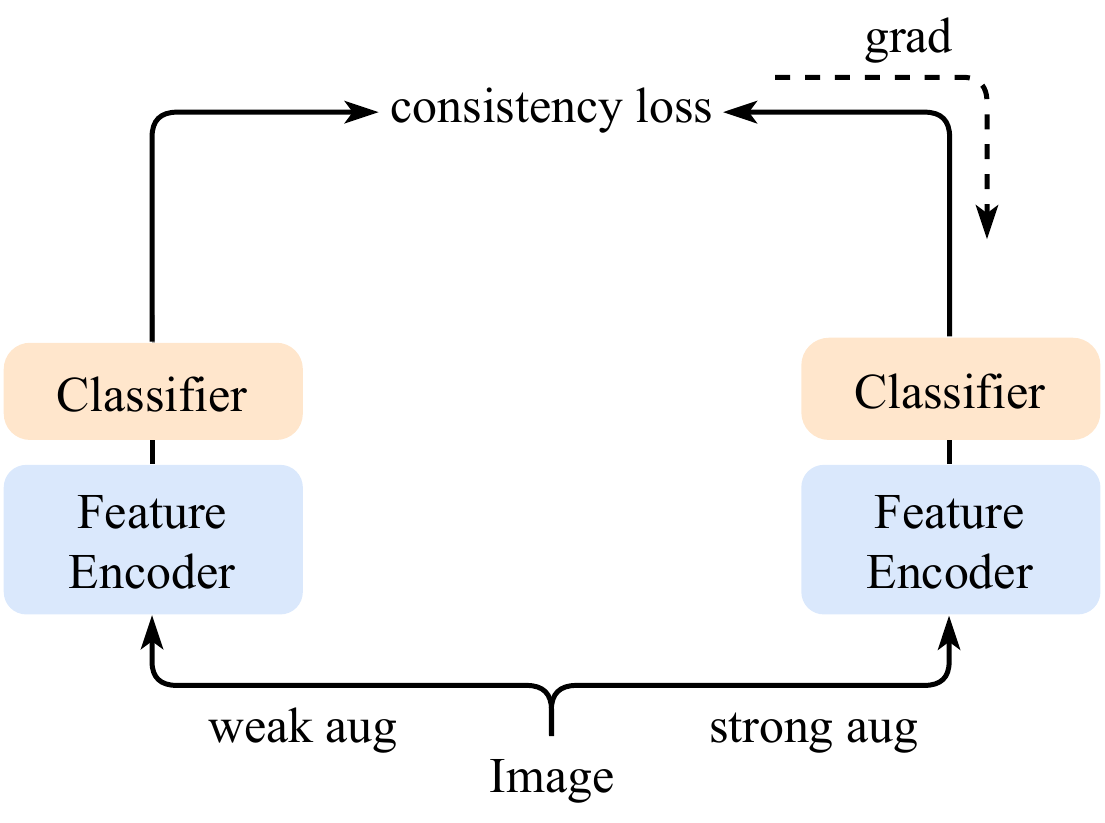}}\hfill
    % \subfigure[\makecell{(b) Self-Sup.\\(Simsiam)}]
    \subfigure[Self-sup.]
	{\includegraphics[width=0.33\linewidth]{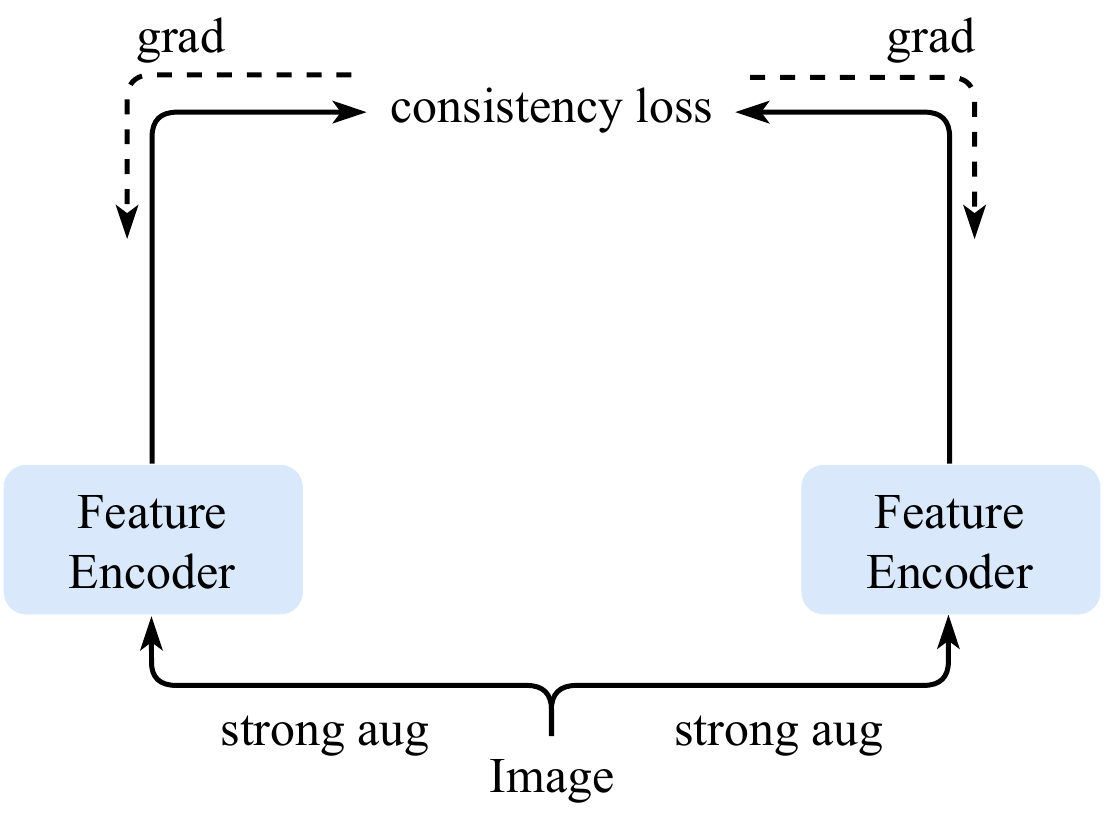}}\hfill
% 	\subfigure[\makecell{(c) Semi with Self-sup.\\(CoMatch)}]
	\subfigure[ Semi-sup. with self-sup.]
	{\includegraphics[width=0.33\linewidth]{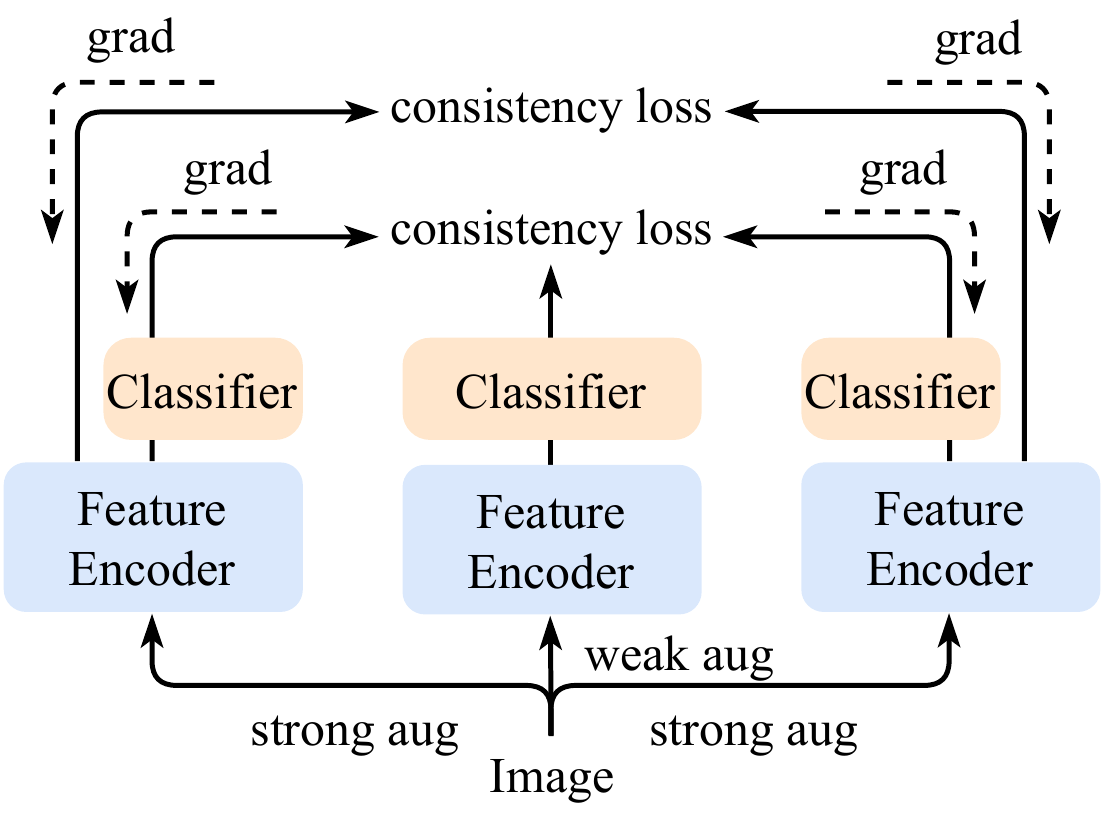}}\hfill\\
	\vspace{-10pt}
	\caption{\textbf{Conceptual illustration of existing methods that leverage unlabeled data:} (a) semi-supervised learning- the model uses the model's prediction itself to produce a pseudo-label for unlabeled data~\cite{lee2013pseudo, sajjadi2016regularization, laine2016temporal, tarvainen2017mean , french2017self, miyato2018virtual, verma2019interpolation, berthelot2019mixmatch, berthelot2019remixmatch, xie2020unsupervised, sohn2020fixmatch}, (b) self-supervised representation learning- the model is learned to generate the same feature embedding for two augmented views from unlabeled data~\cite{bachman2019learning, chen2020simple, he2020momentum, grill2020bootstrap, caron2020unsupervised, grill2020bootstrap, chen2021exploring}, and (c) semi-supervised learning with self-supervision representation learning- the model simultaneously learns a feature representation with self-supervised representation loss, while learning all the networks with semi-supervised learning~\cite{kim2021selfmatch, li2021comatch, lucas2021barely}.}\label{fig:intuition}
	\vspace{-5pt}    
\end{figure}

\subsubsection{Self-supervision in Semi-supervised Learning.}
Many recent state-of-the-art semi-supervised learning methods adopt the self-supervised representation learning methods~\cite{zhai2019s4l,chen2020big} to jointly learn good feature representation. Self-supervised pre-training, followed by supervised fine-tuning, has shown strong performance on semi-supervised learning settings. Specifically, SelfMatch~\cite{kim2021selfmatch} adopted SimCLR~\cite{chen2020simple} for self-supervised pre-training and FixMatch~\cite{sohn2020fixmatch} for semi-supervised fine-tuning. However, it may learn sub-optimal representation for the image classification task due to the task-agnostic learning. On the other hand, some methods~\cite{li2021comatch, lee2022contrastive} unify pseudo-labeling and self-supervised learning. \cite{lerner2020boosting} alternates between self- and semi-supervised learning. There lacks a study to effectively use self-supervision, rather than simply adopting this. \vspace{-10pt}

\subsubsection{Confidence Estimation in Semi-supervised Learning.}
In semi-supervised learning, a confidence-based strategy has been widely used along with pseudo labeling so that the unlabeled data are used only when the predictions are sufficiently confident. Such confidence in pseudo-labeling has been often measured by the peak values of the predicted probability distribution~\cite{xie2020unsupervised, sohn2020fixmatch, zhang2021flexmatch, xu2021dash, rizve2021defense}. Although the selection of unlabeled samples with high confidence predictions moves decision boundaries to low density regions~\cite{chapelle2005semi}, many of these selected predictions are incorrect due to the poor calibration of neural networks~\cite{guo2017calibration}, which has the discrepancy between the confidence level of a network's individual predictions and its overall accuracy and leads to noisy training and poor generalization~\cite{dawid1982well, degroot1983comparison}. However, there was no study how to learn the confidence of pseudo-labels, which is the topic of this paper. 

%On the other hands, in stereo confidence estimation, many recent methods leverage CNNs to infer confidence maps~\cite{poggi2016learning, seki2016patch, kim2017deep, poggi2017learning, poggi2017quantitative, tosi2018beyond, kim2018unified, kim2020adversarial}. We have adopted the confidence estimation step that detects unreliable initial disparities and reduces the effect of poor network calibration. 

\section{Methodology}
\subsection{Preliminaries}
Let us define a batch of \textit{labeled} instances as $
\mathcal{X}=\left\{\left(x_{b}, y_{b}\right)\right\}_{b=1}^{B}$, where $x_b$ is an instance and $y_b$ is a label representing one of $Y$ labels. In addition, let us define a batch of \textit{unlabeled} instances as $\mathcal{U}=\left\{u_{b}\right\}_{b=1}^{\mu B}$, where $\mu$ is a hyper-parameter that determines the size of $\mathcal{U}$ relative to $\mathcal{X}$. The objective of semi-supervised learning is to use both $\mathcal{X}$ and $\mathcal{U}$ to train a model with parameters $\theta$ taking an instance $r \in \mathcal{X}\cup\mathcal{U}$ as input and outputting a distribution over class labels $y$ such that $p_\mathrm{model}(y|r;\theta)$. The model generally consists of an feature encoder $f(\cdot)$ and a classifier $g(\cdot)$, and thus, $p_\mathrm{model}(y|r;\theta) = g(f(r))$.

For semi-supervised learning, most state-of-the-art methods are based on consistency regularization approaches~\cite{bachman2014learning,sajjadi2016regularization, laine2016temporal} that rely on the assumption that the model should generate similar predictions when perturbed versions of the same instance are fed, e.g., using data augmentation~\cite{miyato2018virtual}, or model perturbation~\cite{laine2016temporal, tarvainen2017mean}. These methods formally extract a pseudo-label from one branch, filtered by confidence, and use this as a target for another branch. For instance, FixMatch~\cite{sohn2020fixmatch} utilizes two types of augmentations such as \textit{weak} and \textit{strong}, denoted by $\alpha(\cdot)$ and $\mathcal{A}(\cdot)$, and a pseudo-label from weakly-augmented version of an image is used as a target for strongly-augmented version of the same image. This loss function is formally defined such that
\begin{equation}
    \mathcal{L}_\mathrm{un}=c(r) \mathcal{H}\left(q(r), p_{\mathrm{model}}\left(y| \mathcal{A}(r);\theta\right)\right),
    \label{eqn:fixmatch-loss}
\end{equation}
where $c(r)$ denotes a confidence of $q(r)$, and $q(r)$ denotes a pseudo-label generated from $p_{\mathrm{model}}(y| \alpha(r);\theta)$, which can be either an one-hot label~\cite{sohn2020fixmatch,zhang2021flexmatch, rizve2021defense, xu2021dash} or a sharpened one~\cite{xie2020unsupervised, berthelot2019mixmatch, berthelot2019remixmatch}, and $\mathcal{H}(\cdot,\cdot)$ is often defined as a cross-entropy loss. In this framework, measuring confidence $c(r)$ is of prime importance, but conventional methods simply measure this, e.g., by the peak value of the softmax predictions~\cite{xie2020unsupervised, sohn2020fixmatch, zhang2021flexmatch,rizve2021defense,xu2021dash}.

On the other hands, semi-supervised learning framework can definitely benefit from existing self-supervised representation learning~\cite{li2021comatch, kim2021selfmatch, lucas2021barely} in that good representation from the feature encoder $f(\cdot)$ yields better performance with semi-supervised learner. In this light, some methods attempted to combine semi-supervised learning and self-supervised representation learning to achieve the better feature encoder~\cite{li2021comatch,kim2021selfmatch}. 
Concurrent to the race for better semi-supervised learning methods, substantial progress has been
made in self-supervised representation learning, especially with contrastive learning~\cite{bachman2019learning, chen2020simple, he2020momentum, grill2020bootstrap, caron2020unsupervised,chen2021exploring}. The loss function for this task can also be defined as a consistency regularization loss, similar to~\cite{xie2020unsupervised, sohn2020fixmatch, zhang2021flexmatch,rizve2021defense,xu2021dash} but in the feature-level, such that 
\begin{equation}
    \begin{split}
      \mathcal{L}_\mathrm{self} = \mathcal{D}(F_i(r),F_j(r)),
    \end{split}
    \label{eqn:fixmatch-loss}
\end{equation}
where $F_i(r) = f(\mathcal{A}_i(r))$ and $F_j(r) = f(\mathcal{A}_j(r))$ extracted from images with two different strongly-augmented images $\mathcal{A}_i(\cdot)$ and $\mathcal{A}_j(\cdot)$, respectively. $\mathcal{D}(\cdot,\cdot)$ can be defined as contrastive loss~\cite{he2020momentum} or negative cosine similarity~\cite{chen2021exploring}. %Compared to contrastive learning~\cite{bachman2019learning, chen2020simple, he2020momentum}, this loss does not require other samples for negative terms.
Even though this loss helps to boost learning the feature encoder $f(\cdot)$, the mechanism that simply pulls the features $F_{i}(r)$ and $F_{j}(r)$ may not be optimal to boost a semi-supervised learner and break the latent feature space, without considering a direction representing which branch is better. 

\begin{figure}[t]
\centering
\includegraphics[width=0.8\linewidth]{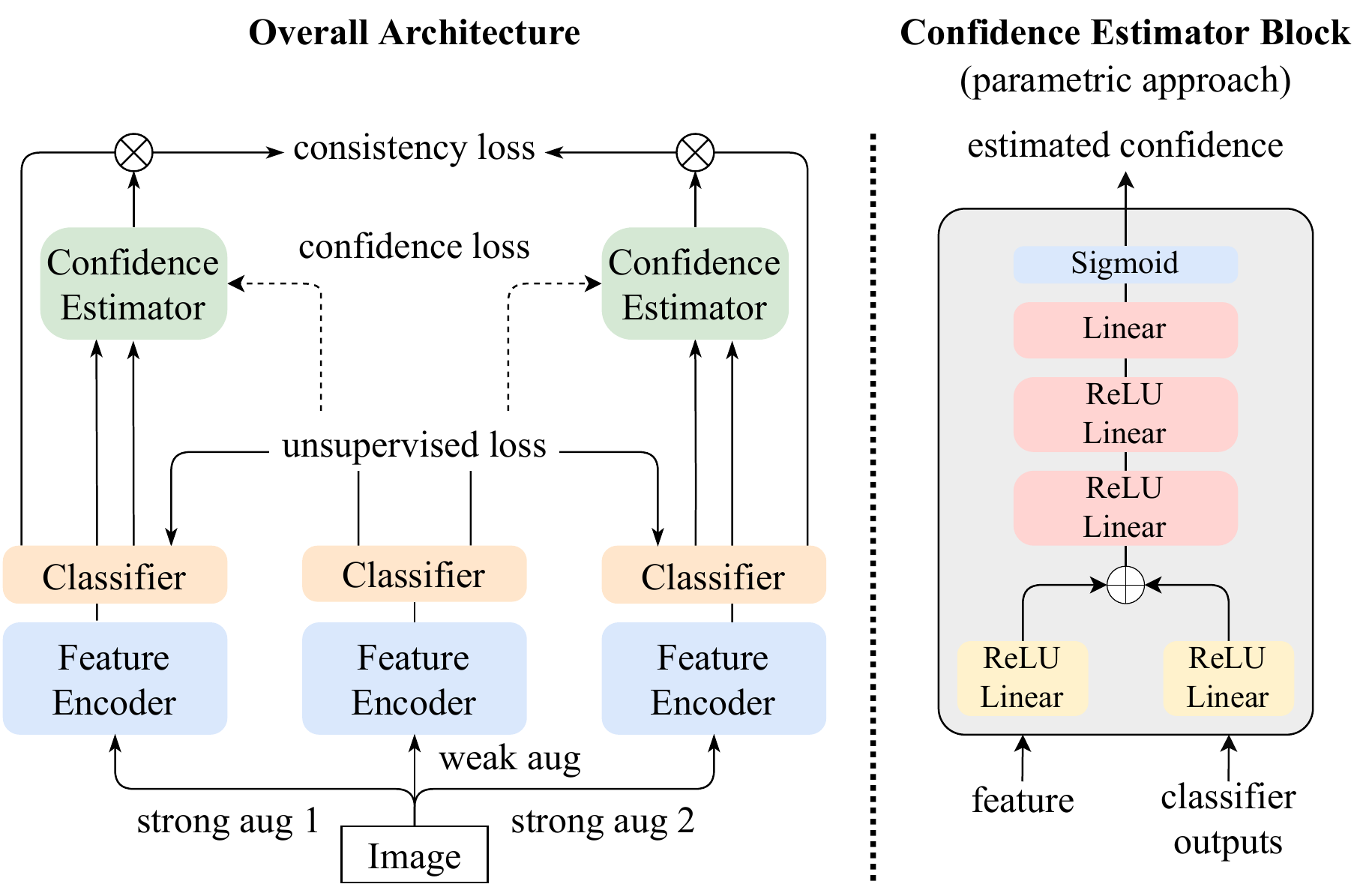}\\
\vspace{-5pt}
\caption{\textbf{Network configuration of ConMatch.} A semi-supervised learning framework built upon consistency loss with an additional strong branch to leverage confidence loss between two strong branches. In the parametric approach, the confidence estimator block takes a concatenated heterogeneous feature as an input and produces the estimated confidence of pseudo label.}
\label{fig:main-network}\vspace{-5pt}
\end{figure}

\subsection{Formulation}
To combine the semi- and self-supervised learning paradigm in a boosting fashion, unlike~\cite{li2021comatch, kim2021selfmatch, lucas2021barely}, we present to effectively exploit a self-supervision between two \textit{strong} branches tailored for boosting semi-supervised learning, called ConMatch. Unlike existing self-supervised representation learning methods, e.g., SimSiam~\cite{chen2021exploring}, we formulate the consistency regularization loss at class logit-level\footnote{In the paper, class logit means the output of the network, i.e., $p_{\mathrm{model}}(y|r;\theta)$ for $r$.}, as done in semi-supervised learning methods~\cite{sohn2020fixmatch,zhang2021flexmatch}, and estimate the confidences of each pseudo-label from two \textit{strongly}-augmented images, $\mathcal{A}_i(r)$ and $\mathcal{A}_j(r)$ for $r$, and use them to consider the probability of each direction between them. Since measuring such confidences is notoriously challenging, we present novel confidence estimators by using the output from \textit{weak}-augmented image $\alpha(r)$ as an anchor in non-parametric and parametric approaches.

An overview of our ConMatch is illustrated in Fig.~\ref{fig:main-network}.  Specifically, there exist two branches for \textit{strongly}-augmented images (called strong branches) and one branch for \textit{weakly}-augmented image (called weak branch). Similar to existing semi-supervised representation learning methods~\cite{sohn2020fixmatch, zhang2021flexmatch, xu2021dash}, we attempt to apply the consistency loss between a pair of each strong branch and weak branch. But, tailored to semi-supervised learning, we present a confidence-guided consistency regularization loss $\mathcal{L}_\mathrm{ccr}$ between two strong branches such that \begin{equation}
\begin{split}
    \mathcal{L}_\mathrm{ccr} = c_i(r) \mathcal{H}(q_i(r),p_\mathrm{model}(y|\mathcal{A}_j(r);\theta)) + c_j(r) \mathcal{H}(q_j(r),p_\mathrm{model}(y|\mathcal{A}_i(r);\theta)),
\end{split}
\label{equ:self_loss}
\end{equation}
where $q_{i}(r)$ and $q_{j}(r)$ denote the pseudo-labels generated from $p_\mathrm{model}(y|\mathcal{A}_i(r);\theta)$ and $p_\mathrm{model}(y|\mathcal{A}_j(r);\theta)$, respectively. $c_i(r)$ and $c_j(r)$ denote estimated confidences of $q_{i}(r)$ and $q_{j}(r)$. Our proposed loss function is different from conventional self-supervised representation learning loss $\mathcal{L}_\mathrm{self}$ in that the consistency is applied in the logit-level (not feature-level) similar to~\cite{sohn2020fixmatch,zhang2021flexmatch}, and adjusted by the estimated confidence. However, unlike~\cite{xie2020unsupervised, sohn2020fixmatch, zhang2021flexmatch,rizve2021defense,xu2021dash}, we can learn the better feature representation by considering two strongly-augmented views, while improving semi-supervised learning performance at the same time. It should be noted that this simple loss function can be incorporated with any semi-supervised learners, e.g., FixMatch~\cite{sohn2020fixmatch} or FlexMatch~\cite{zhang2021flexmatch}. 

To measure the confidences $c_i(r)$ and $c_j(r)$, we present two kinds of confidence estimators, based on non-parametric and parametric approaches. In the following, we explain how to measure these confidences in detail.

\subsection{Measuring Confidence: Non-parametric Approach}
Existing semi-supervised learning methods~\cite{lee2013pseudo, sajjadi2016regularization, sohn2020fixmatch} have selected unlabeled samples with high confidence as training targets (i.e., pseudo-labels) in a straightforward way; which can be viewed as a form of entropy minimization~\cite{grandvalet2004semi}. It has been well known that it is non-trivial to set an appropriate threshold for such handcrafted confidence estimation, and thus, confidence-based strategies commonly suffer from a dilemma between pseudo-label exploration and accuracy depending on the threshold~\cite{arazo2020pseudo,lucas2021barely}.

In our framework, estimating the confidence of pseudo-labels from strong branches may suffer from similar limitations if the conventional handcrafted methods~\cite{lee2013pseudo, sajjadi2016regularization,sohn2020fixmatch} are simply used. To overcome this, we present a novel way to measure the confidences, $c_i(r)$ and $c_j(r)$, based on the similarity between outputs of \textit{strongly}-augmented images and \textit{weakly}-augmented images. Based on the hypothesis that the similarity between the logits or probabilities from strongly-augmented images and weakly-augmented images can be directly used as a confidence estimator, we present to measure confidence of each strong branch loss by the cross-entropy loss value itself between strongly-augmented and weakly-augmented images. Specifically, we measure such a confidence with the following:

\begin{equation}
    s_i(r) = \frac{1}{\mathcal{H}(p_\mathrm{model}(y|\alpha(r);\theta),p_\mathrm{model}(y|\mathcal{A}_i(r);\theta))},
\end{equation}
where the smaller $\mathcal{H}(p_\mathrm{model}(y|\alpha(r);\theta),p_\mathrm{model}(y|\mathcal{A}_i(r);\theta))$, the higher $s_i(r)$ is. $s_j(r)$ can be similarly defined with $\alpha(r)$ and $\mathcal{A}_j(r)$. Finally, $c_i(r)$ is computed such that $c_i(r) = s_i(r)/(s_i(r) + s_j(r))$, and $c_j(r)$ is similarly computed. 

In this case, the total loss for the non-parametric approach is as follows:
\begin{equation}
    \mathcal{L}^\mathrm{np}_\mathrm{total} = \lambda_\mathrm{sup} \mathcal{L}_\mathrm{sup} + \lambda_\mathrm{un} \mathcal{L}_\mathrm{un} + \lambda_\mathrm{ccr} \mathcal{L}_\mathrm{ccr},
\end{equation}
where $\lambda_\mathrm{sup}$, $\lambda_\mathrm{un}$, and $\lambda_\mathrm{ccr}$ are weights for $\mathcal{L}_\mathrm{sup}$, $\mathcal{L}_\mathrm{un}$, and $\mathcal{L}_\mathrm{ccr}$, respectively. Note that for weakly-augmented labeled images $\alpha(x_b)$ with labels $y_b$, a simple classification loss $\mathcal{L}_\mathrm{sup}$ is applied as $\mathcal{H}(y_b,p_\mathrm{model}(y|\alpha(x_b);\theta))$, as done in~\cite{sohn2020fixmatch}.

\begin{algorithm}[t]
\caption{ConMatch-P (Parametric Approach)}
\label{alg:flexmatch}
\begin{algorithmic}[1]
\STATE{\textbf{Notation:} strong augmentation $\mathcal{A}$, weak augmentation $\alpha$, model $p_\mathrm{model}(\cdot;\theta)$ consisting of feature encoder $f$ and classifier $g$, confidence estimator $h(\cdot;\theta_{\mathrm{conf}})$}, pseudo label $q$, leranable confidence $c$
\STATE{\textbf{Input:} $\mathcal{X}=\{(x_b,y_b):b \in (1,\dots, B)\},\;\;\mathcal{U}=\{u_b:b \in (1,\dots,  \mu B)\}$}
% \COMMENT{M labeled data and N unlabeled data.}
% \COMMENT{Initialize predictions of all unlabeled data as -1 indicating unused.}
% \WHILE{not reach the maximum iteration}
% \FOR{$(\alpha(x_b),y_b), (\alpha(u_b),\mathcal{A}_i(u_b), \mathcal{A}_j(u_b)) $ in $\mathcal{X}$, $\mathcal{U}$}
\FOR {$b=1$ to $B$ } 
\STATE{$F(\alpha(x_b)), L(\alpha(x_b)) = f(\alpha(x_b)), g(f(\alpha(x_b))$)}
\STATE{$c(\alpha(x_b)) = h(F(\alpha(x_b)), L(\alpha(x_b));\theta_{\mathrm{conf}})$}
\IF {$y_b == \mathrm{argmax}_y{p_\mathrm{model}(y|\alpha(x_b);\theta))}$}
\STATE{$c_\mathrm{GT}(\alpha(x_b)) = 1 $} 
\ELSE 
\STATE {$c_\mathrm{GT}(\alpha(x_b)) = 0 $}
\ENDIF
\ENDFOR
\STATE{$\mathcal{L}_\mathrm{sup}=\sum_{b=1}^B \mathcal{H}(y_b,p_\mathrm{model}(y|\alpha(x_b);\theta))$}
\STATE{ $\mathcal{L}_\mathrm{conf-sup}= \mathcal{H}(c_\mathrm{GT}(\alpha{(x_b)}) ,h(F(\alpha(x_b)),L(\alpha(x_b));\theta_\mathrm{conf}))$}

\FOR {$b=1$ to $\mu B$ } 
\STATE{$(F_i, F_j), (L_i, L_j)  = f(\mathcal{A}_i(u_b), \mathcal{A}_j(u_b)), g(f(\mathcal{A}_i(u_b), \mathcal{A}_j(u_b)))$}
\STATE{$c_i, c_j =h(F_i, L_i;\theta_{\mathrm{conf}}), h(F_j, L_j;\theta_{\mathrm{conf}})$}
\STATE {Generate pseudo labels for differently augmented versions $\alpha, \mathcal{A}_i, \mathcal{A}_j$ }
\ENDFOR
\STATE{Calculate $\mathcal{L}_\mathrm{un}$ using $c, q$ from $\alpha({u_b})$, and $p_\mathrm{model}(y|\mathcal{A}(u_b);\theta)$} via Eq.~\ref{eqn:fixmatch-loss}
\STATE{Calculate $\mathcal{L}_\mathrm{ccr}$ using $c, q$ from $ \mathcal{A}(u_b)$ and $p_\mathrm{model}(y|\mathcal{A}(u_b);\theta)$ via Eq.~\ref{equ:self_loss}}
\STATE{Calculate $\mathcal{L}_\mathrm{conf}$ using $c$ from $\mathcal{A}(u_b)$ and $p_\mathrm{model}(y|\alpha(u_b);\theta)$ via Eq.~\ref{equ:l_conf_sup}}
\STATE{Update $\theta$ by minimizing $\mathcal{L}_\mathrm{sup}, \mathcal{L}_\mathrm{un}$ and $\mathcal{L}_\mathrm{ccr}$}
\STATE {Update $\theta_\mathrm{conf}$  by minimizing $\mathcal{L}_\mathrm{conf-sup}$ and $\mathcal{L}_\mathrm{conf}$ }
\STATE{\textbf{Return:} Model parameters \{$\theta$, $\theta_\mathrm{conf}$\}}
\end{algorithmic}
\label{alg:conmatch}
\end{algorithm}

\subsection{Measuring Confidence: Parametric Approach}
\label{sec:Parametric}
Even though the above confidence estimator with non-parametric approach yields comparable performance to some extent (which will be discussed in experiments), it solely depends on each image, and thus it may be sensitive to outliers or errors without any modules to learn a prior from the dataset. To overcome this, we present an additional parametric approach for confidence estimation. Motivated by stereo confidence estimation~\cite{poggi2016learning, seki2016patch, tosi2018beyond, choi2021adaptive}, obtaining a confidence measure from the networks by extracting the confidence features from input and predicting the confidence with a classifier, we also introduce a learnable confidence measure for pseudo-labels. Unlike existing methods that simply use the model output as confidence~\cite{xie2020unsupervised, sohn2020fixmatch, zhang2021flexmatch, xu2021dash, rizve2021defense}, such learned confidence can intelligently select a subset of pseudo-labels that are less noisy, which helps the network to converge significantly faster and achieve improved performance by utilizing the false negative samples excluded from training by high threshold at early training iterations.  

Specifically, we define an additional network for learnable confidence estimation such that $c(r) = h(F(r),L(r);\theta_\mathrm{conf})$, where $h(\cdot)$ is a confidence estimator with model parameters $\theta_\mathrm{conf}$, $F(r)$ is a feature, and $L(r)$ is a logit from an instance $r$, as shown in Fig.~\ref{fig:main-network}. For the network architecture, the concatenation of feature $F(r)$ and logit $L(r)$ transformed by individual non-linear projection heads is used, based on the intuition that a direct concatenation of two their heterogeneous confidence features does not provide an optimal performance~\cite{kim2018unified}, followed by the final classifier for confidence estimation. The detailed network architecture is described in the supplementary material.  

The confidence estimator is learned with the following loss function:
\begin{equation}
    \mathcal{L}_\mathrm{conf} = c_i(r) \mathcal{H}(p_\mathrm{model}(y|\alpha(r);\theta_\mathrm{freeze}),p_\mathrm{model}(y|\mathcal{A}_i(r);\theta_\mathrm{freeze})) + \mathrm{log}(1/c_i(r)),   
    \label{equ:l_conf_sup}
\end{equation}
where $\theta_\mathrm{freeze}$ is a freezed network parameter with a stop gradient. The intuition behind is that during the confidence network training, we just want to make the network learn the confidence itself, rather than collapsing to trivial solution to learn the feature encoder simultaneously. In addition, we also use the supervised loss for confidence estimator $\mathcal{L}_\mathrm{conf-sup}=\mathcal{H}(c_\mathrm{GT} ,h(F(\alpha(x_b)),L(\alpha(x_b));\theta_\mathrm{conf}))$; $c_\mathrm{GT}$ = 1 if $y_{b}$ is equal to $\mathrm{argmax}_y\, p_\mathrm{model}(y|\alpha(x_b);\theta)$, and $c_\mathrm{GT}=0$ otherwise. 

The total loss for the parametric case can be written as  
\begin{equation}
\mathcal{L}^\mathrm{param}_\mathrm{total} = \lambda_\mathrm{sup} \mathcal{L}_\mathrm{sup} + \lambda_\mathrm{un} \mathcal{L}_\mathrm{un} + \lambda_\mathrm{conf} \mathcal{L}_\mathrm{conf} + 
\lambda_\mathrm{conf-sup}
\mathcal{L}_\mathrm{conf-sup} +
\lambda_\mathrm{ccr} \mathcal{L}_\mathrm{ccr}  
\label{equ:conf} 
\end{equation}
where $\lambda_\mathrm{conf}$ and $\lambda_\mathrm{conf-sup}$ are the weights for $\mathcal{L}_\mathrm{conf}$ and $\mathcal{L}_\mathrm{conf-sup}$, respectively. 
We explain an algorithm for ConMatch of parametric approach in Alg.~\ref{alg:conmatch}.

\subsection{Stage-Wise Training}
\label{sec:Stage-Wise}
Even though our framework can be trained in an end-to-end manner, we further propose a stage-wise training strategy to boost the convergence of training. This stage-wise training consists of three stages, 1) pre-training for the feature encoder, 2) pre-training for the confidence estimator (for parametric approach only), and 3) fine-tuning for both feature encoder and confidence estimator (for parametric approach only). Specifically, we first warm up the feature encoder by solely using the standard semi-supervised loss functions with $\mathcal{L}_\mathrm{sup}$ and $\mathcal{L}_\mathrm{un}$. We then train the confidence estimator based on the outputs of the pre-trained feature encoder in the parametric approach. As mentioned in~\cite{kim2020adversarial}, this kind of simple technique highly boosts the convergence to discriminate between confident and unconfident outputs from the networks. Finally, we fine-tune all the networks with the proposed confidence-guided self-supervised loss $\mathcal{L}_\mathrm{ccr}$. We empirically demonstrate the effectiveness of the stage-wise training by achieving state-of-the-art results on standard benchmark datasets~\cite{krizhevsky2009learning,netzer2011reading,coates2011analysis}.

\section{Experiments}
\subsection{Experimental Settings}
% \vspace{-10pt}
In experiments, we extensively evaluate the performance of our ConMatch on various standard datasets~\cite{krizhevsky2009learning,netzer2011reading,coates2011analysis} with various label fraction settings in comparison to state-of-the-art algorithms, such as UDA~\cite{xie2020unsupervised}, FixMatch~\cite{sohn2020fixmatch}, FlexMatch~\cite{zhang2021flexmatch}, SelfMatch~\cite{kim2021selfmatch}, LESS~\cite{lucas2021barely} and Dash~\cite{xu2021dash}. Our proposed methods have two variants; ConMatch-NP (non-parametric approach), and ConMatch-P (parametric approach) integrated to FlexMatch~\cite{zhang2021flexmatch}, which is the state-of-the-art semi-supervised learner, even though it can be easily integrated to others~\cite{xie2019unsupervised,sohn2020fixmatch, li2021comatch}. 
\vspace{-10pt}
\subsubsection{Datasets.}
We consider four standard benchmarks, including CIFAR-10/100~\cite{krizhevsky2009learning}, SVHN~\cite{netzer2011reading}, and STL-10~\cite{coates2011analysis}.
CIFAR-10~\cite{krizhevsky2009learning} contains 50,000 training images and 10,000 test images, which have resolution 32$\times$32 with ten classes. Similar to CIFAR-10, CIFAR-100~\cite{krizhevsky2009learning} has the same number of training/test images and image size, but it differently classifies as 100 fine-grained classes. SVHN~\cite{netzer2011reading} consists of 73,257 training images with 26,032 test images, having also 32$\times$32 resolution images, belonging to ten different classes of numeric digits. STL-10~\cite{coates2011analysis} contains 5,000 labeled images with size of 96$\times$96 from 10 classes and 100,000 unlabeled images with size of 96$\times$96. \vspace{-10pt}

\subsubsection{Evaluation Metrics.}
For quantitative evaluation, we compute the mean and standard deviation of error rates, when trained on 3 different folds for labeled data, based on the standard evaluation protocol of selecting a subset of the training data while keeping the remainder unlabeled. In addition, as in~\cite{li2021comatch, zhang2021flexmatch}, we evaluate the quality of pseudo labels by training curves of precision, recall, and F1 values.\vspace{-10pt}

\subsection{Implementation Details}
For a fair comparison, we generally follow the same hyperparameters with FixMatch~\cite{sohn2020fixmatch}. Specifically, we use Wide ResNet (WRN)\cite{zagoruyko2016wide} as a feature encoder for the experiments, especially WRN-28-2 for CIFAR-10~\cite{krizhevsky2009learning} and SVHN~\cite{netzer2011reading}, WRN-28-8 for CIFAR-100\cite{krizhevsky2009learning}, and WRN-37-2 for STL-10~\cite{coates2011analysis}. We use a batch size of labeled data $B=64$, the ratio of unlabeled data $\mu=7$, and SGD optimizer with a learning rate starting from 0.03, The detailed hyperparameter settings are described in the supplementary material. For a weakly-augmented sample, we use a crop-and-flip, and for a strongly-augmented sample, we use RandAugmnet~\cite{cubuk2020randaugment}. \vspace{-10pt}

\subsection{Comparison to State-Of-The-Art Methods}
On standard semi-supervised learning benchmarks, we evaluate the performance of our frameworks, ConMatch-P and ConMatch-NP, compared to various state-of-the-art methods, as shown in Table~\ref{table:main_table-1} and Table~\ref{table:main_table-2}. We observe that the performance difference between ConMatch-NP and ConMatch-P is not large, except in the label-scare setting. This may be explained by the fact that non-parametric method highly depends on baseline performance since it does not consider other samples which can be modeled as a prior. We show our superiority on most benchmarks with extensive label setting, but we mainly focus the label-scare setting, since it corresponds to the central goal of semi-supervised learning, reducing the need for labeled data. We achieves {4.43}\% and {38.89}\% error rate for CIFAR-10 and CIFAR-100 settings~\cite{krizhevsky2009learning} with only 4 labels per class respectively. Compared to the results of SelfMatch~\cite{kim2021selfmatch} and CoMatch~\cite{li2021comatch}, closely related to ours, adopting self-supervised methods, we can prove the competitiveness of our method by achieving 2.38\% and 2.48\% improvements at CIFAR-10 with 40 labels.
On the other datasets, CIFAR-100~\cite{netzer2011reading} and STL-10~\cite{coates2011analysis}, we record the lowest error rate of 38.89\% and 25.39\% with 400 and 2500 labels setting, and also slightly better than baseline~\cite{zhang2021flexmatch} by recording 5.26\% in STL-10 dataset. \vspace{-10pt}
% To sum up, among the state-of-the-art methods~\cite{sohn2020fixmatch,zhang2021flexmatch,xu2021dash,lucas2021barely}, with the same training and evaluation settings, our method ConMatch achieves the most lower error rate at all datasets except SVHN.

\begin{table}[t]
\caption{
\textbf{Comparison on error rates on CIFAR-10~\cite{krizhevsky2009learning} and CIFAR-100~\cite{krizhevsky2009learning} benchmarks on 3 different folds.}}
\label{table:main_table-1}\vspace{-5pt}
\centering
\resizebox{0.85\textwidth}{!}{
\begin{tabular}{l|ccc|cc}
\toprule
&  \multicolumn{3}{c|}{\textbf{CIFAR-10}} & \multicolumn{2}{c}{\textbf{CIFAR-100}} \\
% &   \multicolumn{3}{c}{\# Labeled samples} & \multicolumn{3}{c}{\# Labeled samples} \\
\cline{2-4} \cline{5-6}  
Methods & 40 & 250 & 4,000  & 400 & 2,500  \\
\midrule

% $\Pi$-model~\cite{laine2016temporal}& - &54.26 $\pm$ 3.97 & 14.01 $\pm$ 0.38 & - &  57.25 $\pm$ 0.48 & 37.88 $\pm$ 0.11\\ % Fixmatch refer
% Pseudo-Labeling~\cite{lee2013pseudo}                & - & 49.78 $\pm$ 0.43 & 16.09 $\pm$ 0.28 & - & 57.38 $\pm$ 0.46\\ % Fixmatch refer
% Mixup~\cite{zhang2018mixup}                     & - & 47.43 $\pm$ 0.92 & 13.15 $\pm$ 0.20 & - & - & -  \\ % Featmatch refer
% VAT~\cite{miyato2018virtual}                    & - & 36.03 $\pm$ 2.82 & 18.68 $\pm$ 0.40 &  8.41 $\pm$ 1.01 &  5.98 $\pm$ 0.21 \\ % Not yet checked
% Mean Teacher~\cite{tarvainen2017mean}           & - & 32.32 $\pm$ 2.30 & 9.19 $\pm$ 0.19 &  - &  53.91 $\pm$ 0.57 & 35.83 $\pm$ 0.24 \\ %Fixmatch refer
% MixMatch~\cite{berthelot2019mixmatch}     & 47.54 $\pm$ 11.50 &  11.05 $\pm$ 0.86 &  6.42 $\pm$ 0.10 &  67.61 $\pm$ 1.32 &  39.94 $\pm$ 0.37 & 28.31 $\pm$ 0.33\\ % Fixmatch refer
UDA~\cite{xie2020unsupervised}       &  29.05$\pm$5.93 &  8.82$\pm$1.08 &  4.88$\pm$0.18 &  59.28$\pm$0.88 &  33.13$\pm$0.22   \\
% ReMixMatch~\cite{berthelot2019remixmatch}       &  19.10 $\pm$ 9.64 &  5.44 $\pm$ 0.05 &  4.72 $\pm$ 0.13 &  44.28 $\pm$ 2.06 &  27.43 $\pm$ 0.31 & 23.03 $\pm$ 0.56  \\ % Fixmath refer
FixMatch (RA)~\cite{sohn2020fixmatch}       &  13.81$\pm$3.37 &  5.07$\pm$0.65 & 4.26$\pm$0.06 &  48.85$\pm$1.75 &  28.29$\pm$0.11 \\ % Fixmatch refer
FlexMatch~\cite{zhang2021flexmatch}       &  4.97$\pm$0.06 &  4.98$\pm$0.09 &  4.19$\pm$0.01 & \underline{39.94$\pm$1.62} &  \underline{26.49$\pm$0.20} \\
SelfMatch~\cite{kim2021selfmatch}       &  6.81$\pm$1.08 &  4.87$\pm$0.26 &  \underline{4.06$\pm$0.08} &  - &  -  \\
CoMatch~\cite{li2021comatch}       &  6.91$\pm$8.47 &  4.91$\pm$0.33 &  - &  - &  -  \\

LESS~\cite{lucas2021barely}      & 6.80$\pm$1.10 & 4.90$\pm$0.80 & -& 48.70$\pm$12.40 & - \\
%have to be filled
Dash (RA)~\cite{xu2021dash}       &  13.22$\pm$3.75 &  \textbf{4.56$\pm$0.13} &  4.08$\pm$0.06 &  44.76$\pm$0.96 &  27.18$\pm$0.21 \\
\midrule
% [accuracy]
% ConMatch-NP                        &  95.14 &  -    &  95.64      &  57.98,60.27 &  72.13,74.05 \\
ConMatch-NP &  \underline{4.89$\pm$0.07} & 5.00$\pm$0.37 &  4.36$\pm$0.42  & 44.90$\pm$1.34 & 26.91$\pm$1.35 \\
% [accuracy]
% ConMatch-P                         &  95.32 &  95.3 &  95.4, 95.9 &  - &  -  \\
ConMatch-P   & \textbf{4.43$\pm$0.13}  & \underline{4.70$\pm$0.25} &\textbf{3.92$\pm$0.08}&\textbf{38.89$\pm$2.18}  & \textbf{25.39$\pm$0.20}
 \\
\bottomrule
\end{tabular}
}
\end{table}

\begin{table}[t]
\centering
\renewcommand{\arraystretch}{1}
\caption{
\textbf{Comparison on error rates on SVHN~\cite{netzer2011reading} and STL-10~\cite{coates2011analysis} benchamarks on 3 different folds. }
}
\label{table:main_table-2}\vspace{-5pt}
\resizebox{0.62\textwidth}{!}{
\begin{tabular}{l|cc|c}
\toprule
&  \multicolumn{2}{c|}{\textbf{SVHN}} & {\textbf{STL-10}}  \\
% &   \multicolumn{3}{c}{\# Labeled samples} & \multicolumn{3}{c}{\# Labeled samples} \\
\cline{2-3} \cline{4-4} 
Method & 40 & 250  & 1,000 \\
\midrule
UDA ~\cite{xie2020unsupervised} & 52.63$\pm$20.51 & 5.69$\pm$2.76 & 7.66$\pm$0.56\\ % Fixmatch refer
FixMatch (RA)~\cite{sohn2020fixmatch} & \textbf{3.96$\pm$2.17} & \underline{2.48$\pm$0.38} & 7.98$\pm$1.50\\ % Fixmatch refer
FlexMatch~\cite{zhang2021flexmatch}       &  4.97$\pm$0.06 &  4.98$\pm$0.09 & \underline{5.77$\pm$0.18}\\
SelfMatch~\cite{kim2021selfmatch}       &  3.42$\pm$1.02 &  2.63$\pm$0.43  &- \\
CoMatch~\cite{li2021comatch}       &  6.91$\pm$8.47 &  4.91$\pm$0.33 & 20.20$\pm$0.38  \\

% LESS~\cite{lucas2021barely}      & -& -& -&-  \\
Dash (RA)~\cite{xu2021dash}       &  \textbf{3.03$\pm$1.59} &  2.17$\pm$0.10 &7.26$\pm$0.40\\

\midrule
%5.34는 FIXMATCH, SEED=0 에서 가져옴.
ConMatch-NP                       &6.20$\pm$3.44 & 5.80$\pm$0.74 &  6.02$\pm$0.08 \\
ConMatch-P                      &\underline{3.14$\pm$0.57}& 3.13$\pm$0.72
 &  \textbf{5.26$\pm$0.04} \\
\bottomrule
\end{tabular}
}
\end{table}

\begin{table}[t]

\caption{
\textbf{Ablation study of different semi-supervised baselines.} We evaluate non-parametric (ConMatch-NP) and parametric (ConMatch-P) approaches with different baselines, Fixmatch~\cite{sohn2020fixmatch} and FlexMatch~\cite{zhang2021flexmatch}.
}
\label{table:ablation_baseline}\vspace{-5pt}
\centering
\resizebox{0.54\textwidth}{!}{
\begin{tabular}{l|cc|c}
\toprule
&  \multicolumn{2}{c|}{\textbf{CIFAR-10}} & \multicolumn{1}{c}{\textbf{CIFAR-100}}   \\
% &   \multicolumn{3}{c}{\# Labeled samples} & \multicolumn{3}{c}{\# Labeled samples} \\
% \cline{3-5} \cline{6-8} \cline{9-12} 
\cline{2-4}

Methods & 40 & 250  & 400  \\
\midrule

FixMatch~\cite{sohn2020fixmatch}            &  13.81  &  5.07  & 48.85 \\

ConMatch-NP w/~\cite{sohn2020fixmatch}       &  \underline{6.83} & 4.73  & \underline{48.73}   \\ 
ConMatch-P w/~\cite{sohn2020fixmatch}       &  \textbf{5.13} & {4.64} &  \textbf{48.00}\\ 
\midrule

FlexMatch~\cite{zhang2021flexmatch}         &4.97 & 4.98 & \underline{39.94} \\
ConMatch-NP w/~\cite{zhang2021flexmatch}     &\underline{4.84} &\underline{4.74} & 44.90\\ 
ConMatch-P w/~\cite{zhang2021flexmatch}     & \textbf{4.68} & \textbf{4.70} & \textbf{38.89}  \\ 
\bottomrule
\end{tabular}
}
\end{table}

% Comparison of our ConMatch with different backbone and stage-wise training:} \texttt{E} means end-to-end training and \texttt{S} means stage-wise training. We consider non-parametric (ConMatch-NP) and parametric (ConMatch-P) approaches with different baselines, Fixmatch~\cite{sohn2020fixmatch} and FlexMatch~\cite{zhang2021flexmatch}.

% Comparison of our ConMatch with different backbone and stage-wise training:} \texttt{E} means end-to-end training and \texttt{S} means stage-wise training. We consider non-parametric (ConMatch-NP) and parametric (ConMatch-P) approaches with different baselines, Fixmatch~\cite{sohn2020fixmatch} and FlexMatch~\cite{zhang2021flexmatch}.
\begin{table}[t]
\caption{
\textbf{Ablation study of training schemes.} \texttt{E} means end-to-end training and \texttt{S} means stage-wise training.
}
\label{table:ablation_status}\vspace{-5pt}
\centering

\resizebox{0.73\textwidth}{!}{
\begin{tabular}{l|c|cc|c}
\toprule
& \multirow{2}{*}{   Status   } & \multicolumn{2}{c|}{\textbf{  CIFAR-10  }} & \multicolumn{1}{c}{\textbf{  CIFAR-100  }}   \\
% &   \multicolumn{3}{c}{\# Labeled samples} & \multicolumn{3}{c}{\# Labeled samples} \\
% \cline{3-5} \cline{6-8} \cline{9-12} 
\cline{3-5}
Methods &  & 40 & 250  & 400  \\
\midrule
 \multirow{2}{*}{ConMatch w/ FixMatch~\cite{sohn2020fixmatch}}  
 & \texttt{E} &   4.85  &  4.77 &  47.81   \\ 
  & \texttt{S} & 5.13 & 4.60 & 48.00
\\ \hline
 \multirow{2}{*}{ConMatch w/ FlexMatch~\cite{zhang2021flexmatch}} & \texttt{E} & 4.68 & 4.70 &  57.16\\ 
  & \texttt{S} &  4.43 &  4.70 & 38.89  \\ 
\bottomrule
\end{tabular}
}
\end{table}

% Comparison of our ConMatch with different backbone and stage-wise training:} \texttt{E} means end-to-end training and \texttt{S} means stage-wise training. We consider non-parametric (ConMatch-NP) and parametric (ConMatch-P) approaches with different baselines, Fixmatch~\cite{sohn2020fixmatch} and FlexMatch~\cite{zhang2021flexmatch}.

\newcolumntype{M}[1]{>{\centering\arraybackslash}m{#1}}

\begin{table}[t]
\caption{
\textbf{Ablation study of our component on CIFAR-10~\cite{krizhevsky2009learning} with 40 labels.}
}
\label{table:ablation}\vspace{-5pt}
\centering

\resizebox{0.8\textwidth}{!}{
\begin{tabular}{M{.1\linewidth}|M{.14\linewidth}|M{.14\linewidth}|M{.14\linewidth}|M{.14\linewidth}|M{.14\linewidth}}
\toprule
& Three & Logit-level & \multicolumn{2}{c|}{Confidence net. input}  & \multirow{2}{*}{Error rate}\\
\cline{4-5}
  & branches & self-sup. & logits  & features & \\ \midrule
\textbf{(I)} & \markgood{\cmark} &  \xmark & \xmark & \xmark & 18.11 \\
\textbf{(II)} & \markgood{\cmark} &  \markgood{\cmark} & \xmark & \xmark & 77.50 \\ 
\textbf{(III)} & \markgood{\cmark} & \markgood{\cmark} & \markgood{\cmark} & \xmark & 7.05\\ 
\textbf{(IV)} & \markgood{\cmark} &  \markgood{\cmark} & \markgood{\cmark} & \markgood{\cmark} & 5.13\\ 
\bottomrule
\end{tabular}
}
\end{table}
\subsection{Ablation Study}
% \jy{
%  In this section, we conduct ablation studies on Confidence Guided Self-Supervision (CGS) to the different baselines and different training schemes. We also conduct ablation studies to evaluate the effectiveness of the each key component in our method. 
% \vspace{-10pt}
% }
\subsubsection{Effects of Different Baseline.}
We first evaluate our ConMatch with two baselines, FixMatch~\cite{sohn2020fixmatch} and FlexMatch~\cite{zhang2021flexmatch}, in both parametric (ConMatch-P) and non-parametric (ConMatch-NP) approaches as shown in Table~\ref{table:ablation_baseline}. ConMatch-P w/\cite{sohn2020fixmatch} boosts the performance significantly on CIFAR-10 with 40 labels from 13.81\% to 5.13\%, achieving the state-of-the-art result. The performance gains of ConMatch-P w/\cite{sohn2020fixmatch} is relatively higher than one w/\cite{zhang2021flexmatch} on most setting since~\cite{sohn2020fixmatch} does not adaptively adjust the threshold depending on the difficulty level of samples. Note that the thresholds of FixMatch~\cite{sohn2020fixmatch} and FlexMatch~\cite{zhang2021flexmatch} are used only for $\mathcal{L}_\mathrm{un}$.
%{, which is more likely to have confirmation bias.}
\vspace{-10pt}

\subsubsection{Effectiveness of Confidence Measure.}
In Table~\ref{table:ablation_baseline}, we evaluate two confidence measures in non-parametric and parametric approach. In extremely label-scare setting, such as CIFAR-10 with 4 labels per class, the non-parametric approach achieves relatively lower performance, {1.70}\% and {0.16}\%, in both FixMatch and FlexMatch baseline, while the parametric approach (ConMatch-P w/\cite{zhang2021flexmatch}) reaches the state-of-the-art performance. But, as the number of labels increases, the gap between non-parametric and parametric approach decreases, indicating that a certain number of labeled samples are required to measure the confidence without the confidence estimator.
\vspace{-10pt}

\subsubsection{Effectiveness of Stage-Wise Training.}
In Table~\ref{table:ablation_status}, we report the performance difference between end-to-end training and stage-wise training. We can observe that ConMatch-P has obtained meaningful enhancements in both training schemes, but stage-wise training shows more larger gap between baseline.\vspace{-10pt}
% Even in FlexMatch, the difference between the two training schemes maintains relatively smaller gap than that of FixMatch. Thus, based on the previous results, we can prove the effectiveness of the proposed stage-wise stage. 

%  Based on the fact that both $\mathcal{L}_{cgs}$ and $\mathcal{L}_{un}$ converge to almost 0, 
% We can confirm that the performance improvement of the parametric approach by showing {}\% and {}\% of CIFAR-10 with 250 labels and CIFAR-100 with 400 labels. Through the robustness and effectiveness as mentioned above, confidence measure based on the confidence estimator can boost the confidence guidance and also lead to the performance improvement.

\subsubsection{Architecture.}
Here we analyze the key components of ConMatch, the confidence estimator and guided consistency regularization as shown in Table~\ref{table:ablation}. For the fair comparison, we construct three branches on FixMatch~\cite{sohn2020fixmatch} as baseline \textbf{(I)}, one branch for a weakly-augmented sample and two branches for strongly-augmented samples. \textbf{(II)} uses logit-level self-supervised loss, but not weighted by confidence, i.e., $\mathcal{L}_\mathrm{ccr}$ with $c_i(r),c_j(r) = 1/2$. \textbf{(III)} and \textbf{(IV)} weight confidences of strongly-augmented instances to logit-level self-supervised loss. \textbf{(III)} only takes logits as an input of confidence estimator while both logits and features are fed into \textbf{(IV)}. 
% 이 때, 3은 estimator의 입력으로 logit만 받으며 4는 feature도 함께 받는다. 
The result of this ablation study shows that logit-level self-supervised loss without confidence guidance causes network collapse. The collapse occurred in \textbf{(II)} is one of the reasons why other semi-supervised methods~\cite{li2021comatch} could not use self-supervision at logit-level and should use negative pairs. \textbf{(III)} and \textbf{(IV)} show a significant performance improvement compared to \textbf{(I)} without such collapse.
\vspace{-10pt} 

% The collapse occurred in \textbf{(II)} 
% collapse occurs when loss is given without guidance at logit-level. 

% However, the aforementioned methods applied the consistency loss at the feature-level and do not consider the logit-level (as done in recent semi-supervised learning methods~\cite{sohn2020fixmatch,zhang2021flexmatch}). It is because a direction should be determined to assign a loss at the logit-level. To the best of our knowledge, we are the first to show the framework to apply a self-supervised learning framework at logit-level by guidance using estimated confidence.

% There is a performance difference between the self-supervised loss w/o confidence \textbf{(II)} and w/ confidence \textbf{(III)}. We also prove the effectiveness of another key component, confidence loss based on the difference in similarity between weak and strong augmentation by showing {}\% improvement compared to \textbf{(III)} and \textbf{(IV)}.

\begin{figure}[t!] 
\centering
    \renewcommand{\thesubfigure}{}
    \subfigure[(a) Precision ]
	{\includegraphics[width=0.33\linewidth]{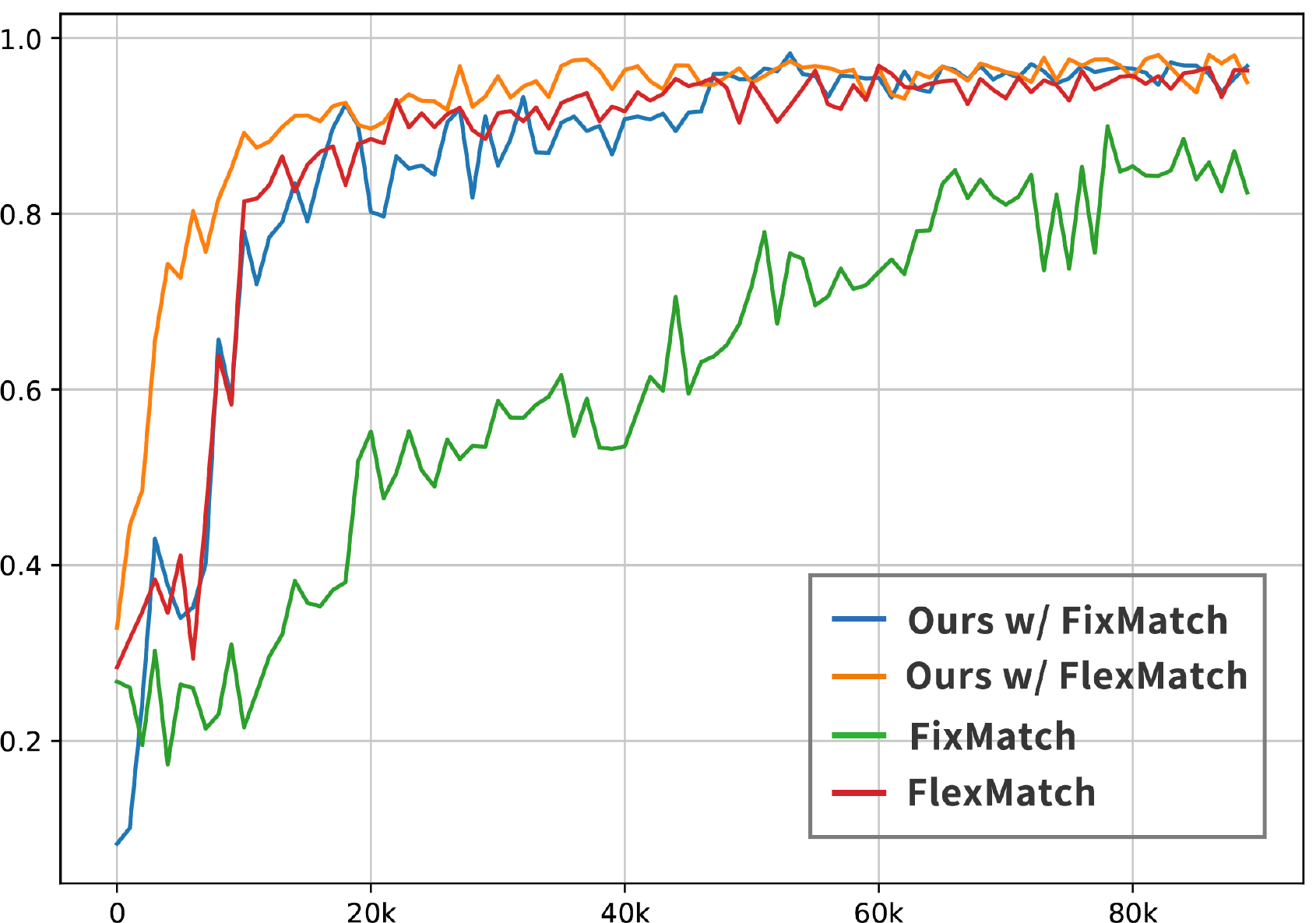}}\hfill
    \subfigure[(b) Recall ]
	{\includegraphics[width=0.33\linewidth]{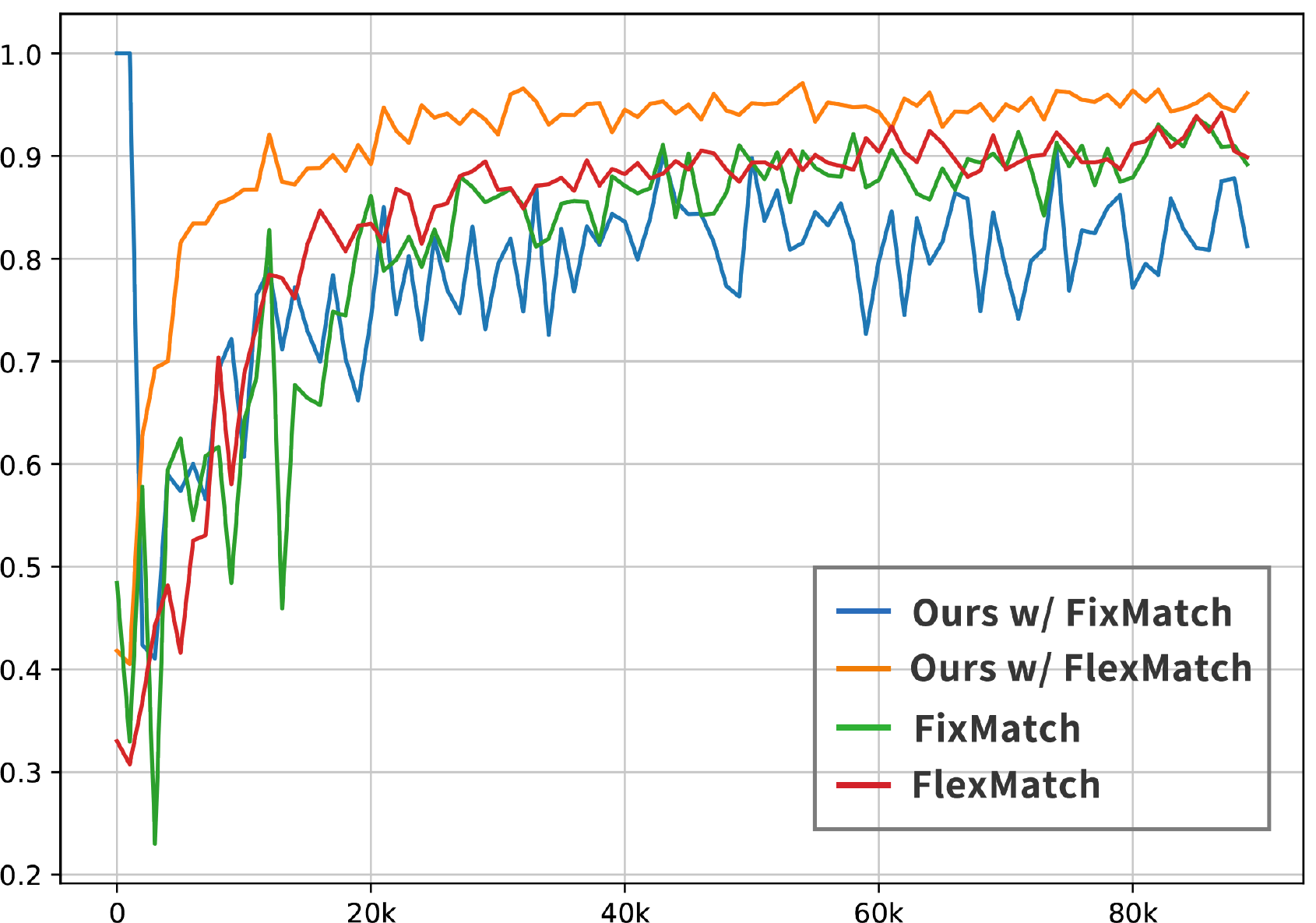}}\hfill	
	\subfigure[(c) F1-score ]
	{\includegraphics[width=0.33\linewidth]{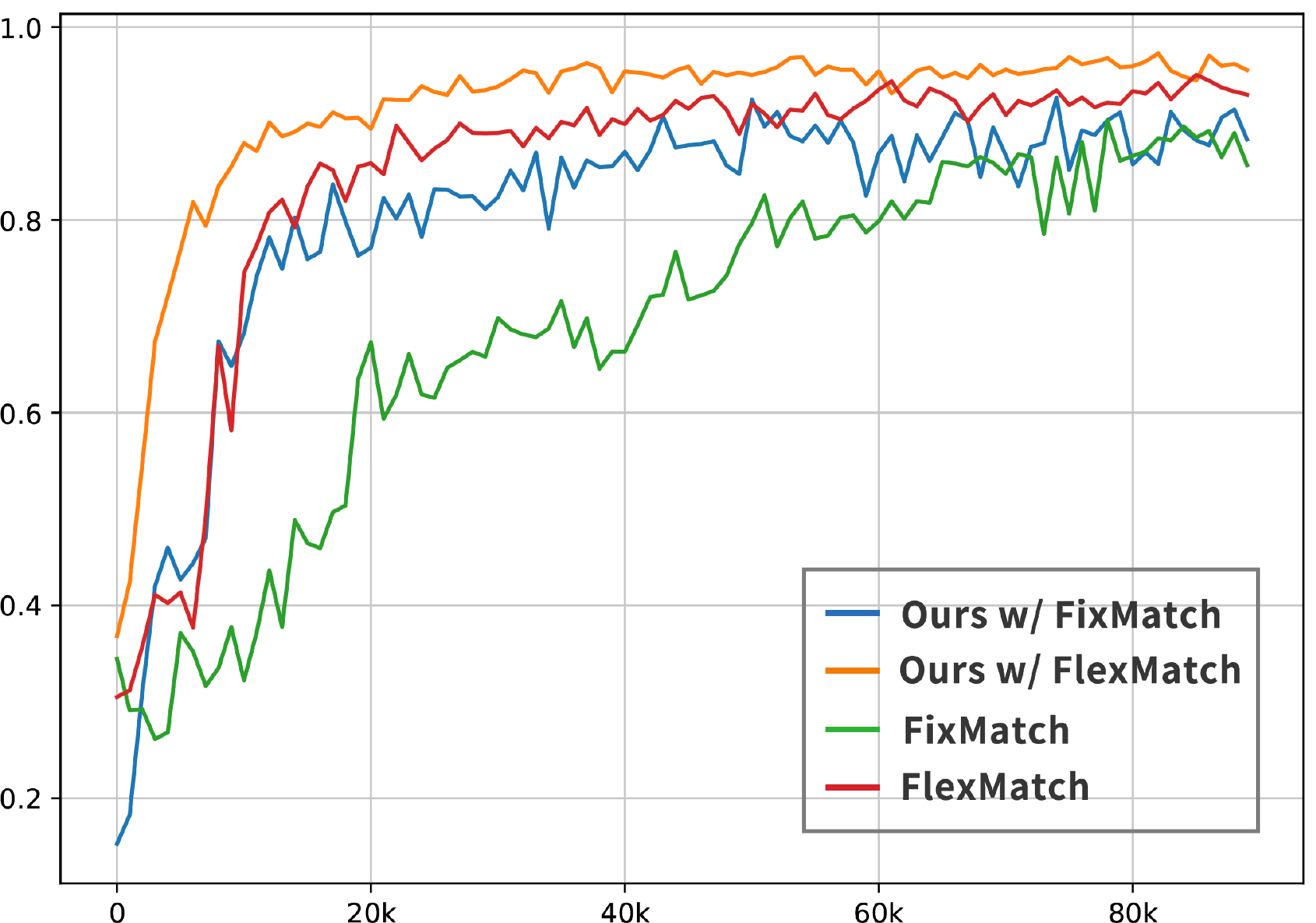}}\hfill\\
	\vspace{-10pt}
	\caption{\textbf{Plots of evolution of pseudo-labeling between ours and baselines~\cite{sohn2020fixmatch, zhang2021flexmatch} as training progresses on CIFAR-10~\cite{krizhevsky2009learning} with 40 labels:} in terms of (a) Precision, (b) Recall, and (c) F1-Score.}
	\label{fig:ana_2}
	\vspace{-5pt}    
\end{figure}

\subsubsection{Evaluating Confidence Estimation.}
To evaluate the effectiveness of our confidence estimator, we measure precision, recall, and F1-score of ConMatch and FixMatch~
\cite{sohn2020fixmatch} as evolving the training iterations on CIFAR-10~\cite{krizhevsky2009learning} with 40 labels as shown in Fig.~\ref{fig:ana_2}. The confident sample is defined as an unlabeled sample having max probability over than threshold in the baseline and confidence measures over than 0.5 in ConMatch. The quality of the confident sample is important to determine precisely to prevent the confirmation bias problem, significantly degrading the performances. The three classification metric, precision, recall and F1-score, are effective to evaluate the quality of the confidence. By Fig.~\ref{fig:ana_2}, we can observe that ConMatch, starting from the scratch for the fair comparison, shows higher values in all metric compared to the baseline.
 \vspace{-10pt}
 
% Next, the remained three components are added alternatively to analyze the effectiveness of each part of CGS. It is shown that the baseline shows the most high error rate of {}\%, and the improvement of the CGS loss are {}\%, and that of confidence learning is {}\%, which is {}\% larger than our final model. Also, we demonstrate the strength of parametric confidence learning by comparing our parametric \textit{learned} version and non-parametric version of our model, which shows the {}\% lower error rate than parametric model.

\begin{figure}[t!] 
\centering
    \renewcommand{\thesubfigure}{}
    \subfigure[(a)]
	{\includegraphics[width=0.25\linewidth]{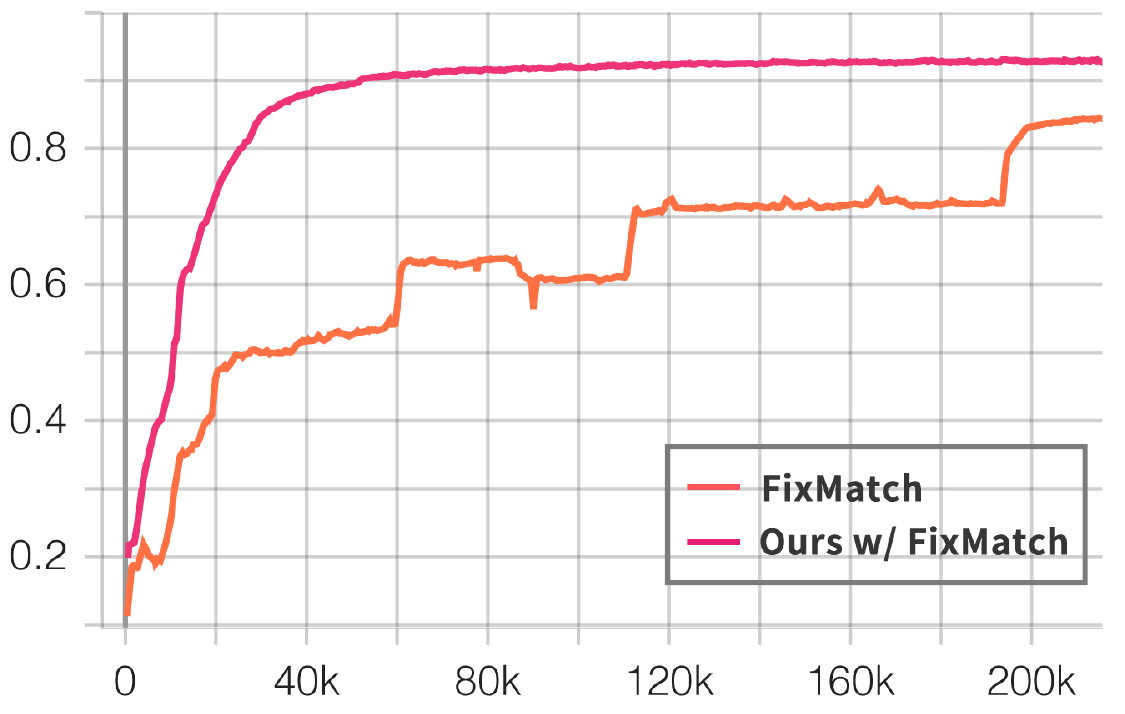}}\hfill
    \subfigure[(b)]
	{\includegraphics[width=0.25\linewidth]{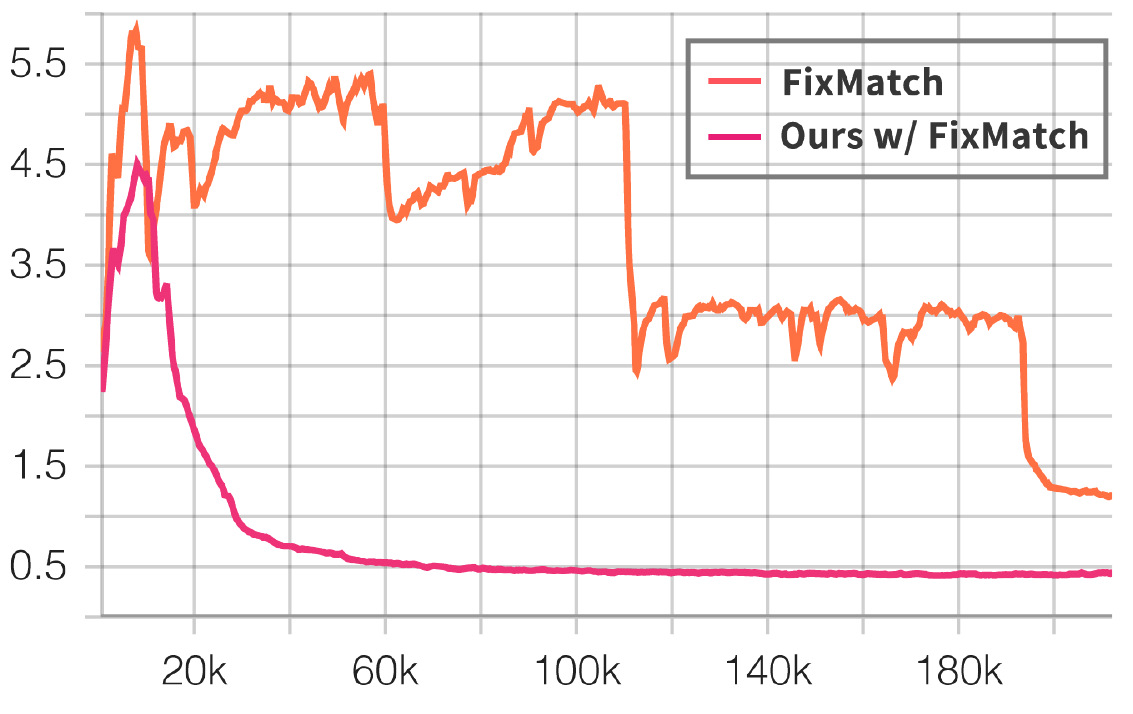}}\hfill
    \subfigure[(c)]
	{\includegraphics[width=0.25\linewidth]{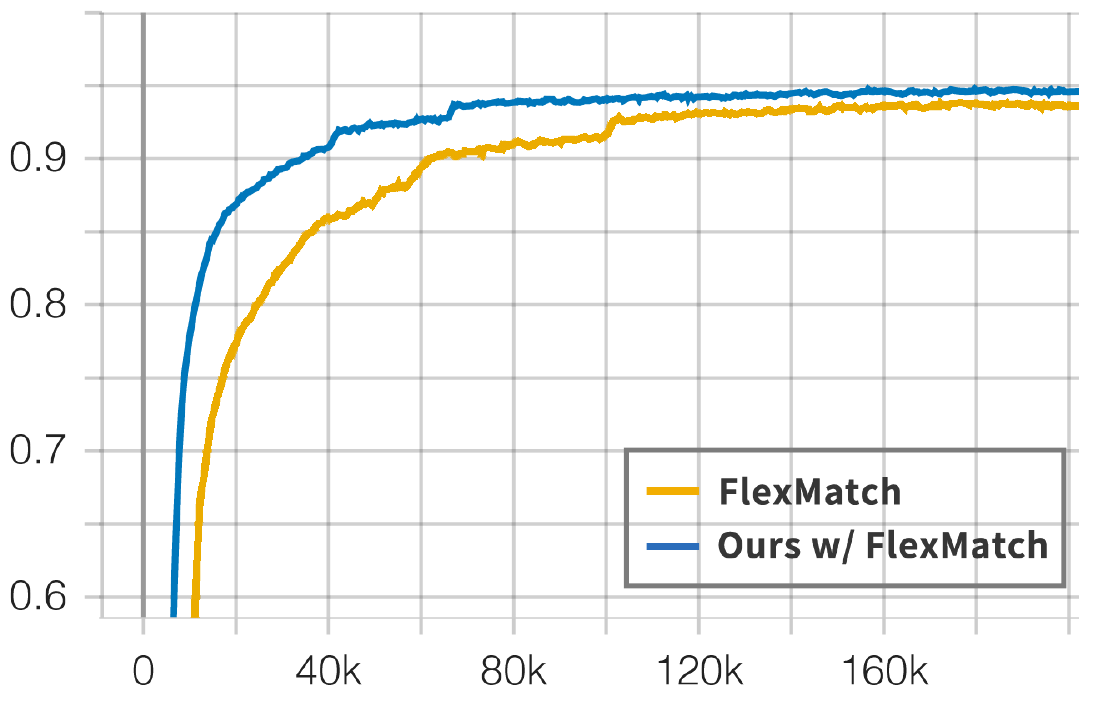}}\hfill
	\subfigure[(d)]
	{\includegraphics[width=0.25\linewidth]{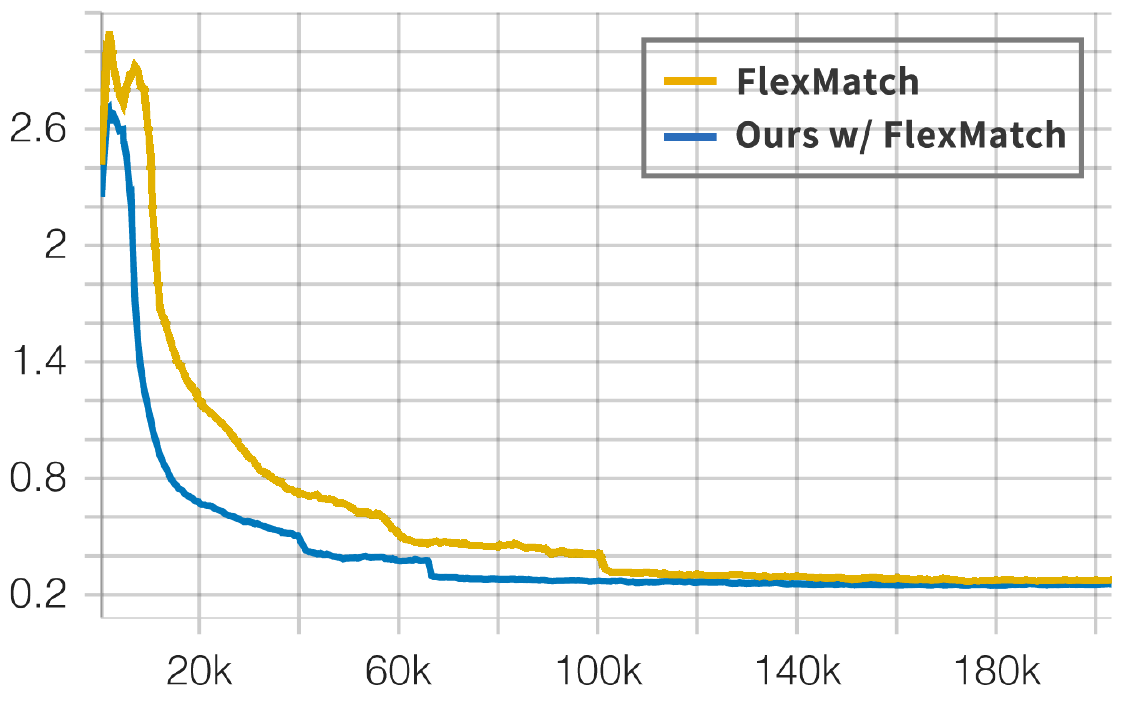}}\hfill\\
	\vspace{-10pt}
	\caption{\textbf{Convergence analysis of baselines~\cite{sohn2020fixmatch, zhang2021flexmatch} and ConMatch:} A comparison of top-1-accuracy and loss between  FixMatch~\cite{sohn2020fixmatch} and ConMatch w/~\cite{sohn2020fixmatch} are shown at (a) and (b). A comparison between FlexMatch~\cite{zhang2021flexmatch} and ConMatch w/\cite{zhang2021flexmatch} is shown at (c) and (d). Evaluations are done every {200}K iterations on CIFAR-10 with 40 labels.}
	\label{fig:ana_1}
	\vspace{-5pt}    
\end{figure}

\subsection{Analysis}
\subsubsection{Convergence Speed.}
One of the advantages of our ConMatch is its superior convergence speed. Based on the results as shown in Fig.~\ref{fig:ana_1} (b) and (d), the loss of ConMatch decreases much faster and smoother than corresponding baseline~\cite{sohn2020fixmatch}, demonstrating our superior convergence speed. Furthermore, the result of the accuracy in Fig.~\ref{fig:ana_1} (a) also proves that the global optimum is quickly reached. We also prove our effectiveness of our method by comparing the another baseline, FlexMatch~\cite{zhang2021flexmatch}. The convergence speed gap is relatively smaller than FixMatch since it dynamically adjust class-wise thresholds at each time step, leading to the stable training, but ConMatch achieves fast convergence at all time step from the early phase where the predictions of the model are still unstable. It is manifest that the introduction of ConMatch successfully encourages the model to proactively improve the overall learning effect.
\vspace{-10pt}
% \subsubsection{Visualization on Transition of Confidence.}
% To explain the effectiveness of confidence-guided consistency regularization between strongly-augmented pair, we show the feature distribution of ConMatch at different training iterations at Tab.~\ref{Tab:ana_tsne}. For visualizing the high dimensional features, the complex feature matrices are transformed into two-dimensional points based on t-SNE~\cite{van2008visualizing}. As training progresses, t-sne of labeled and weakly-augmented samples are clustered into distinct groups, relatively faster than baseline~\cite{sohn2020fixmatch}. And confident strongly-augmented samples are also well-organized near confident weakly-augmented samples. 

% The brightest color dots means the weakly-augmented instances, and the second brightest color means the strongly-augmented instances that having high confidence values than the pale-colored dots, whose confidence values are relatively lower. The high-confidence strongly augmented instances are more closer to high-confidence weekly-augmented instances than the low-confidence strongly-augmented instances. By this visualization, we can confirm that our confidence estimator effectively measures the confidence of samples and makes possible the training between the strongly augmented images. 

\section{Conclusion}
In this paper, we have proposed a novel semi-supervised learning framework built upon conventional consistency regularization frameworks with an additional strong branch to define the proposed confidence-guided consistency loss between two strong branches. 
To account for the direction of such consistency loss, we present confidence measures in non-parametric and parametric approaches. Also, we also presented a stage-wise training to boost the convergence of training. Our experiments have shown that our framework boosts the performance of base semi-supervised learners, and is clearly state-of-the-art on several benchmarks.
\vspace{-10pt}
\subsubsection{Acknowledgements.}
This research was supported by the MSIT, Korea (IITP-2022-2020-0-01819, ICT Creative Consilience program), and National Research Foundation of Korea (NRF-2021R1C1C1006897).

\setcounter{figure}{0}
\setcounter{table}{0}
\begin{center}
	\textbf{\Large Appendix}
\end{center}

In this supplementary document, we provide additional experimental results and implementation details to complement the main paper. The source code and a pre-trained model will be released in the near future.

\section*{Appendix A. Implementation Details - Hyper-parameters}
\vspace{-3pt}
For a fair comparison, we basically followed the default hyperparameters of our baselines, i.e., FixMatch~\cite{sohn2020fixmatch} and FlexMatch~\cite{zhang2021flexmatch} as demonstrated in the Sec 4.2 of the main paper. In addition to this, we also provide the best hyperparameters for the specific loss functions and each training stage, as shown in Table~\ref{Table:hyper-params}.\vspace{-5pt}
\begin{table}[h]
\begin{center}
\caption{\textbf{Additional hyper-parameter list of ConMatch}}
\begin{tabular}{c|c|cc}
\hline
Stage &          \multicolumn{2}{c}{hyper-parameters}                 \\ \hline
\multirow{2}{*}{Feature encoder pre-training} 
& $\lambda_\textrm{{sup}}$        & 1.0                         \\
& $\lambda_\textrm{{un}}$          & 1.0                     \\ \hline

\multirow{2}{*}{Confidence estimator pre-training} 
& $\lambda_\textrm{{conf}}$        & 0.1                         \\
& $\lambda_\textrm{{conf-sup}}$          & 1.0                     \\ \hline
\multirow{5}{*}{Fine-tuning} 
&$\lambda_\textrm{{sup}}$        & 1.0                        \\
& $\lambda_\textrm{{un}}$          & 1.0                           \\
& $\lambda_\textrm{{ccr}}$          &  1.0                       \\
& $\lambda_\textrm{{conf}}$        & 1.0                         \\
& $\lambda_\textrm{{conf-sup}}$          & 1.0                     \\ \hline
\end{tabular}
\label{Table:hyper-params}
\end{center}
\end{table}
\vspace{-30pt}
\section*{Appendix B. Additional Experimental Results}

\subsubsection{Results of ConMatch with FixMatch Baseline}
Since our framework basically works as a plug-and-play module, it can be combined with various semi-supervised learners. We experimented FlexMatch~\cite{zhang2021flexmatch}-based ConMatch (ConMatch with FlexMatch) on general semi-supervised learning benchmarks in Table 1 and Table 2 of the main paper. In this supplementary material, we additionally provide the benchmark results of FixMatch~\cite{sohn2020fixmatch}-based ConMatch (ConMatch with FixMatch) on SVHN and STL-10 datasets. On SVHN with 40 labels and 250 labels, ConMatch with FixMatch records the state-of-the-art accuracy of 96.75\% and 98.14\% respectively as displayed in Table~\ref{table:svhn_stl}. We could observe that ConMatch with FixMatch shows similar or even better results than ConMatch with FlexMatch on SVHN dataset. This is mainly due to the weakness of FlexMatch~\cite{zhang2021flexmatch} on the \textit{unbalanced} dataset. While training feature encoder of FlexMatch on SVHN, class-wise imbalance of SVHN leads the classes with fewer samples to have low thresholds. Such low thresholds allow noisy pseudo-labeled samples to be learned throughout the training process and eventually reduce the accuracy of model's prediction. A feature encoder of FixMatch, on the other hand, is not affected by class-wise imbalance of dataset, since it fixes its threshold at 0.95 to filter out noisy samples. Consequently, this performance gap between two encoders caused by class-wise imbalance of dataset allows ConMatch with FixMatch to achieve similar or even higher accuracy score than ConMatch with FlexMatch.  \vspace{-10pt}
\begin{table}[t]
\centering
\renewcommand{\arraystretch}{1}
\caption{
\textbf{Comparison on error rates on SVHN~\cite{netzer2011reading} and STL-10~\cite{coates2011analysis} benchamarks on 3 different folds. }
}
\label{table:svhn_stl}\vspace{-5pt}
\resizebox{0.7\textwidth}{!}{
\begin{tabular}{l|cc|c}
\toprule
&  \multicolumn{2}{c|}{\textbf{SVHN}} & {\textbf{STL-10}}  \\
% &   \multicolumn{3}{c}{\# Labeled samples} & \multicolumn{3}{c}{\# Labeled samples} \\
\cline{2-3} \cline{4-4} 
Method & 40 & 250  & 1,000 \\
\midrule
FixMatch~\cite{sohn2020fixmatch}  & 3.96$\pm$2.17 & 2.48$\pm$0.38 & 7.98$\pm$1.50\\  
FlexMatch~\cite{zhang2021flexmatch}       &  4.97$\pm$0.06 &  4.98$\pm$0.09 & 5.77$\pm$0.18\\

% \midrule 
%5.34는 FIXMATCH, SEED=0 에서 가져옴.
ConMatch w/FixMatch~\cite{sohn2020fixmatch} & 3.25$\pm$1.06 & \textbf{1.86$\pm$0.06} & 5.90$\pm$0.17  \\
ConMatch w/FlexMatch~\cite{zhang2021flexmatch}                     &\textbf{3.14$\pm$0.57}& 3.13$\pm$0.72
 &  \textbf{5.26$\pm$0.04} \\
\bottomrule
\end{tabular}
}
\end{table}

\subsubsection{Class-wise Results with Other SSL Techniques}
In Table 2 and Table 3 of the main paper, we demonstrated that semi-supervised learners~\cite{sohn2020fixmatch, zhang2021flexmatch} combined with ConMatch outperform their baselines by a significant margin in most SSL benchmark settings. Additionally, to provide a detailed analysis for our performance results on CIFAR-10, we also performed a per-class quantitative evaluation as shown in Table~\ref{table:per-class}. ConMatch w/~\cite{sohn2020fixmatch} achieves performance improvement over FixMatch in all the classes except for the \textit{dog} class. Although FixMatch seems to predict \textit{dog} class better than any other models, its prediction for the \textit{cat} class, which is very similar to \textit{dog}, obtains significantly low accuracy score. This result shows that Fixmatch suffers from a confirmation bias, and its prediction accuracy for \textit{dog} class is a distorted value. ConMatch w/~\cite{sohn2020fixmatch} on the other hand, records high accuracy score for all classes including \textit{cat} class. This verifies that ConMatch w/~\cite{sohn2020fixmatch} effectively alleviate the confirmation bias by making use of confidence estimator. Furthermore, ConMatch w/~\cite{zhang2021flexmatch} outperforms its baseline~\cite{zhang2021flexmatch} on all classes. Based on all these class-wise evaluation and comparison results, we could confirm the effectiveness of combining ConMatch with existing semi-supervised learners. \vspace{-10pt}
%ConMatch w/~\cite{fix} 와 ConMatch w/~\cite{flex} 둘다 모든 클래스에 대해 고른 prediction을 보이는 것을 통해 confidence estimator를 통해서 효과적으로 confirmation bias를 해결한 것이라고 해석할 수 있다. 

\begin{table}[t!]
\begin{center}
\caption{\textbf{Per-class quantitative evaluation on ConMatch and base semi-supervised learners~\cite{sohn2020fixmatch, zhang2021flexmatch} on CIFAR-10 with 40 labels.} The best results are in bold.}
\label{table:per-class}
\begin{tabular}{c|cccccccccc|c}
\toprule
Methods  & air. & auto.  & bird & cat  & deer & dog  & frog  & horse & ship & truck & total \\ 
\midrule
\multicolumn{1}{c|}{FixMatch~\cite{sohn2020fixmatch}} & 94.3 & 97.6 &72.6 & 28.1 & 96.6 & \textbf{94.1}& 98.1 & 95.6 & 97.5 & 96.6 &87.11 \\
\multicolumn{1}{c|}{FlexMatch~\cite{zhang2021flexmatch}} & 96.8 & 98.1 & 92.3 & 88.0 & 95.8 & 88.7 & 98.3 & 97.1 &98.4 &97.2  & 95.07 \\
 \hline
\textbf{ConMatch w/\cite{sohn2020fixmatch}} & \textbf{97.2} & 97.7 & 90.9 & 86.0 & \textbf{96.9} & 88.6 & 98.7 & {97.2}  & {97.6}& \textbf{97.9} & {94.87} \\
\textbf{ConMatch w/\cite{zhang2021flexmatch}}&97.1  & \textbf{98.5} & \textbf{92.9} & \textbf{88.1} & 96.0 &  91.8 & \textbf{98.9} & \textbf{97.6} & \textbf{98.7} & 97.4 & \textbf{95.70}
\\ \bottomrule
\end{tabular}    
\end{center}\vspace{-5pt}
\vspace{-10pt}
\end{table}

\subsubsection{Convergence Time.}
In this section, we analyze the effect of performance-boosting of ConMatch on end-to-end training. Specifically, we analyze the time taken to achieve the best accuracy of FixMatch~\cite{sohn2020fixmatch}, 86.19\% in CIFAR-10 40 labels for a fair comparison between FixMatch~\cite{sohn2020fixmatch}, FlexMatch~\cite{zhang2021flexmatch}, and ConMatch due to their different convergence speeds. As seen from Table~\ref{table:convtime}, FixMatch~\cite{sohn2020fixmatch} takes 261.9k iters to converge, 1,699 minutes, while ConMatch w/FixMatch~\cite{sohn2020fixmatch} converges at 27.1k iters, 117 minutes, which means that ConMatch can boost convergence about 14.5 times faster. Additionally, ConMatch w/FlexMatch~\cite{zhang2021flexmatch} converges at 20.5k iters, 89 minutes, which is 1.5 times faster than FlexMatch~\cite{zhang2021flexmatch} which converges at 44.6k iters, 138 minutes.
% 27.1k 1H 57min 58sec < Fix+Con (117min) 
% 261.9k 1Day 4H 19min 7sec < FixMatch Original @ CIFAR-10 40lbs (1699min)
% 14.5x faster!
\begin{table}[t]
\caption{
\textbf{Comparison of Convergence Time between FixMatch and ConMatch w/FixMatch} Convergence time means the time taken to reach the highest accuracy of FixMatch, 86.19\%, in CIFAR-10 40 labels. Runtime is measured in the same setting using RTX 3090 Ti. The best results are in bold.}
\label{table:convtime}\vspace{-5pt}
\centering

\resizebox{0.6\textwidth}{!}{
\begin{tabular}{c|c|c}
\toprule
Method & Iter & Convergence Time    \\
\midrule
 FixMatch  & 261.9k  & 1,699 mins \\
 ConMatch w/FixMatch  &  \textbf{27.1k}  & \textbf{117 mins} \\ \midrule
 FlexMatch  & 44.6k  & 138 mins \\
 ConMatch w/FlexMatch  & \textbf{20.3k}  & \textbf{89 mins} \\
\bottomrule
\end{tabular}
}
\end{table}

\begin{table}[t]
\caption{
\textbf{Ablation study on warm\_up iterations for feature encoder.} The best results are in bold.}
\label{table:ab_warmup}\vspace{-5pt}
\centering
\resizebox{0.4\textwidth}{!}{
\begin{tabular}{c|cc}
\toprule
\multirow{2}{*}{  Warm\_up iter. } & \multicolumn{2}{c}{\textbf{  CIFAR-10  }}    \\
% &   \multicolumn{3}{c}{\# Labeled samples} & \multicolumn{3}{c}{\# Labeled samples} \\
% \cline{3-5} \cline{6-8} \cline{9-12} 
\cline{2-3}
 &   \multicolumn{2}{c}{40}   \\
\midrule
 0  &   \multicolumn{2}{c}{93.24} \\
1000 &   \multicolumn{2}{c}{93.78} \\
4000 &   \multicolumn{2}{c}{\textbf{94.87}} \\
5000  &   \multicolumn{2}{c}{93.29} \\
10000  &   \multicolumn{2}{c}{93.10} \\
 20000  &   \multicolumn{2}{c}{92.34} \\

\bottomrule
\end{tabular}
}\vspace{-10pt}
\end{table}

\section*{Appendix C. Ablation Studies}
\subsubsection{Warm-up Iteration}
As mentioned in Sec.3.5 of the main paper, we adopt the warm-up stage when training, to stabilize the confidence estimator at the early stage of training and to boost the convergence of training. Therefore, it can be the question that how many iterations of pre-training are we needed, so we perform an ablation study of these warm-up iterations at CIFAR-10 40 labels setting, and the results are shown in Table~\ref{table:ab_warmup}. After warm-up iteration, we perform the pre-training stage for confidence estimator for 10,000 iterations equally for all ablation settings except for the end-to-end training strategy. While without warm-up stage training results the accuracy of 93.24\%, pre-training the feature extractor 10,000 iterations and the confidence estimator 10,000 iterations results 94.87\% accuracy, showing {1.63}\% improvement compared to the end-to-end training strategy. \vspace{-10pt}
%warm-up iteration에 따라 feature encoder에서 confidence estimator에게 제공해주는 정보의 양의 퀄리티에 따라 성능의 차이가 사소하지 않다는 것을 알 수 있었다.
%(초기 iter가 좋으면) semi-supervised setting에서 few number of labels에 대한 prediction은 빠르게 학습되기때문에 logit distribution이 overfit되기전인 초기 iteration에서 confidence estimator pretraining stage에 들어가는 것이 좋다.
%(중간 iter가 좋으면)
%labeled class distribution은L_ conf_sup에 의해서 supervised manner로 정확하게 학습될 수 있는 반면 unlabeled class distribution은 L_conf에 의해 probabilistic manner로 학습되기때문에 feature encoder에서 생성되는 feature와 logit가 informative representation으로 생성되는 iteration에서 confidence estimator pretraining stage에 들어가는 것이 좋다.

%We noticed in our experiments that at the early stage of the training, the model may blindly predict most unlabeled samples into a certain class depending on the parameter initialization(i.e., more likely to have confirmation bias). Hence, the estimated learning status may not be reliable at this stage. Therefore, we introduce a warm-up process by rewriting the denominator in Eq. (6) as:

% Also, only pre-training the feature extractor or the confidence estimator results {}\% or {}\% accuracy, respectively. 
% Finally, we find that pre-training the feature extractor for $0000$ iterations and the confidence estimator for $00000$ iterations shows the best results.
\begin{figure}[t] 
\centering
    % \renewcommand{\thesubfigure}{}
% 	\subfigure[ \makecell{(a) Semi-Sup.\\(FixMatch)}]
	\subfigure[Basic.]
	{\includegraphics[width=0.23\linewidth]{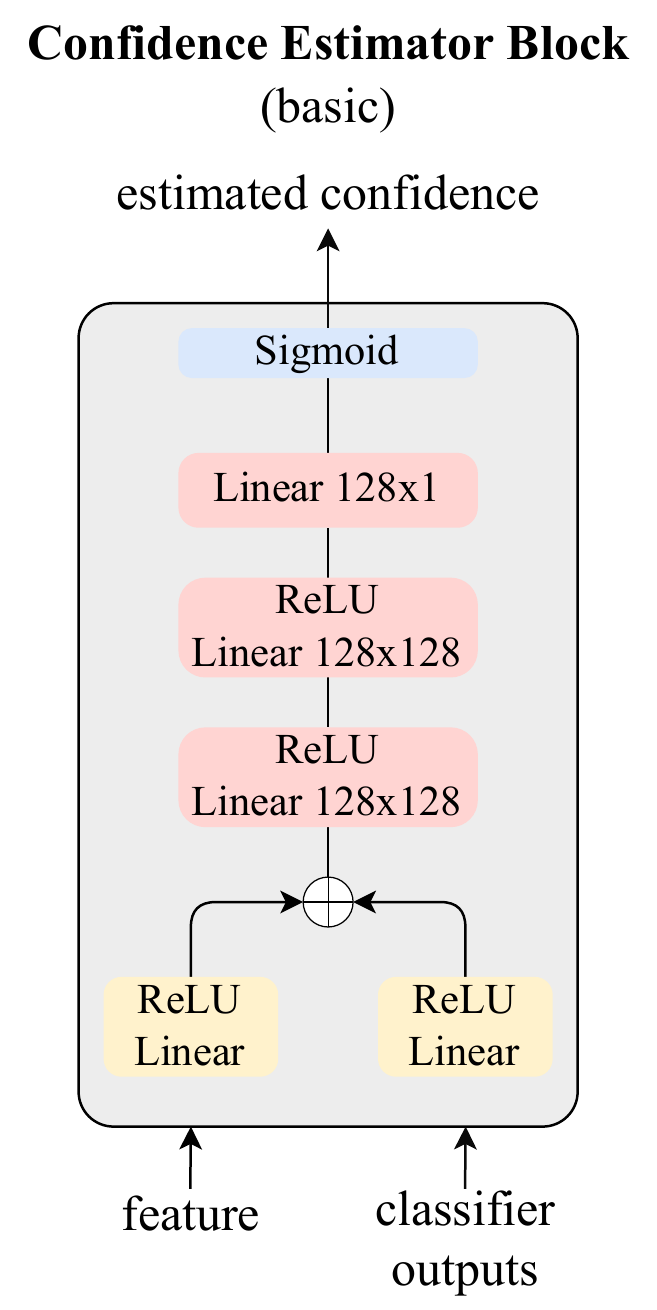}}\hfill
    % \subfigure[\makecell{(b) Self-Sup.\\(Simsiam)}]
	\subfigure[Reduced.]
	{\includegraphics[width=0.23\linewidth]{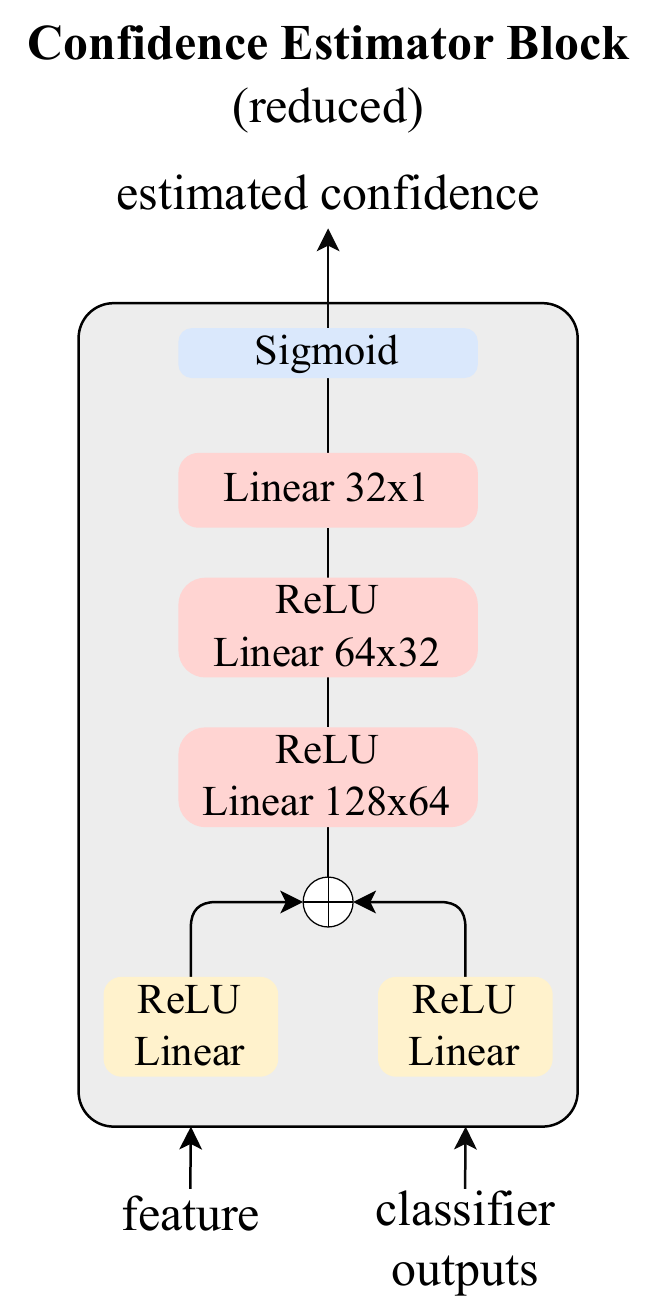}}\hfill
	\subfigure[Top-k.]
	{\includegraphics[width=0.23\linewidth]{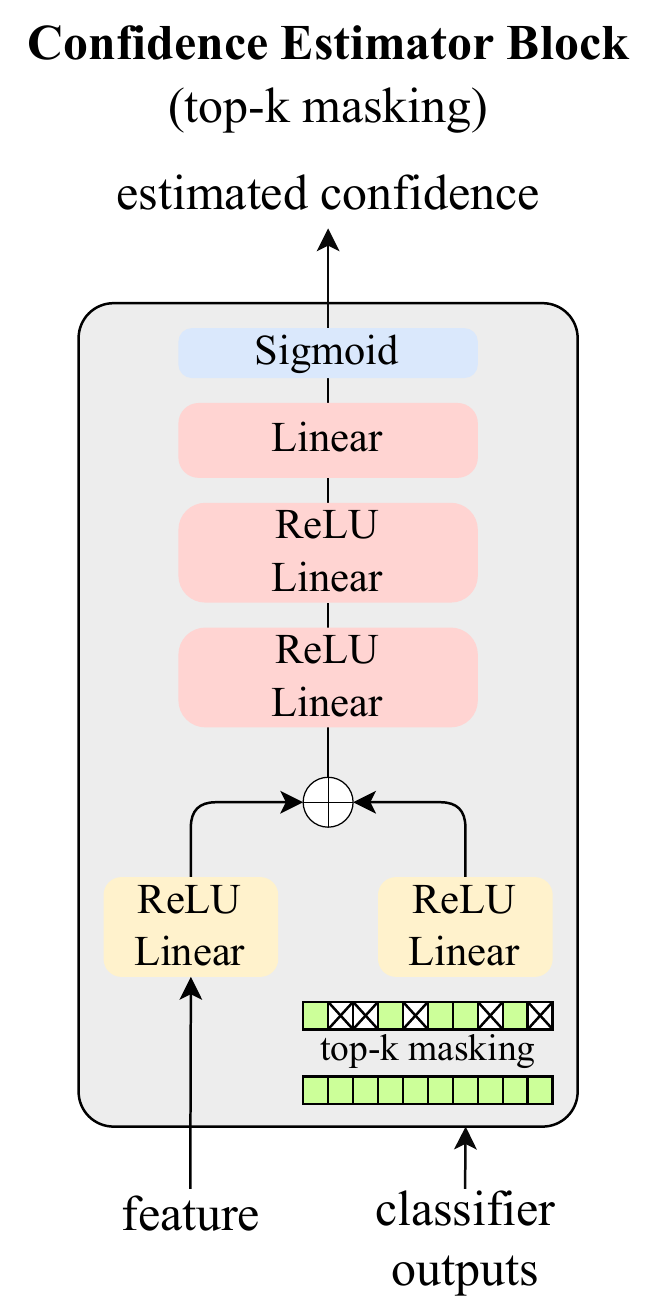}}\hfill    
    \subfigure[BatchNorm.]
	{\includegraphics[width=0.23\linewidth]{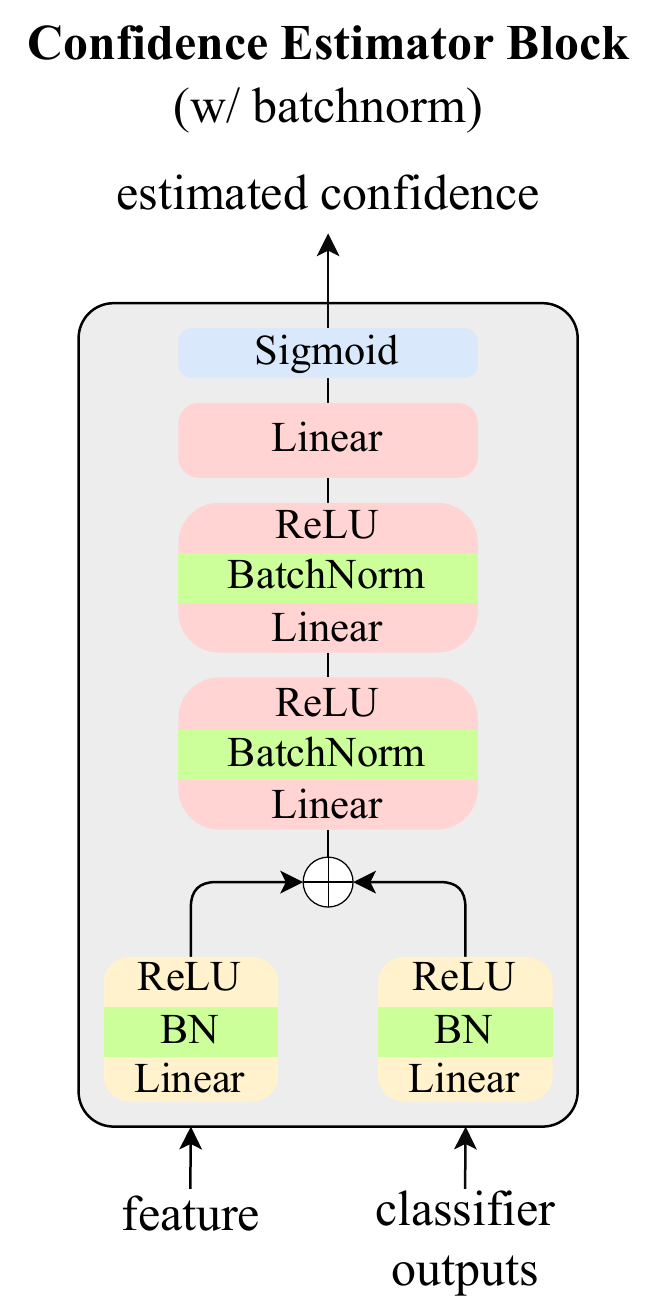}}\hfill
% 	\subfigure[\makecell{(c) Semi with Self-sup.\\(CoMatch)}]

	\vspace{-10pt}
	\caption{\textbf{Network Architecture Variants for Confidence estimator:} (a) Basic-The number of parameters of linear layers is always constant. (b) Reduced-The number of parameters decreases to a multiplier of 2 according to the depth of the linear layers. (c)Top-k - The top-k values of the classifier outputs are only used for confidence estimator, by masking class values that do not belong to the top-k. (d)BatchNorm.-Batch normalization layer is followed by the linear layer.  }
	\label{Fig:conf_net}
	\vspace{-5pt}    
\end{figure}
\subsubsection{Designing Confidence Estimator Network}
The overall confidence estimation network, consisting of two sub-networks, is built as the collection of linear layer and ReLU activations. To input the logit values and features into the confidence estimator, it includes feature projection embedding and logit projection embedding. 
For the first variant of the confidence estimator, as shown in Fig.~\ref{Fig:conf_net}(b), we reduce the channel dimension of each linear layer of the confidence estimator by a factor of two, which is called \textit{reduced} version. It shows {93.26}\% accuracy with the same evaluation setting, which is {0.91}\% increased results compared to the basic architecture as shown in Table~\ref{table:ablation_simfunc} (I) and (II). Given the distributions of all classes from the classifier, we find that passing a part of it, especially masking the lower values of logit makes the estimator more robust and working well, shown in Fig.~\ref{Fig:conf_net}(c). The performance gap between (II) and (III) proves that the top-k approach is effective for the confidence estimator. We also find that the best top-k value is set by k=5 and that \textit{masking} top-k, logit dimension is same as the number of class but masking class value, not belonging to top-k, is better than \textit{reducing} top-k, logit dimension is same as top-k. For the final variant of the confidence estimator, we test the effect of BatchNorm~\cite{ioffe2015batch} that came after the linear layer and a detailed architecture for this version is shown in Fig.~\ref{Fig:conf_net}(d). As shown in Table~\ref{table:ablation_simfunc}, (IV) record the best accuracy, 2.52\%, 1.61\%, and 0.93\% better than (I), (II) and (III), respectively. \vspace{-10pt}

% We perform the ablation study for this masking strategy for the CIFAR-10 40 labels setting, and it is shown at Table.~\ref{}. 

% Passing the highest top-5 values of logit shows the best results of {}\%, and passing the logits as it is shows the accuracy of {}\%. Also, passing the logit values only containing top 5 values, without masking it so make the low values remain zero, its performance significantly decreased to {}\%. 

\subsubsection{Similarity Functions in Non-parametric Approach}
In this section, we analyze the similarity function of the confidence estimation module in a non-parametric approach. We assume that the confidence of a strongly-augmented sample can be represented by the similarity between a weakly-augmented sample $\alpha(r)$ and a strongly-augmented sample $\mathcal{A}(r)$ in a probabilistic way. As seen in Eq.(4), we define a similarity function as a cross-entropy loss between the logit of weakly-augmented sample and the logit of the strongly-augmented sample. The similarity function can be substituted in other ways. Specifically, it can be formulated by L2 distance like $\|p_\mathrm{model}(y|\mathcal{A}(r);\theta)-p_\mathrm{model}(y|\alpha(r);\theta_\mathrm{freeze})\|$ or feature cosine-similarity like $\frac{p_\mathrm{model}(y|\mathcal{A}(r);\theta) \cdot p_\mathrm{model}(y|\alpha(r);\theta_\mathrm{freeze})}{\|p_\mathrm{model}(y|\mathcal{A}(r);\theta)\|\|p_\mathrm{model}(y|\alpha(r);\theta_\mathrm{freeze})\|}$ where $p_\mathrm{model}$ is considered as logit and feature, respectively.  Table~\ref{table:ablation_simfunc} shows the results of replacing the similarity function with the L2 distance of logits and feature cosine-similarity. Through the result, we can find that the similarity function based on cross-entropy is the most effective confidence measure, compared to others.

\begin{table}[t]
\caption{
\textbf{Ablation study on network architecture variants for confidence estimator.}The best results are in bold. }
\label{table:ablation_simfunc}\vspace{-5pt}
\centering
\resizebox{0.8\textwidth}{!}{
\begin{tabular}{c|c|c|c|c|cc}
\toprule
\multirow{2}{*}{ } & \multicolumn{4}{c|}{   Network Arch. for Confidence estimator  } & \multicolumn{2}{c}{\textbf{  CIFAR-10  }}    \\
% &   \multicolumn{3}{c}{\# Labeled samples} & \multicolumn{3}{c}{\# Labeled samples} \\
% \cline{3-5} \cline{6-8} \cline{9-12} 
\cline{2-5} \cline{6-7}
 &  Basic & Reduced & Top-K & BatchNorm & \multicolumn{2}{c}{40}   \\
\midrule
% (I) & \xmark &\xmark &\xmark &\xmark &\multicolumn{2}{c}{??} \\
(I) & \cmark &\xmark &\xmark &\xmark &\multicolumn{2}{c}{92.35} \\
(II) & \xmark &\cmark &\xmark &\xmark &\multicolumn{2}{c}{93.26} \\
(III) & \xmark &\cmark &\cmark &\xmark &\multicolumn{2}{c}{93.94} \\
(IV) & \xmark &\cmark &\cmark &\cmark &\multicolumn{2}{c}{\textbf{94.87}} \\
\bottomrule
\end{tabular}
}
\end{table}

\section*{Appendix D. Visualization}
\subsubsection{Visualization of AUC-ROC curve of Confidence.} 
\begin{figure}[t]
\centering
\includegraphics[width=0.8\linewidth]{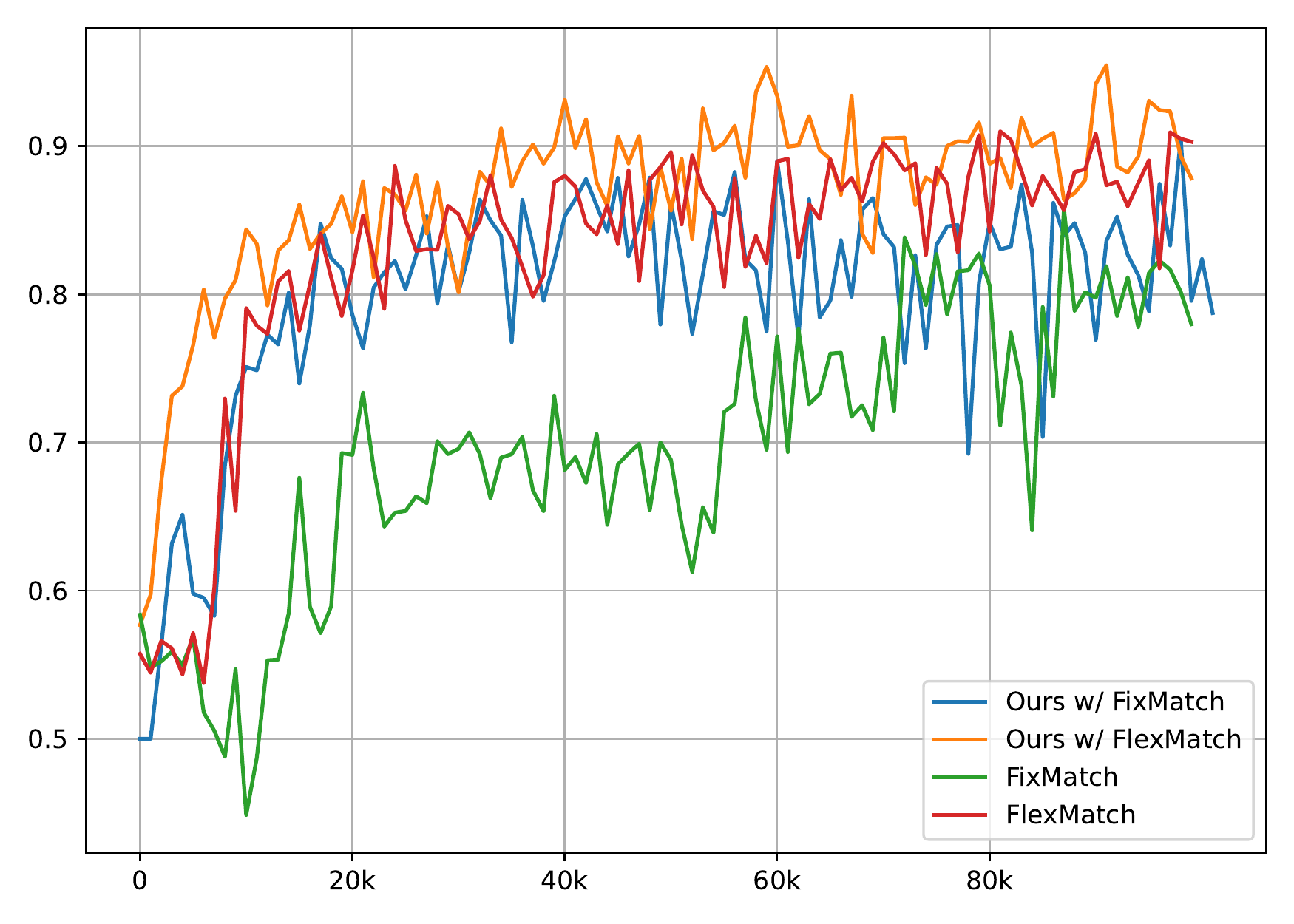}\\
\vspace{-5pt}
\caption{\textbf{AUC-ROC Curve of ConMatch with various baselines} The orange line means FlexMatch w/ ConMatch, the blue line means FixMatch w/ ConMatch, the red line means FlexMatch, and the green line means original FixMatch, respectively.}
\label{fig:roc-auc}\vspace{-5pt}
\end{figure}

The area under the ROC curve (AUC) is used for measuring the accuracy of a classifier and the better confidence prediction is closer to 1. On confidence estimation, it is also used to evaluate the capability of the confidence measure to distinguish correct class assignments from erroneous ones. As shown in Fig.~\ref{fig:roc-auc}, for overall training iterations, ConMatch w/~\cite{sohn2020fixmatch} and ConMatch w/~\cite{zhang2021flexmatch} show the higher AUC-ROC value compared to each baseline, FixMatch~\cite{sohn2020fixmatch}, and FlexMatch~\cite{zhang2021flexmatch}. This explains that the confidence estimator measures the confidences of augmented instances more properly than FixMatch and FlexMatch, which use max class probability as the confidence measure. This will be also proven in the following section. \vspace{-10pt}
 
\begin{table}[t]
\caption{
\textbf{Ablation study on similarity functions in non-parametric approach.}The best results are in bold.}
\label{table:ablation_simfunc}\vspace{-5pt}
\centering
\resizebox{0.6\textwidth}{!}{
\begin{tabular}{c|c|cc}
\toprule
\multirow{2}{*}{   Similarity Function   } & \multirow{2}{*}{   Target   } & \multicolumn{2}{c}{\textbf{  CIFAR-10  }}    \\
% &   \multicolumn{3}{c}{\# Labeled samples} & \multicolumn{3}{c}{\# Labeled samples} \\
% \cline{3-5} \cline{6-8} \cline{9-12} 
\cline{3-4}
 &  & \multicolumn{2}{c}{40}   \\
\midrule
 Cross Entropy  & logit &   \multicolumn{2}{c}{\textbf{4.89}} \\
 L2 Norm  & logit &   \multicolumn{2}{c}{4.93} \\
 Cosine Similarity  & feature &   \multicolumn{2}{c}{5.02} \\
\bottomrule
\end{tabular}
}
\end{table}

\subsubsection{Visualization on Transition of Confidence.}
To explain the effectiveness of ConMatch, we show the feature distribution of ConMatch w/FixMatch~\cite{sohn2020fixmatch}, ConMatch w/FlexMatch~\cite{zhang2021flexmatch}, compared to~\cite{sohn2020fixmatch} and~\cite{zhang2021flexmatch} at different training iterations in Table~\ref{table:tsne-confix}, Table~\ref{table:tsne-fix}, Table~\ref{table:tsne-conflex} and Table~\ref{table:tsne-flex}, respectively. For visualizing the high dimensional features, the complex feature matrices are transformed into two-dimensional points based on t-SNE~\cite{van2008visualizing}. As training progresses, we can confirm that t-SNE visualization of labeled and weakly-augmented samples are clustered into distinct groups, relatively faster than baseline~\cite{sohn2020fixmatch} by comparing Table~\ref{table:tsne-confix} and Table~\ref{table:tsne-fix}. And we can also find that confident strongly-augmented samples are also well-organized near confident weakly-augmented samples. 

The color of dots means the confidence of each sample. The darkest color dot means the high-confidence instances, and the empty dot means the low- or zero-confidence instances, whose confidence values are relatively lower. For ConMatch, we consider the confidence estimator output as the confidence of each instance, and for the FixMatch~\cite{sohn2020fixmatch} and FlexMatch~\cite{zhang2021flexmatch}, we consider the maximum probability of the output distribution as the confidence of each instance. 
Both methods, ConMatch and FixMatch / FlexMatch show that the high-confident strongly-augmented instances are closer to high-confident weakly-augmented instances than the low-confident strongly-augmented instances. About FixMatch and FlexMatch, they measure the confidence of the insufficiently clustered strongly augmented samples as relatively high around $0.5$, expressed as the \textit{light-blue} color of figures. Otherwise, different to FixMatch and FlexMatch, ConMatch shows sharp changes in confidence values, distinguishing the high- and low-confidence instances definitely so the insufficiently clustered samples have low confidence values around zero, expressed as the \textit{empty} circle of figures, even from the early stage of training. By this visualization, we can confirm that our confidence estimator effectively measures the confidence of samples and makes possible the training between the strongly augmented images. 

\begin{table*}[]
\begin{center}
\scalebox{0.8}{
\begin{tabular}{c|c|c|c} 

\textbf{Con+Fix} & Labeled samples  &  Weakly augmented samples &  Strongly augmented samples\\ \hline
10000 iter & {\includegraphics[width=0.33\linewidth]{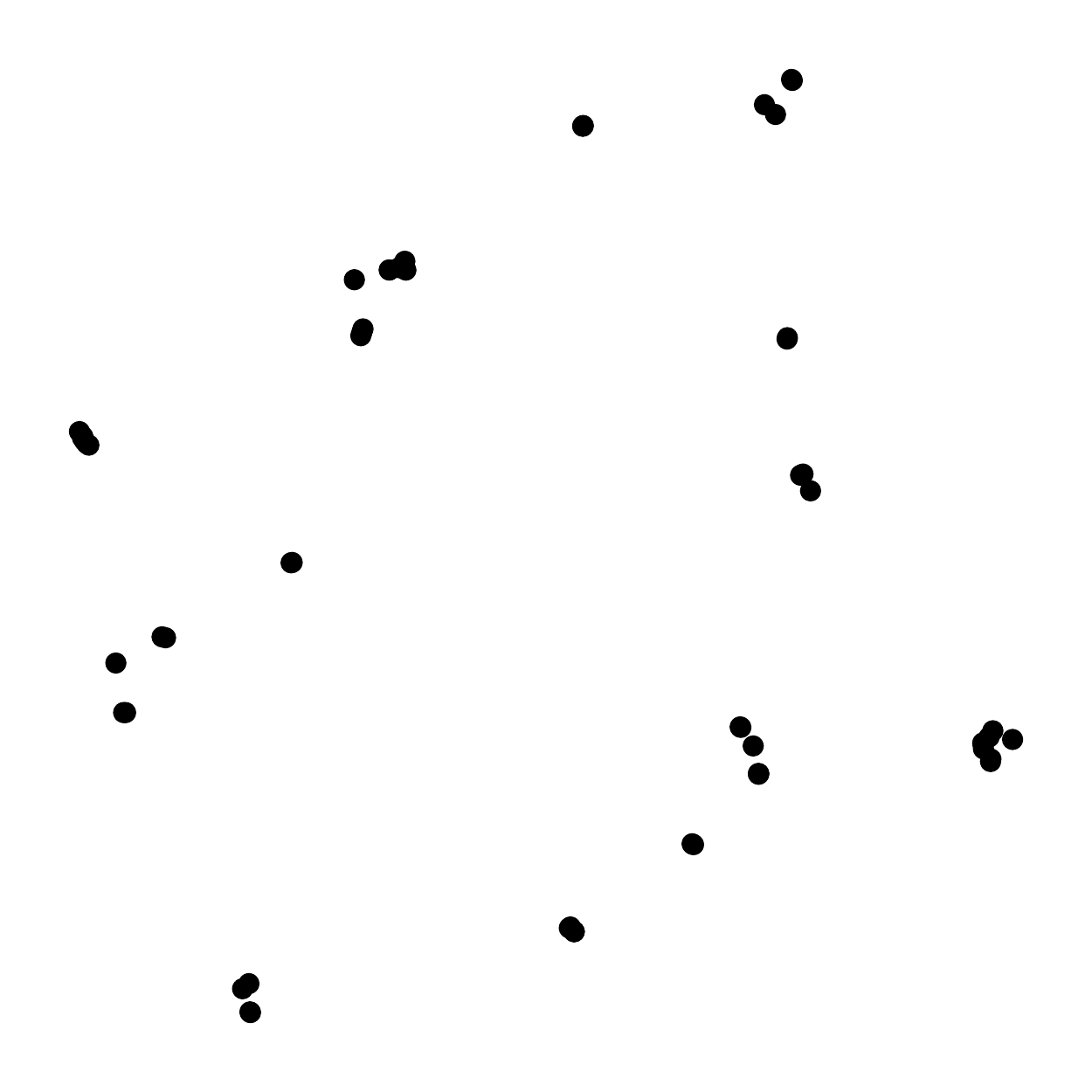}}\hfill &                     {\includegraphics[width=0.33\linewidth]{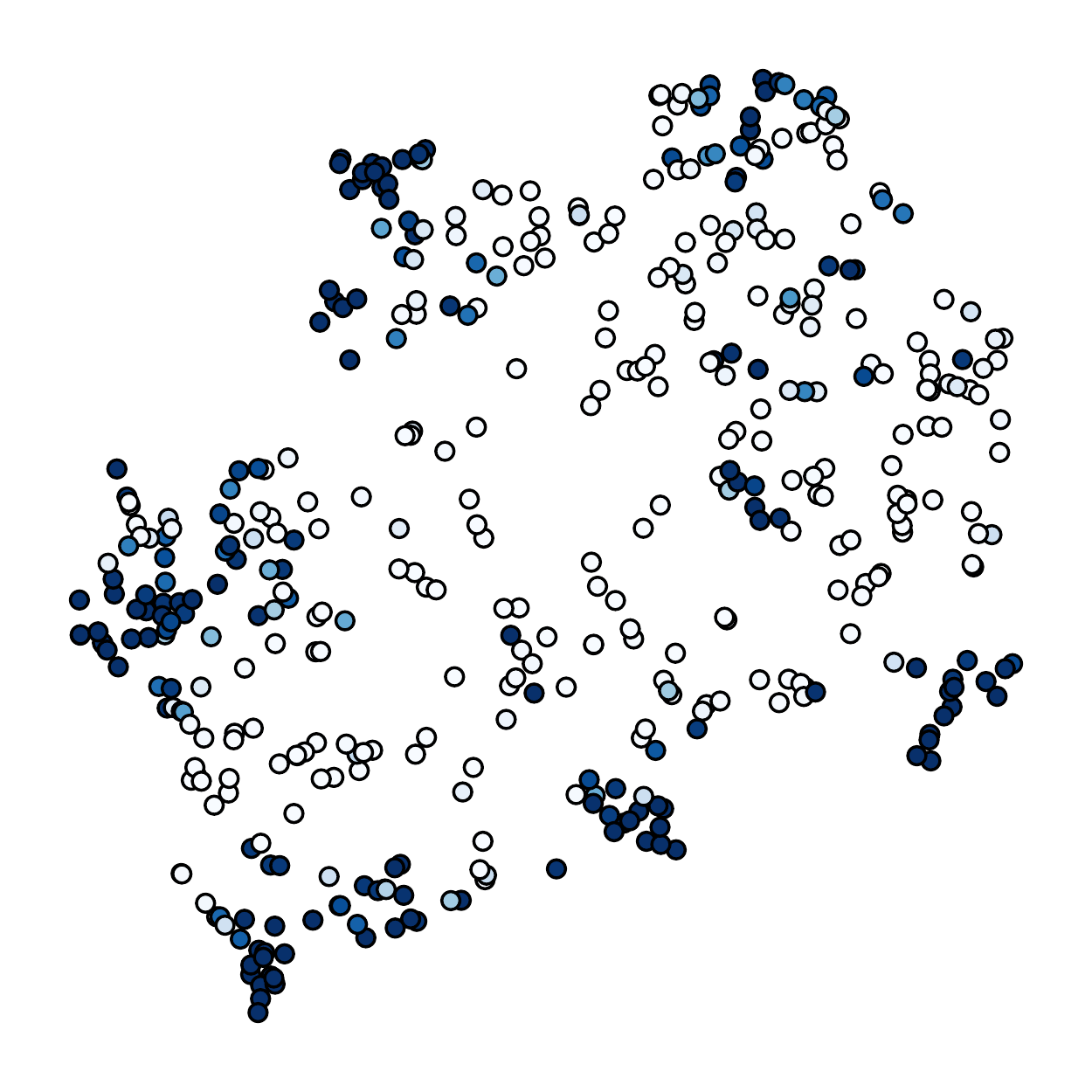}}\hfill & {\includegraphics[width=0.33\linewidth]{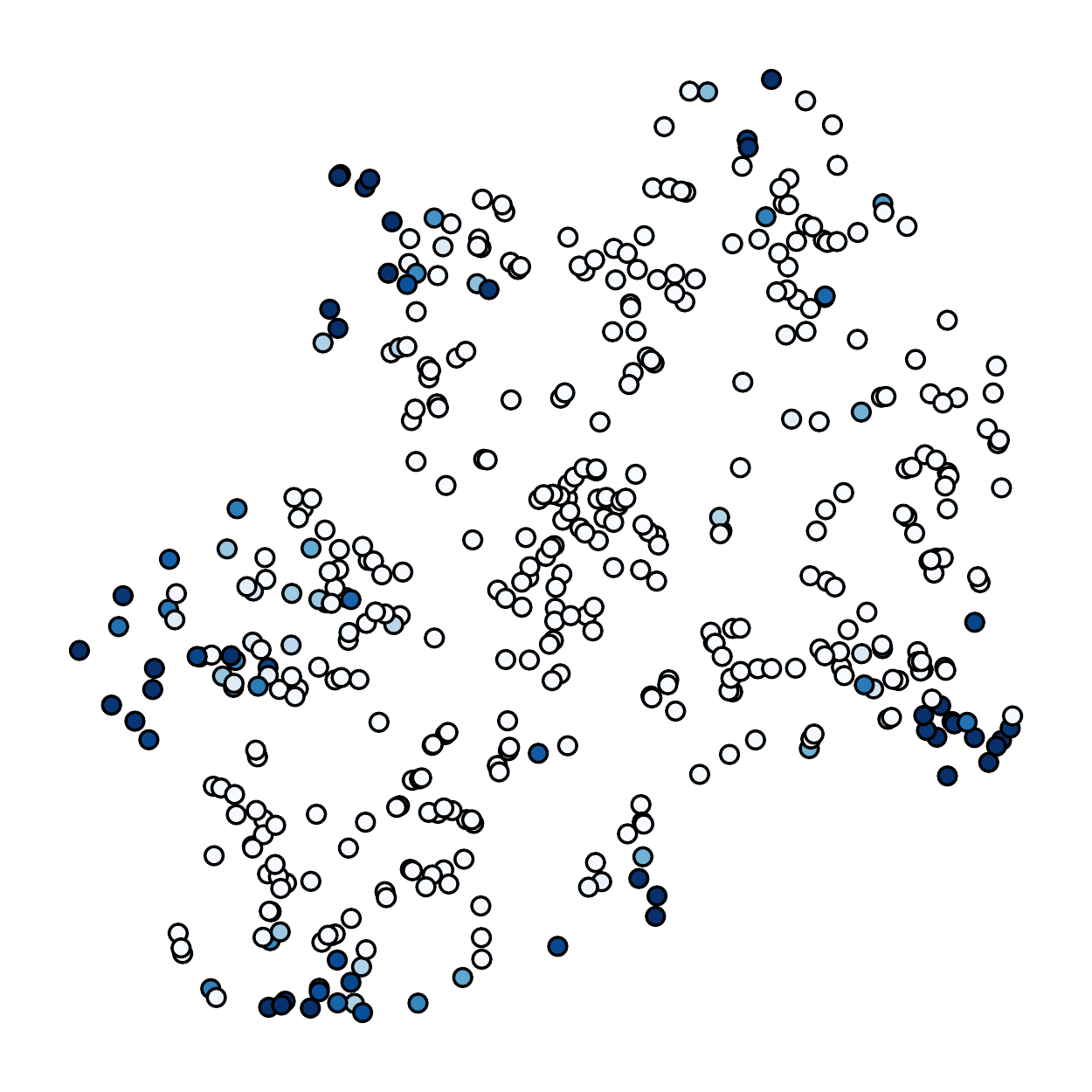}}\hfill  \\ \hline
30000 iter & {\includegraphics[width=0.33\linewidth]{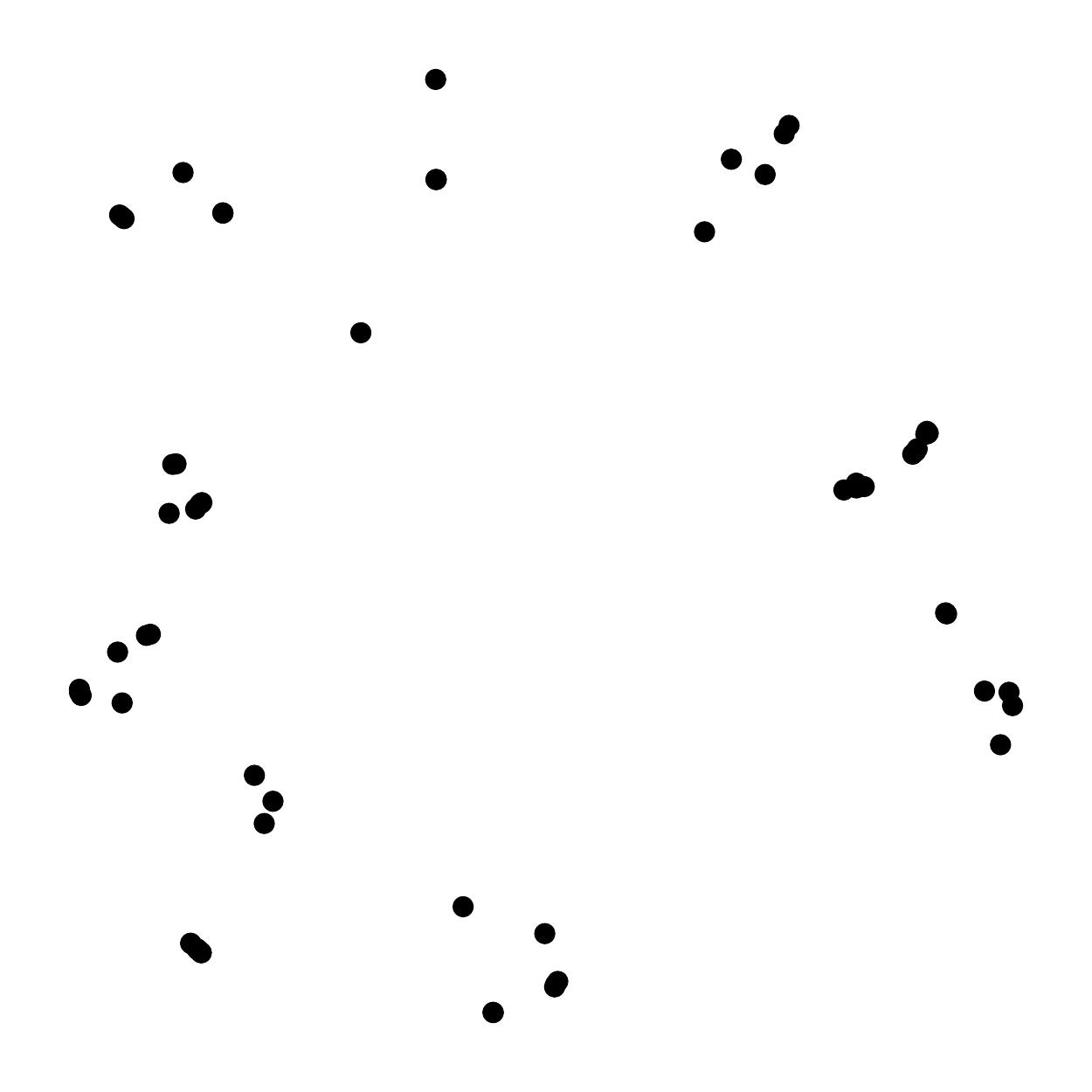}}\hfill &                     {\includegraphics[width=0.33\linewidth]{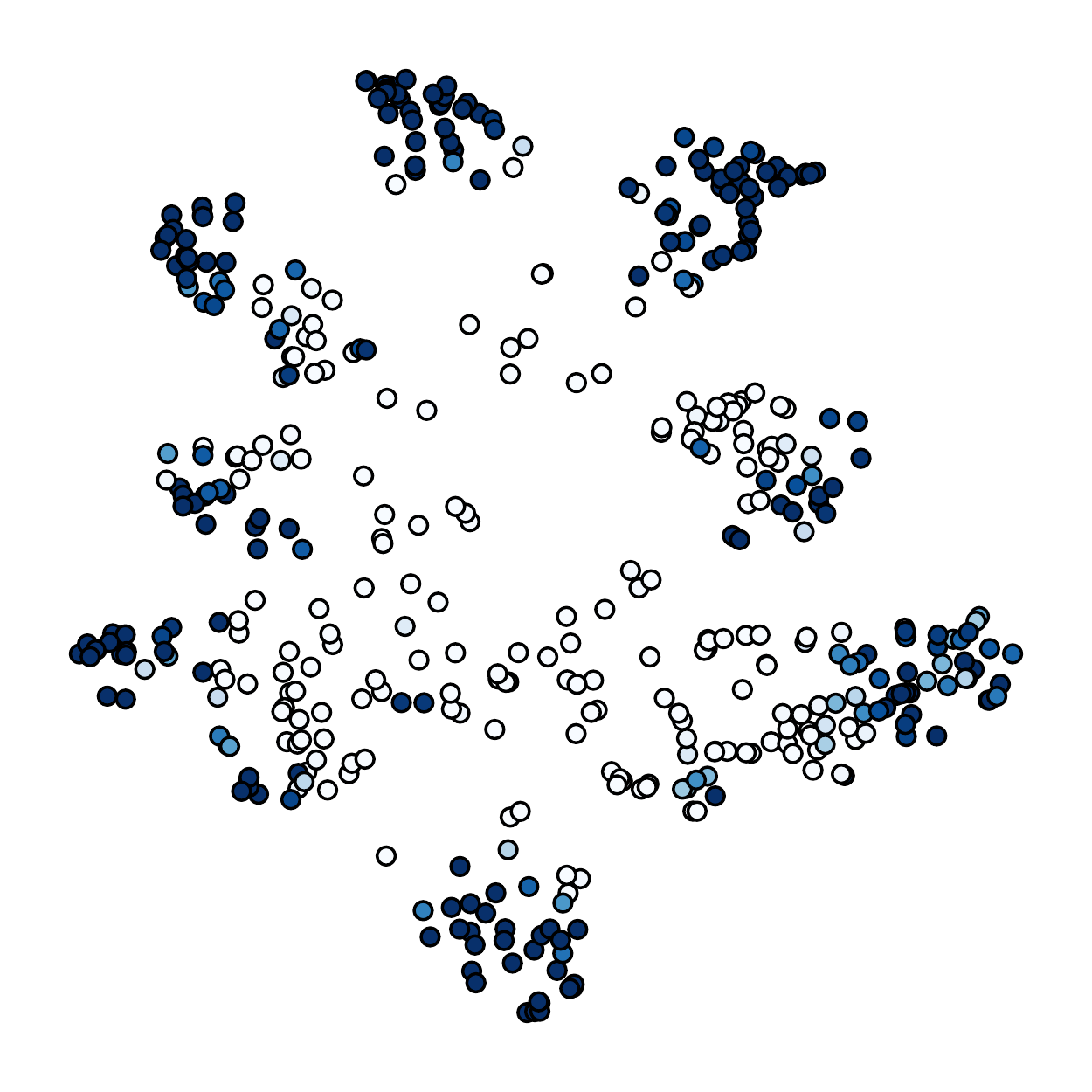}}\hfill & {\includegraphics[width=0.33\linewidth]{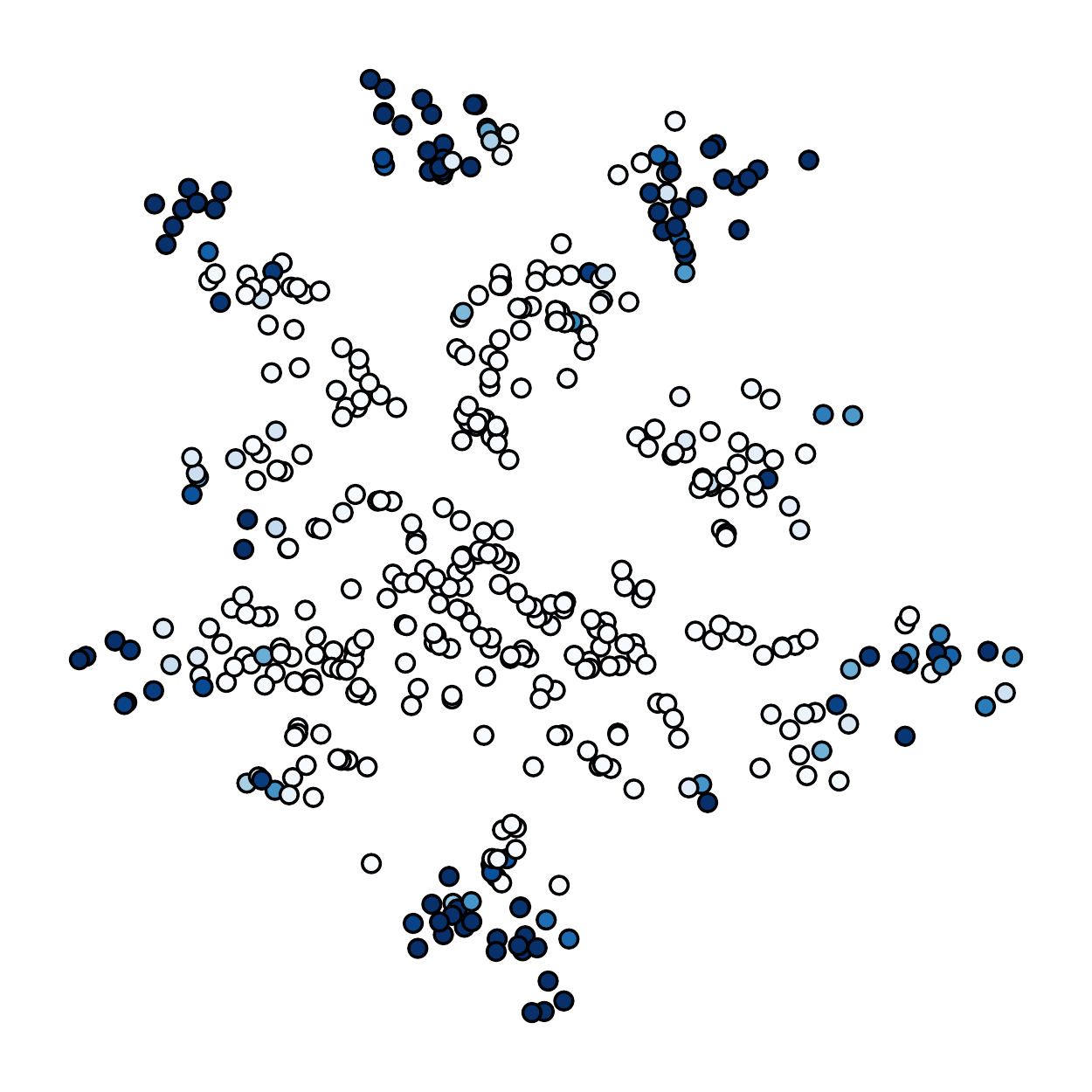}}\hfill  \\ \hline
50000 iter & {\includegraphics[width=0.33\linewidth]{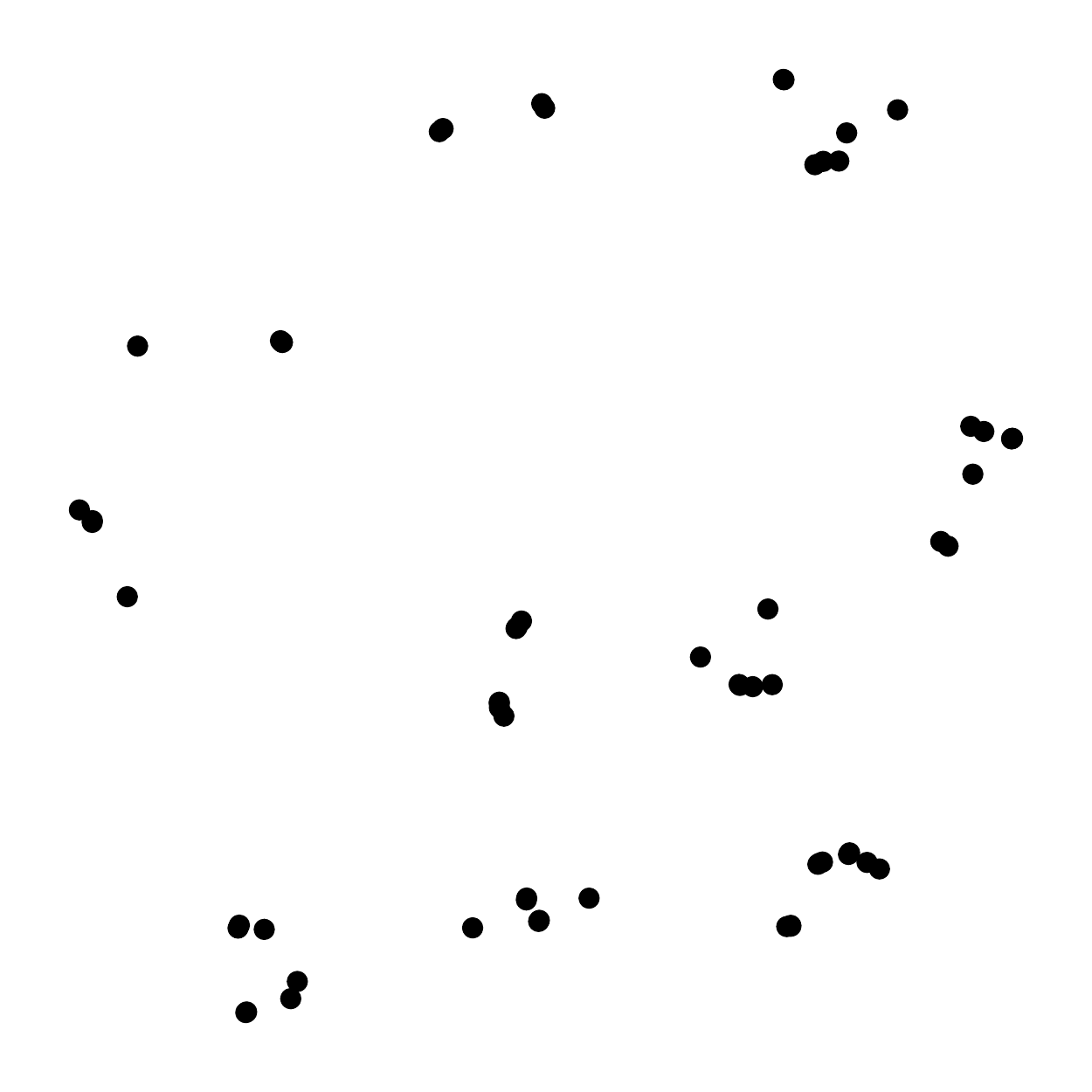}}\hfill &                     {\includegraphics[width=0.33\linewidth]{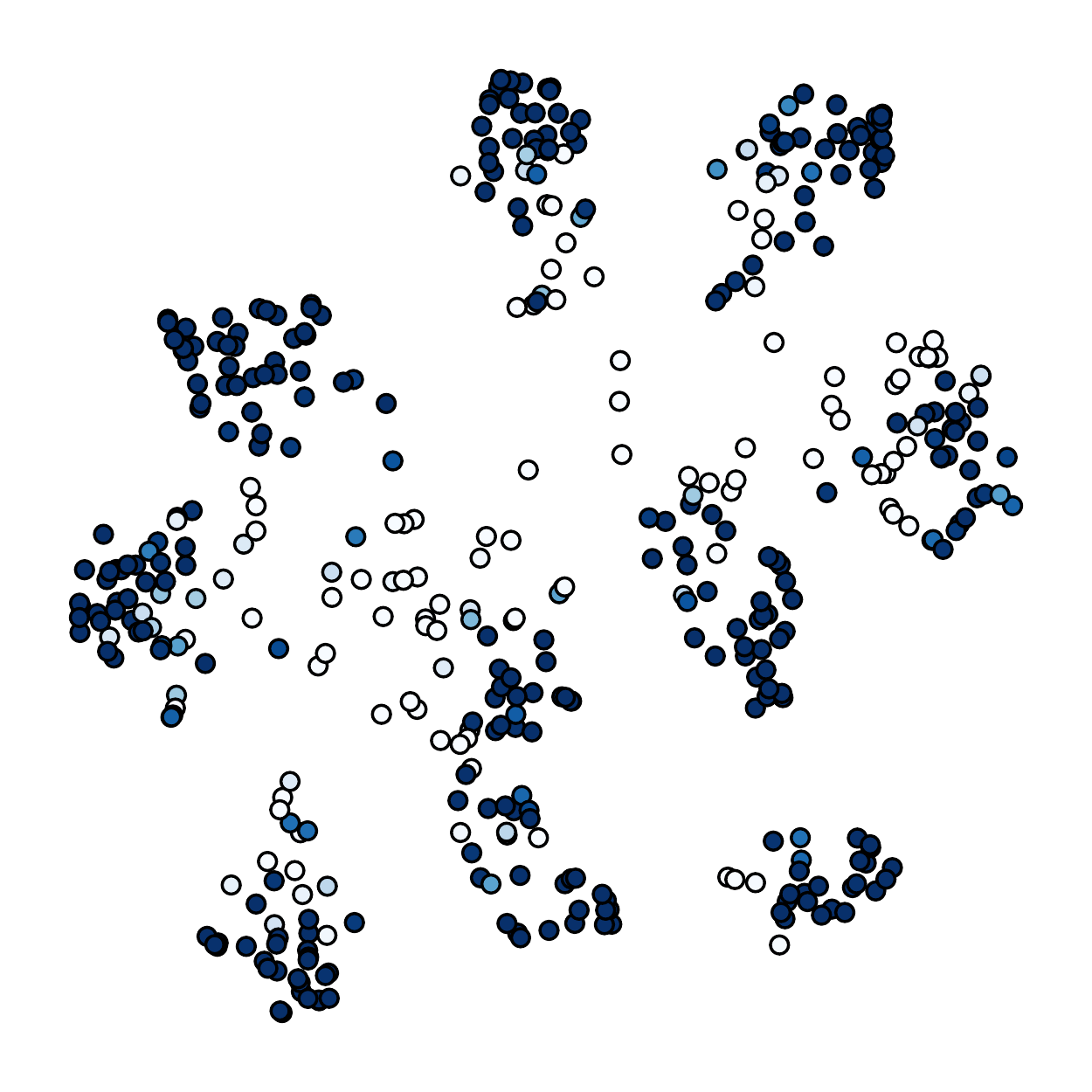}}\hfill & {\includegraphics[width=0.33\linewidth]{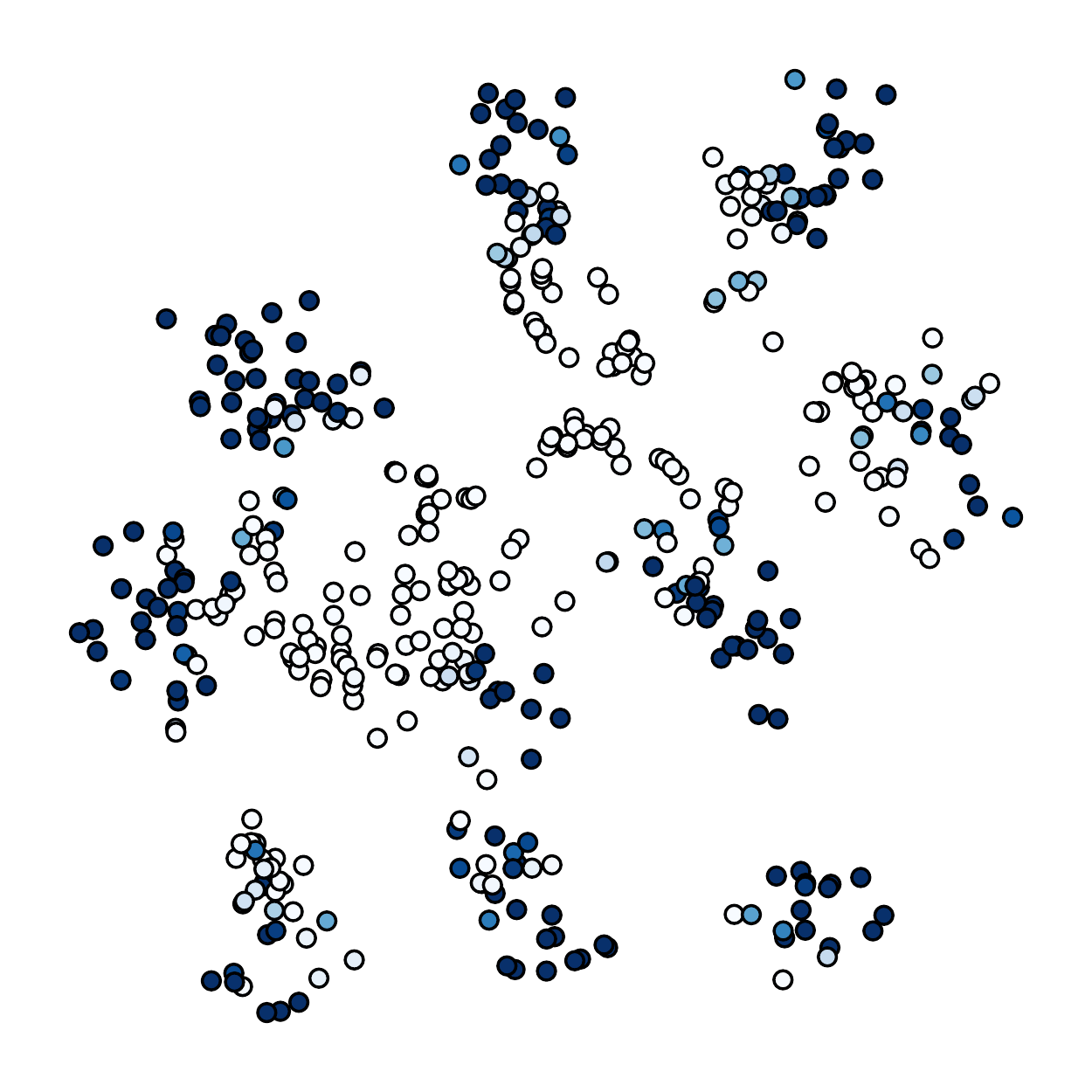}}\hfill  \\ \hline
70000 iter & {\includegraphics[width=0.33\linewidth]{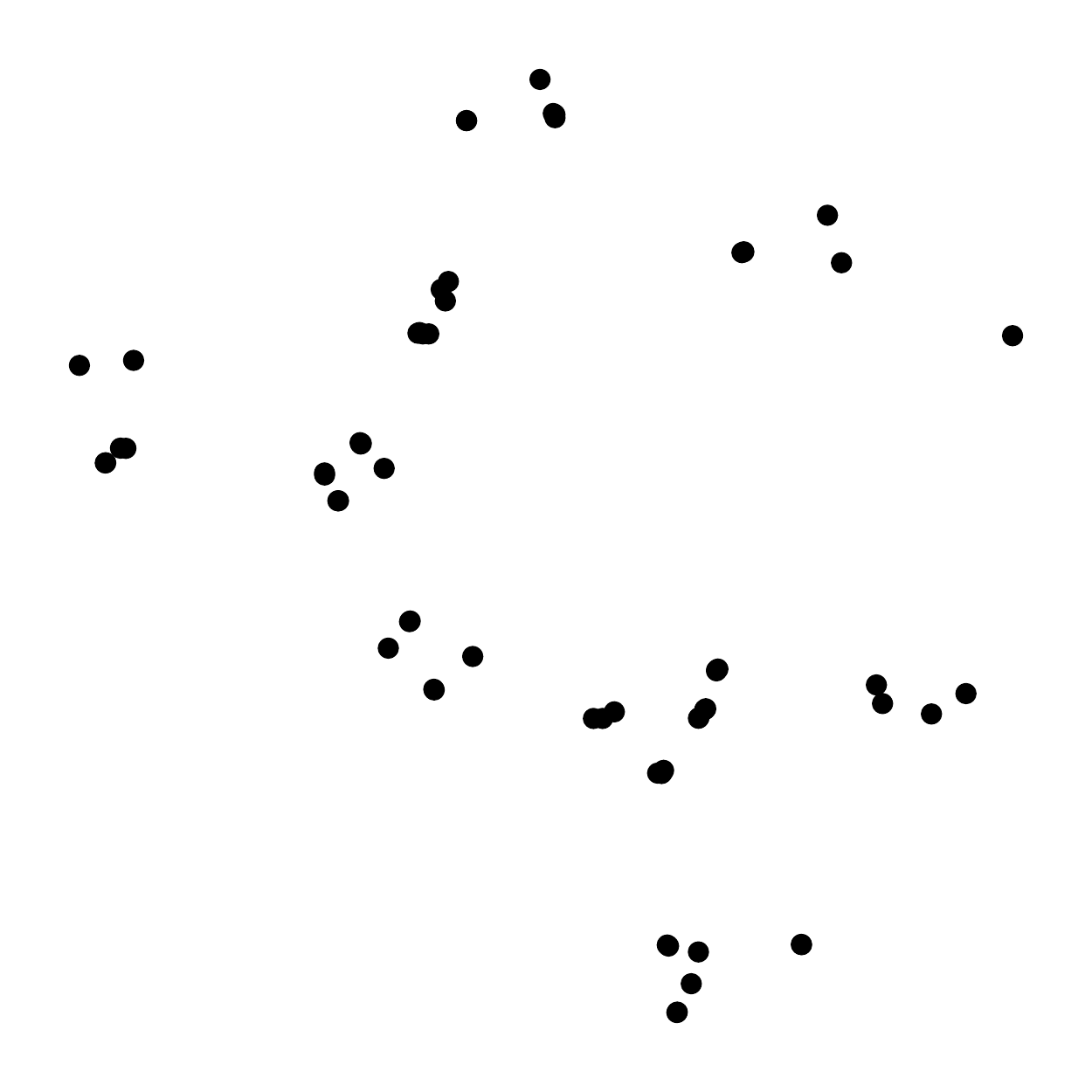}}\hfill &                     {\includegraphics[width=0.33\linewidth]{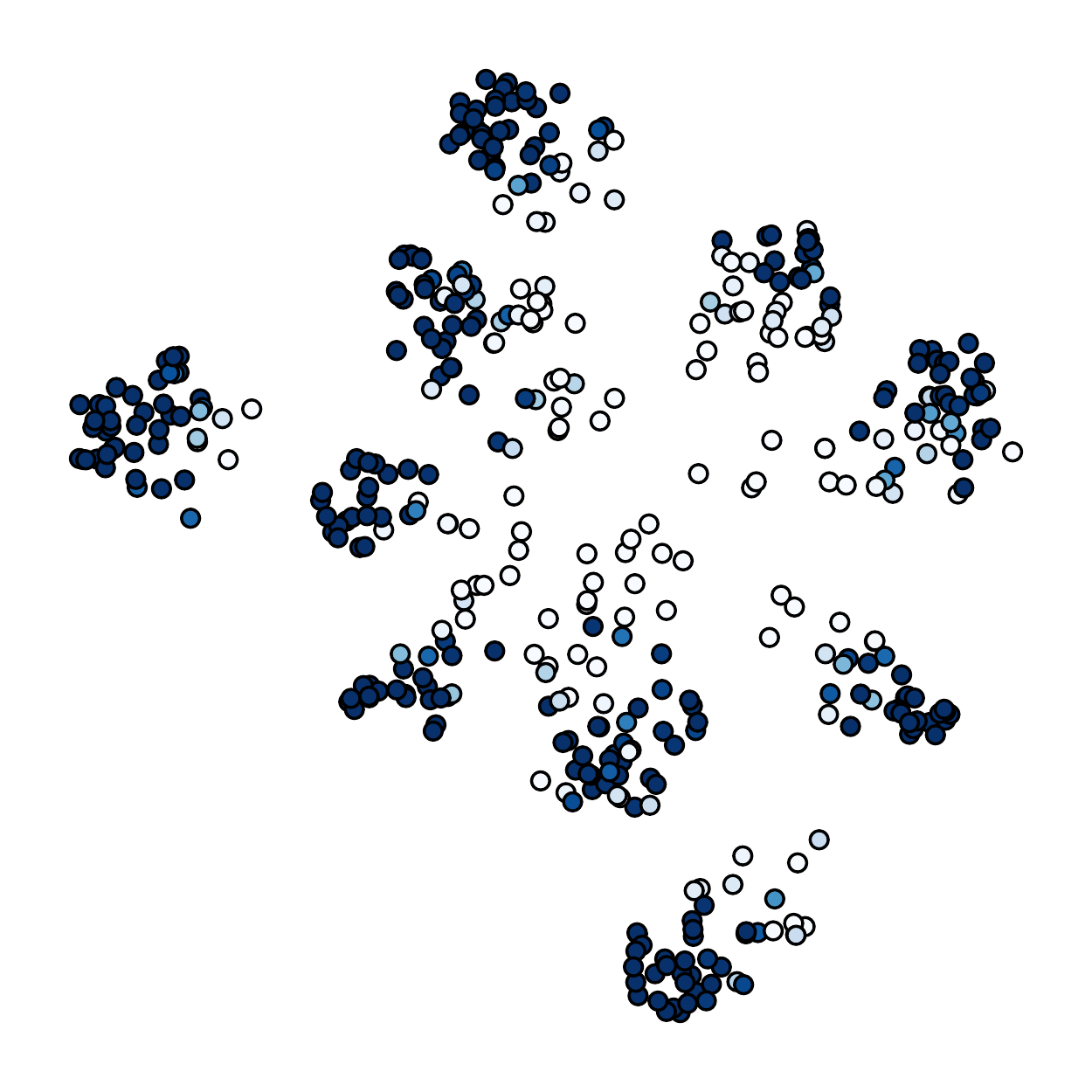}}\hfill & {\includegraphics[width=0.33\linewidth]{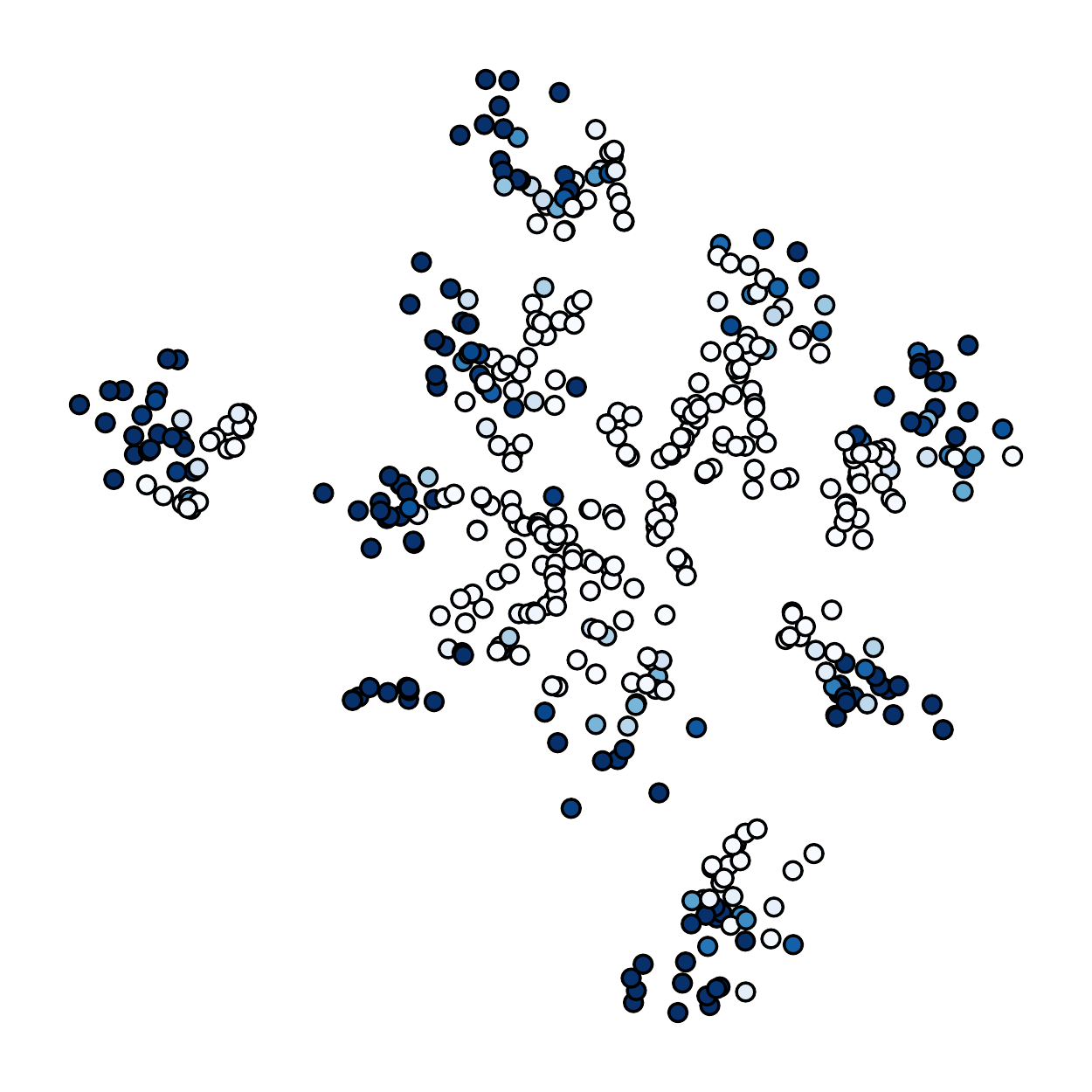}}\hfill  \\ \hline
90000 iter & {\includegraphics[width=0.33\linewidth]{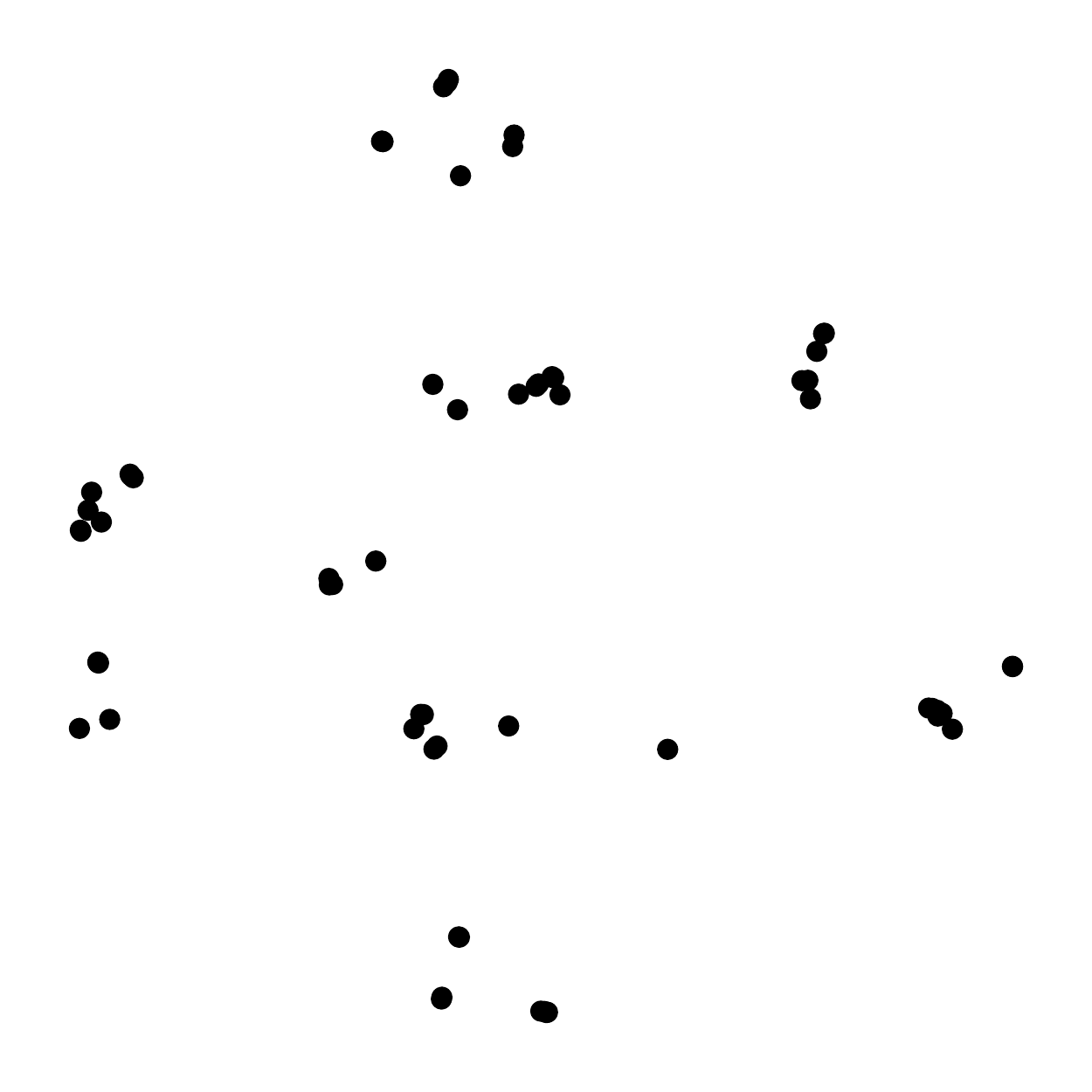}}\hfill &                     {\includegraphics[width=0.33\linewidth]{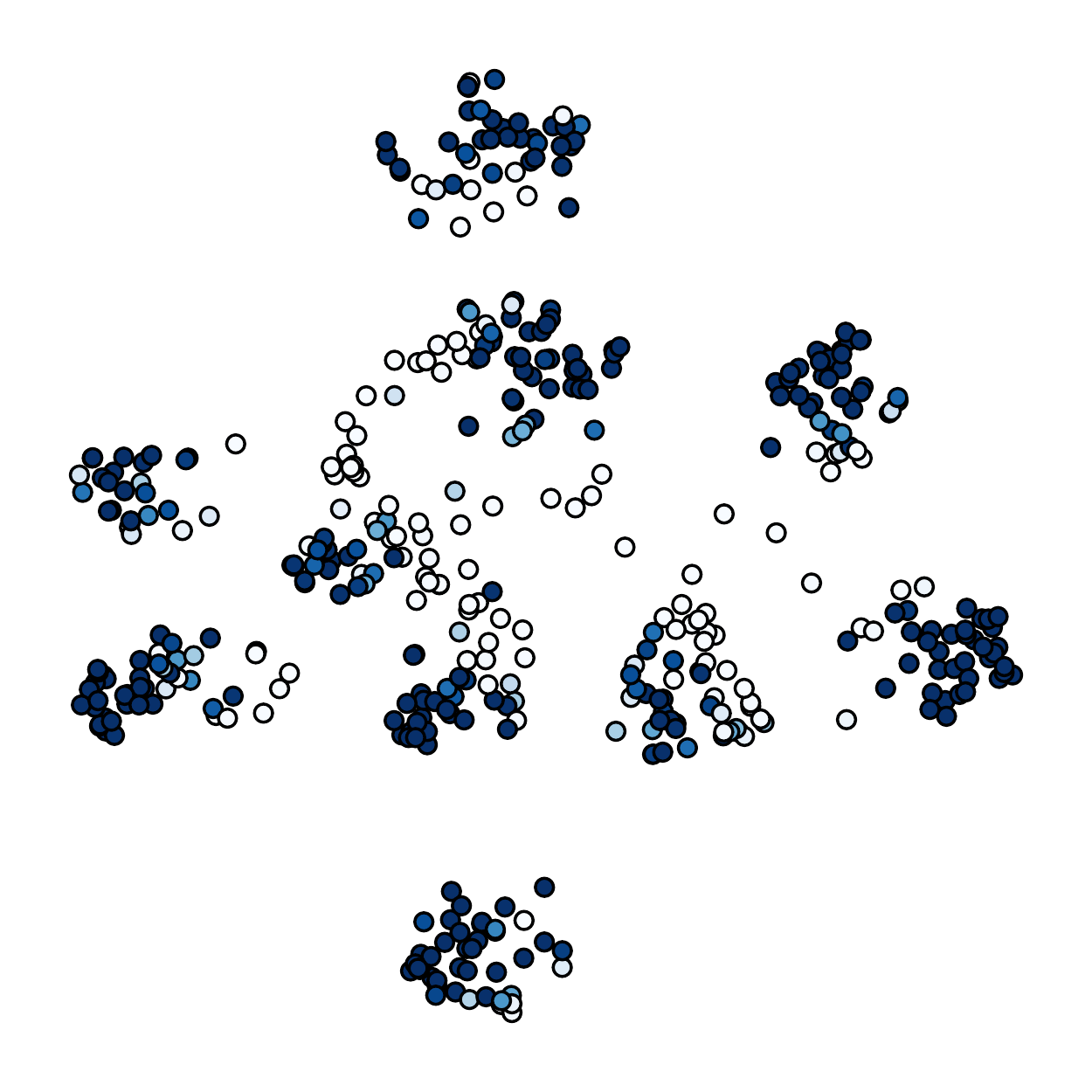}}\hfill & {\includegraphics[width=0.33\linewidth]{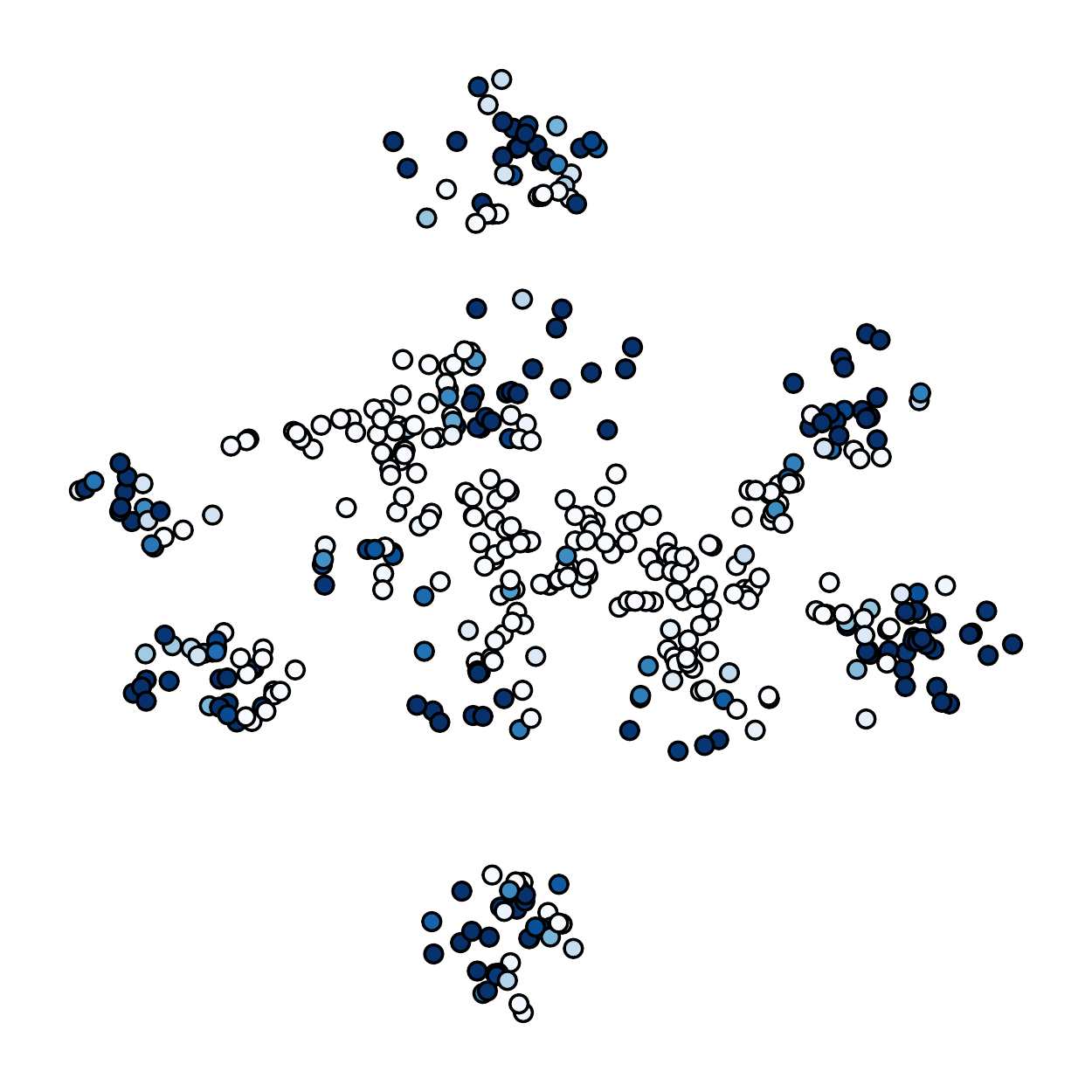}}\hfill  \\ \hline
\end{tabular}
}\vspace{-10pt}
\end{center}
\caption{Visualization on the transitions feature distribution with confidence in \textbf{ConMatch w/FixMatch} (denoted as blue dots in colors) of training samples from CIFAR-10 in the feature space at different training stages. Data projection in 2-D space is attained by t-SNE based on the feature representation. The dark blue dots represent the most confident samples, and the empty dots represent the least confident samples.}
\label{table:tsne-confix}
\end{table*}

\begin{table*}[]
\begin{center}
\scalebox{0.8}{
\begin{tabular}{c|c|c|c} 

\textbf{FixMatch} & Labeled samples  &  Weakly augmented samples &  Strongly augmented samples\\ \hline
10000 iter & {\includegraphics[width=0.33\linewidth]{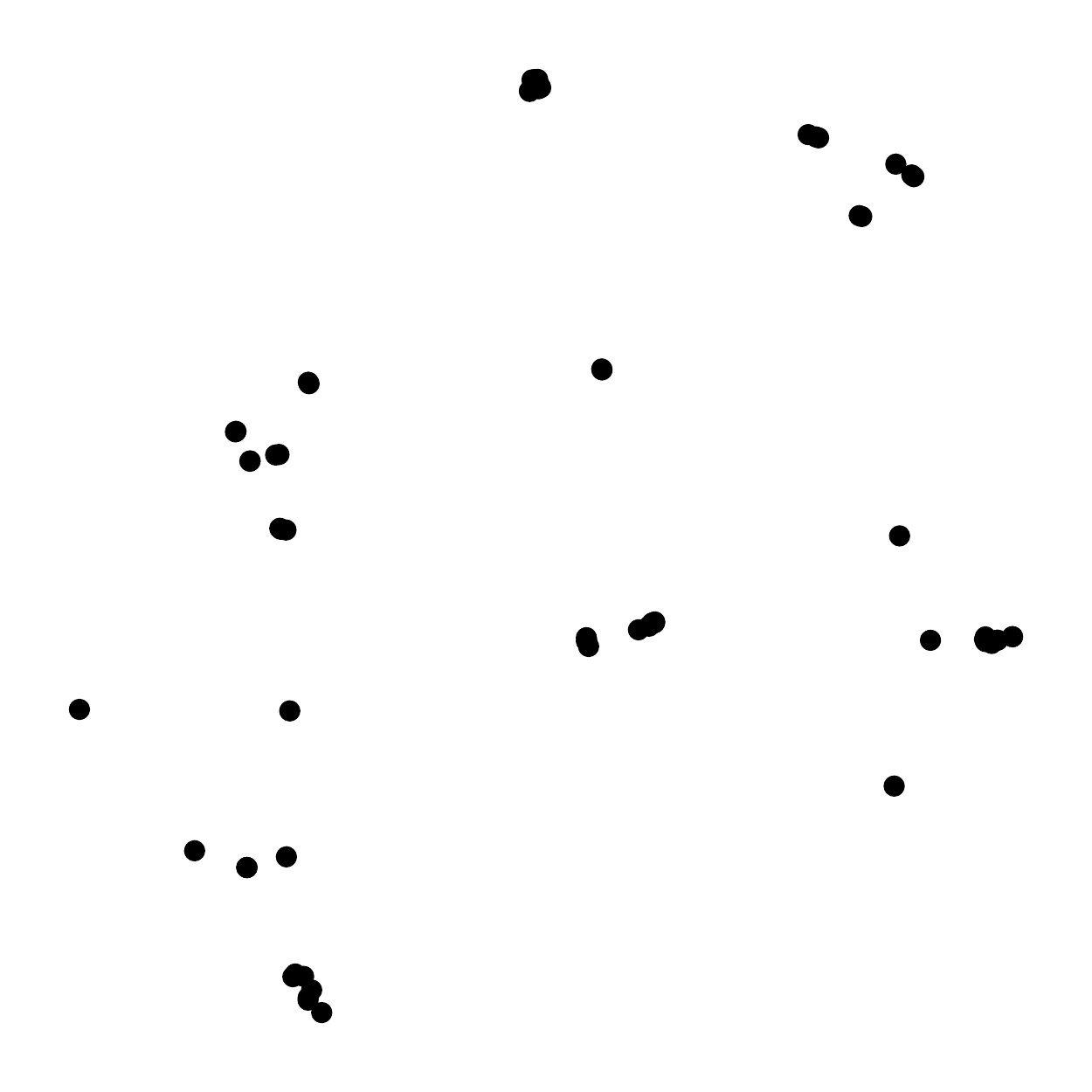}}\hfill &                     {\includegraphics[width=0.33\linewidth]{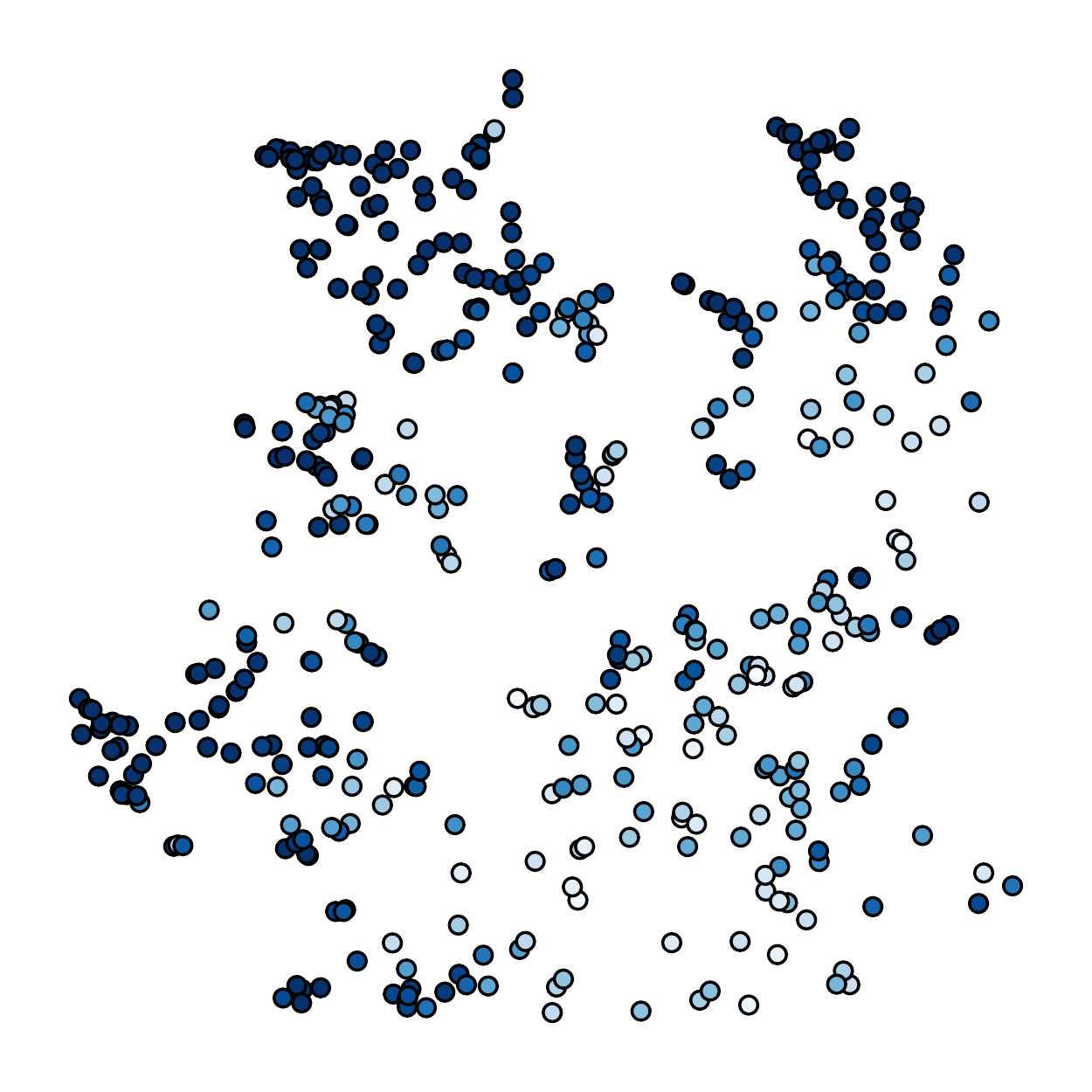}}\hfill & {\includegraphics[width=0.33\linewidth]{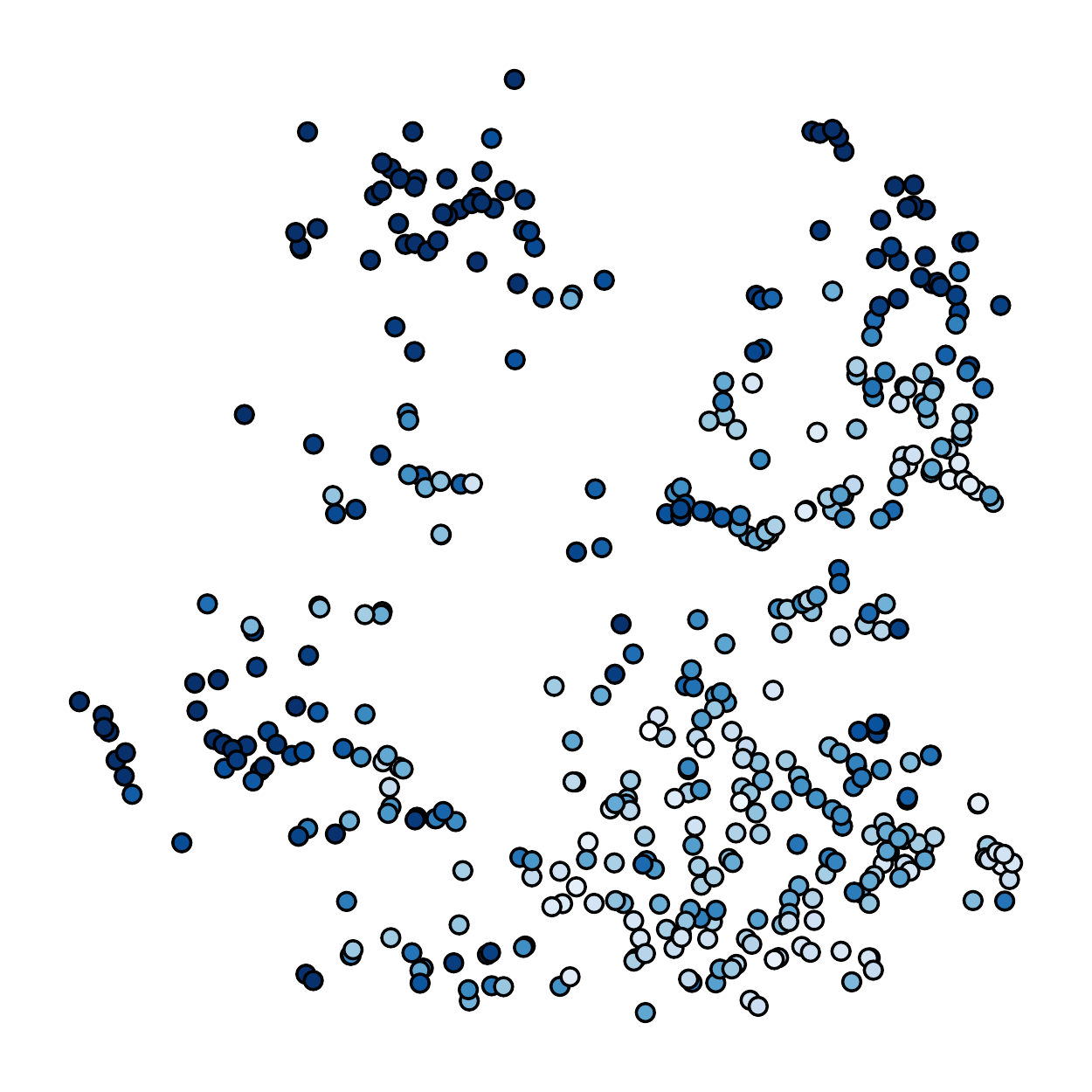}}\hfill  \\ \hline
30000 iter & {\includegraphics[width=0.33\linewidth]{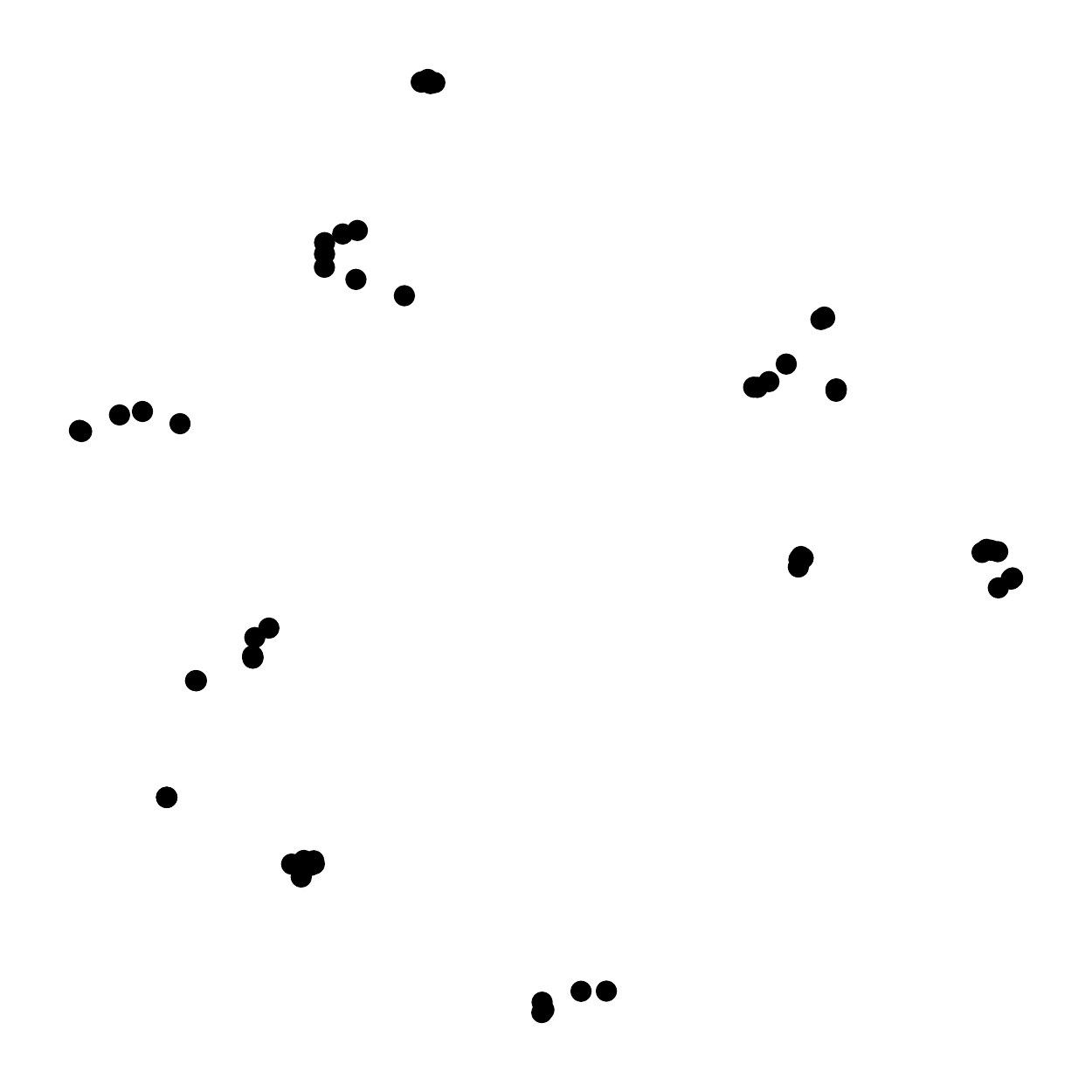}}\hfill &                     {\includegraphics[width=0.33\linewidth]{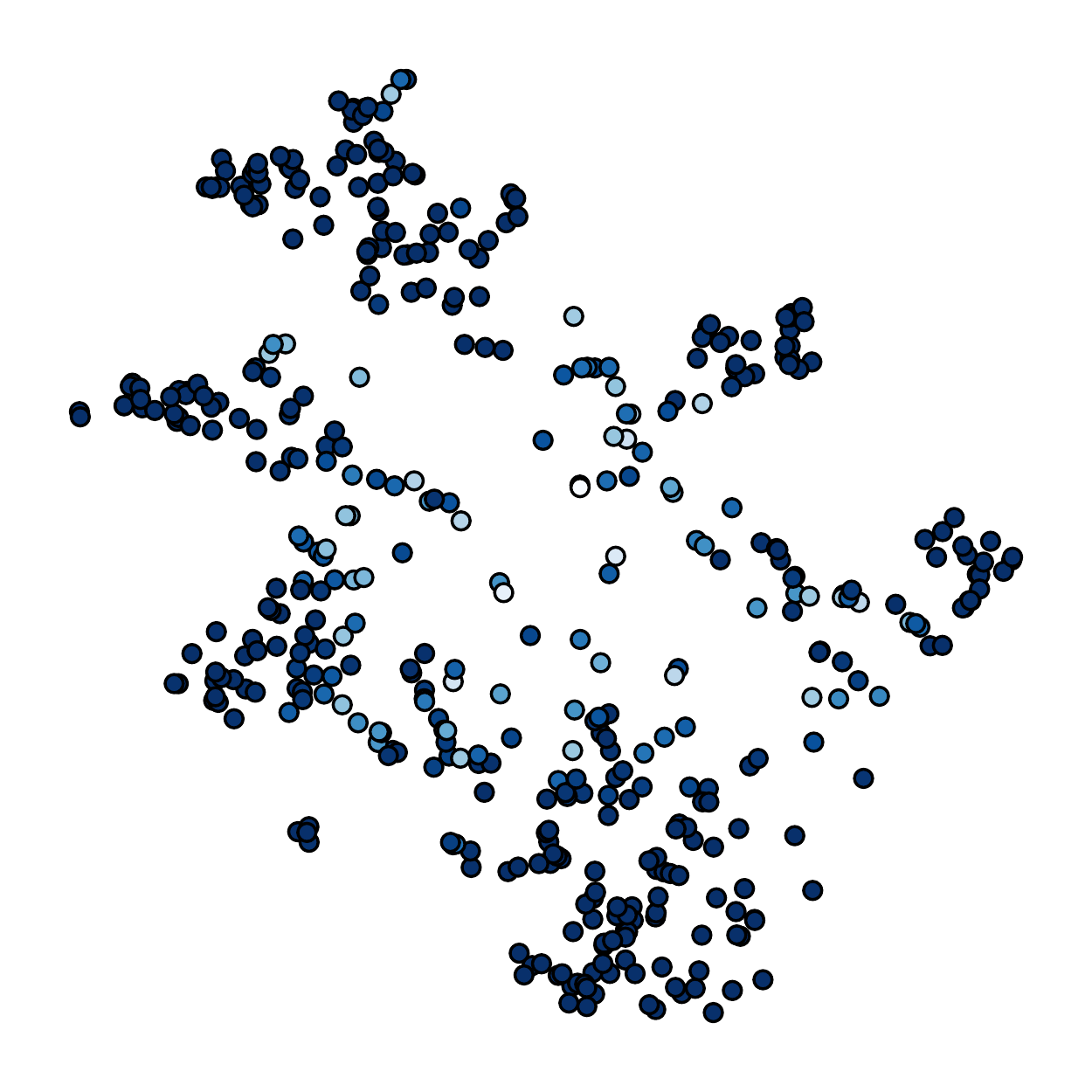}}\hfill & {\includegraphics[width=0.33\linewidth]{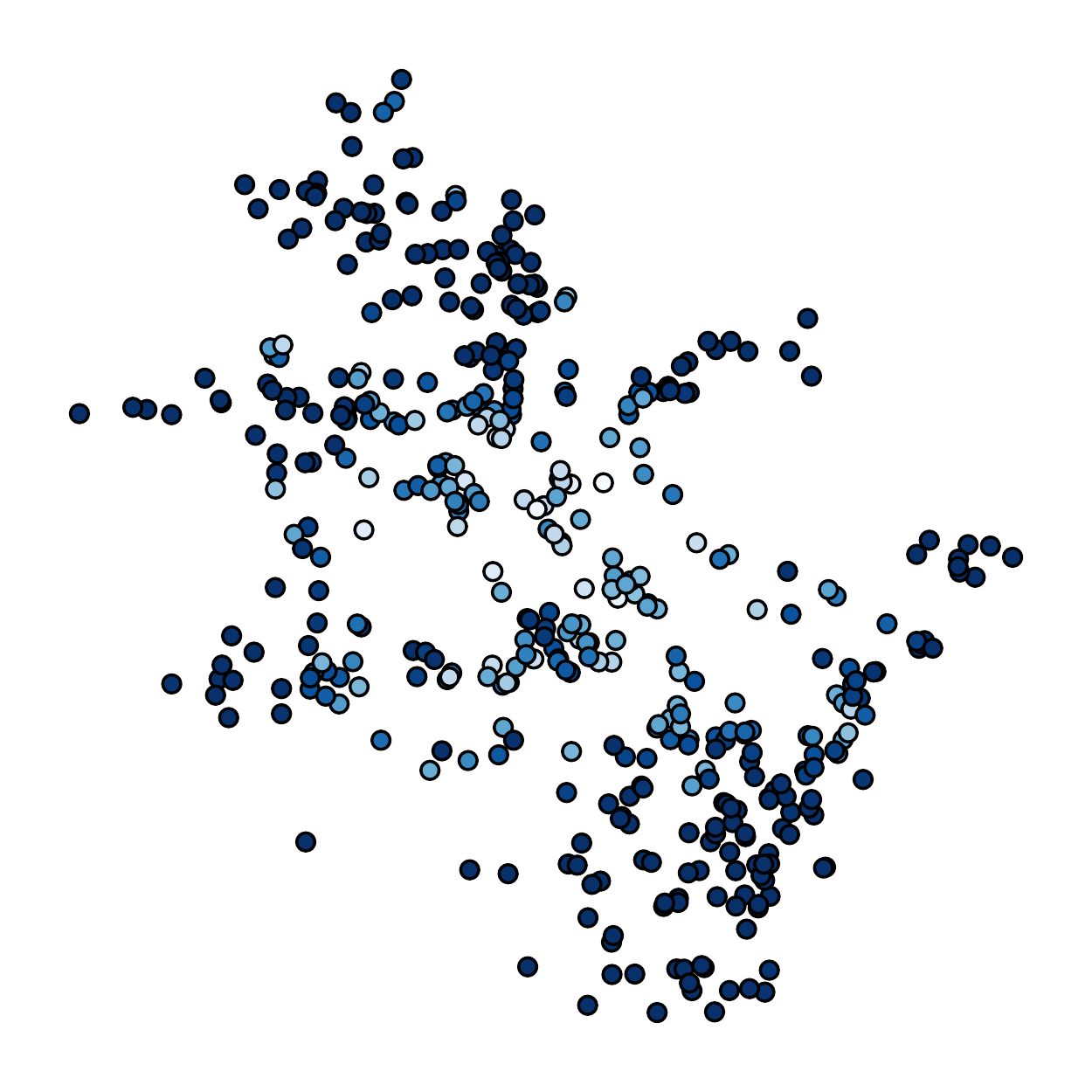}}\hfill  \\ \hline
50000 iter & {\includegraphics[width=0.33\linewidth]{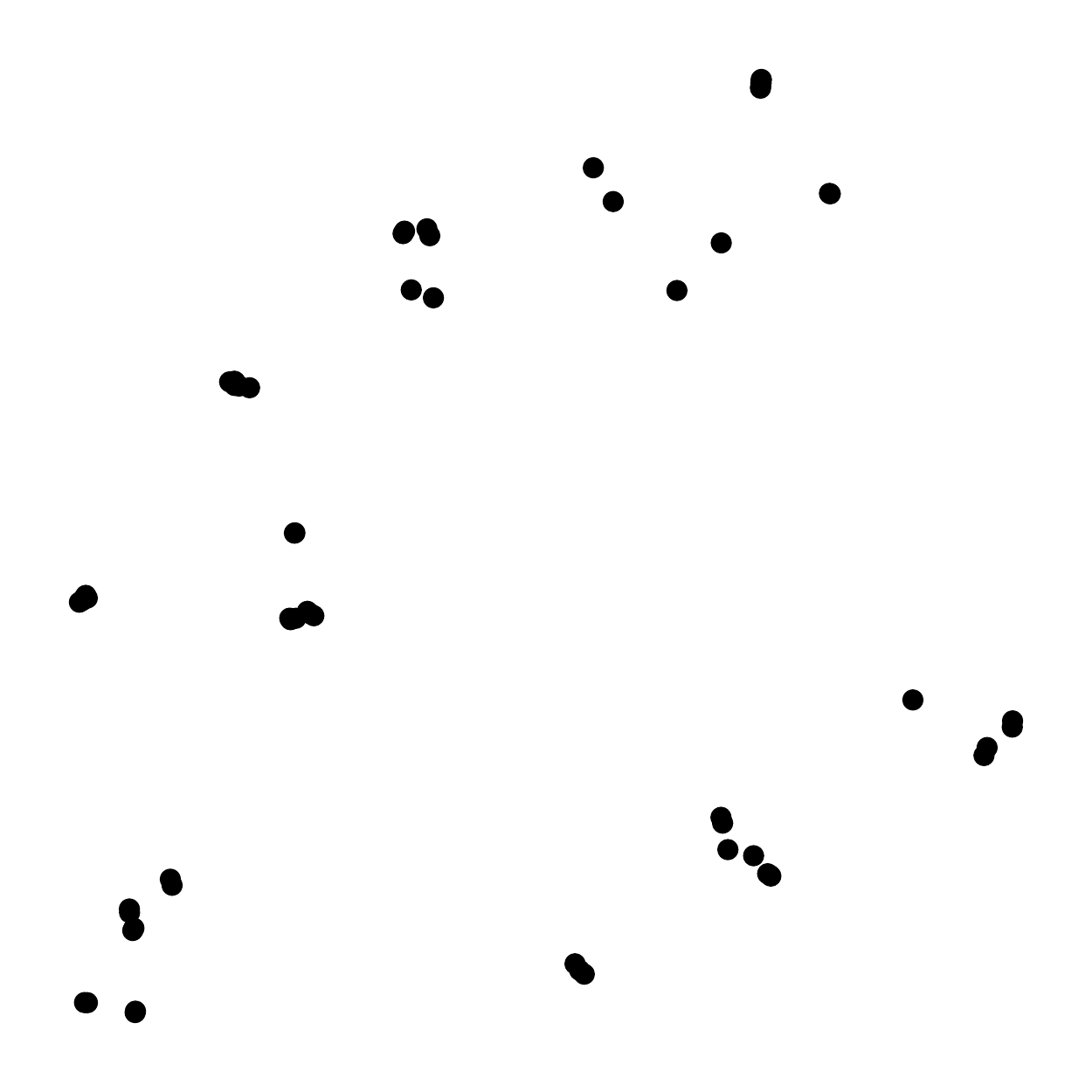}}\hfill &                     {\includegraphics[width=0.33\linewidth]{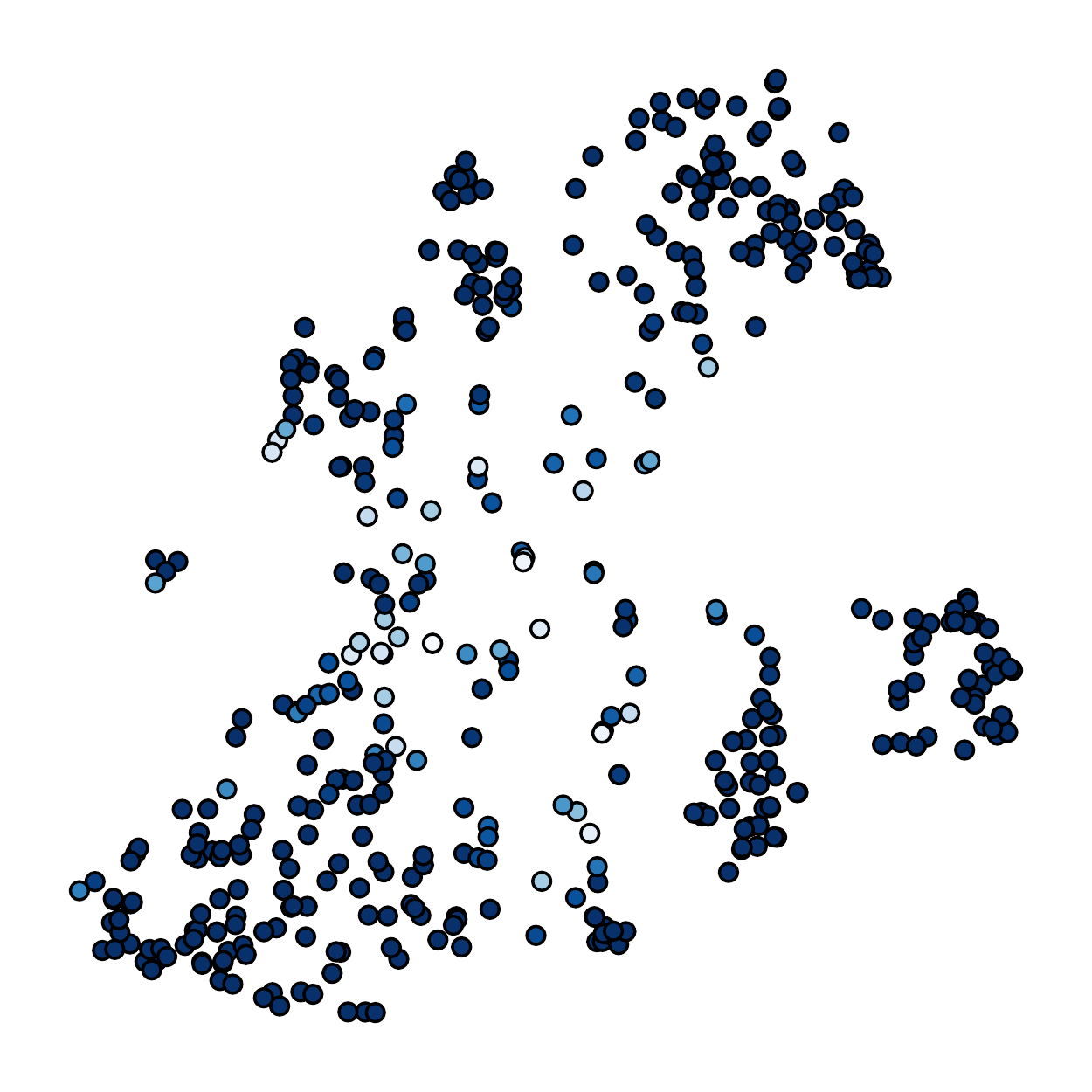}}\hfill & {\includegraphics[width=0.33\linewidth]{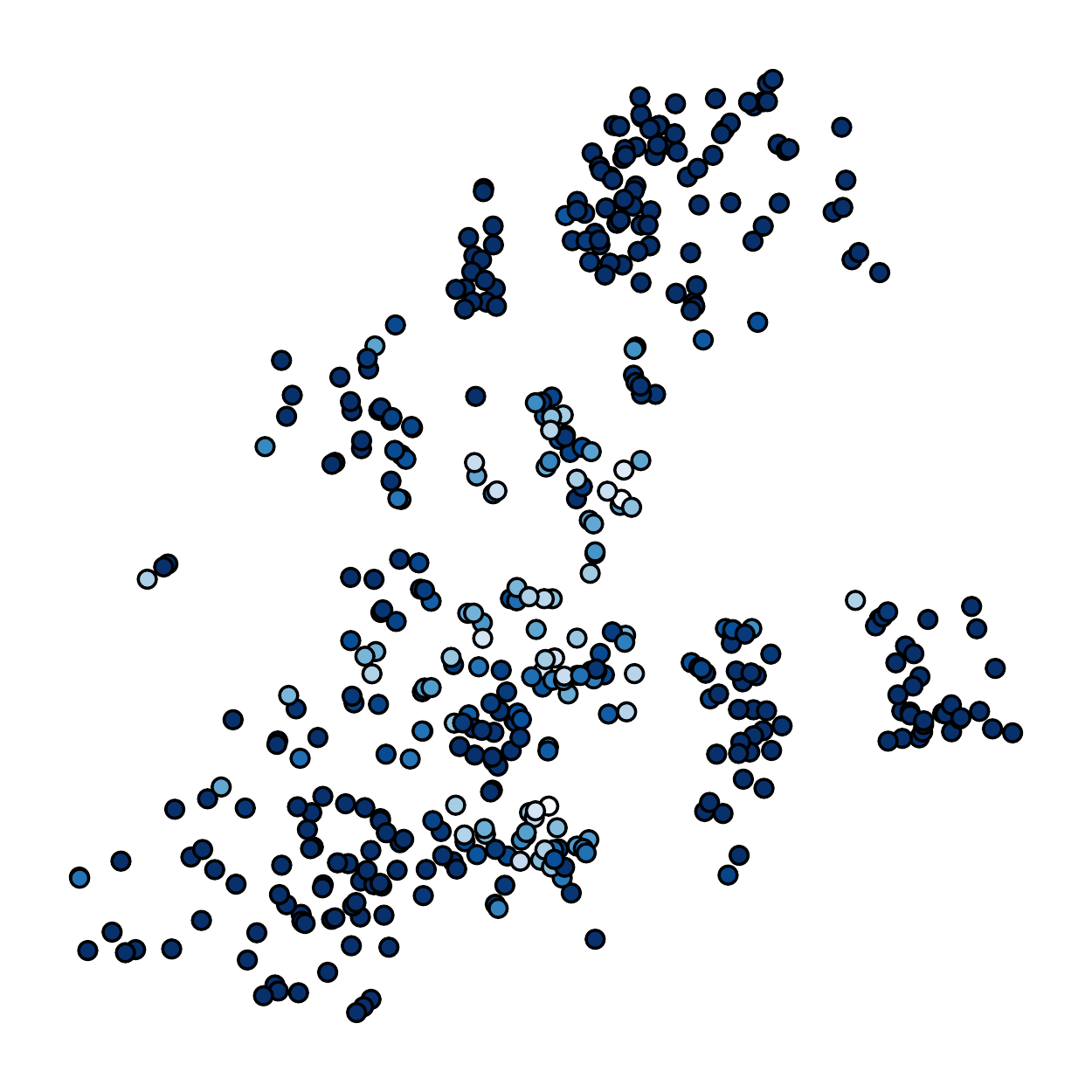}}\hfill  \\ \hline
70000 iter & {\includegraphics[width=0.33\linewidth]{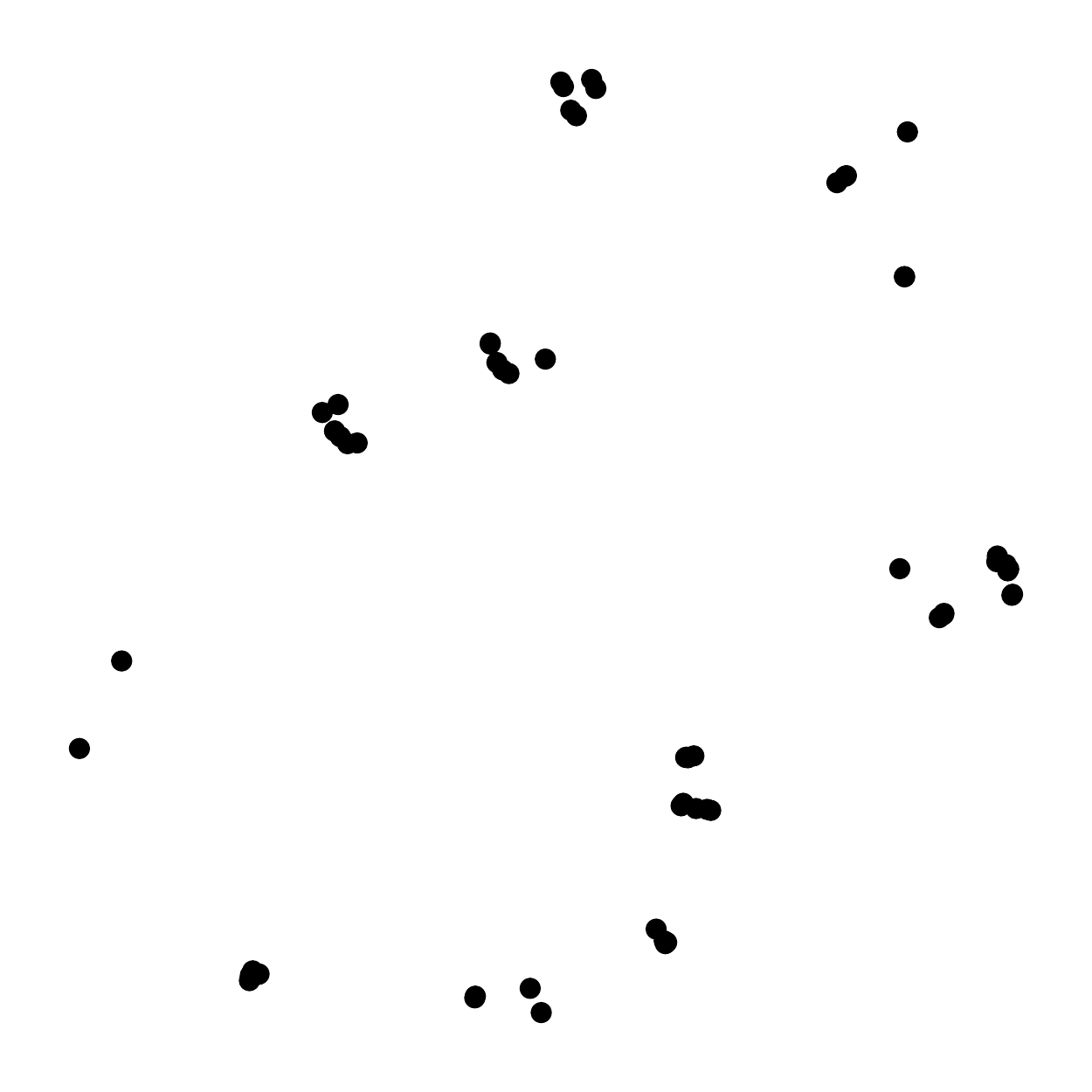}}\hfill &                     {\includegraphics[width=0.33\linewidth]{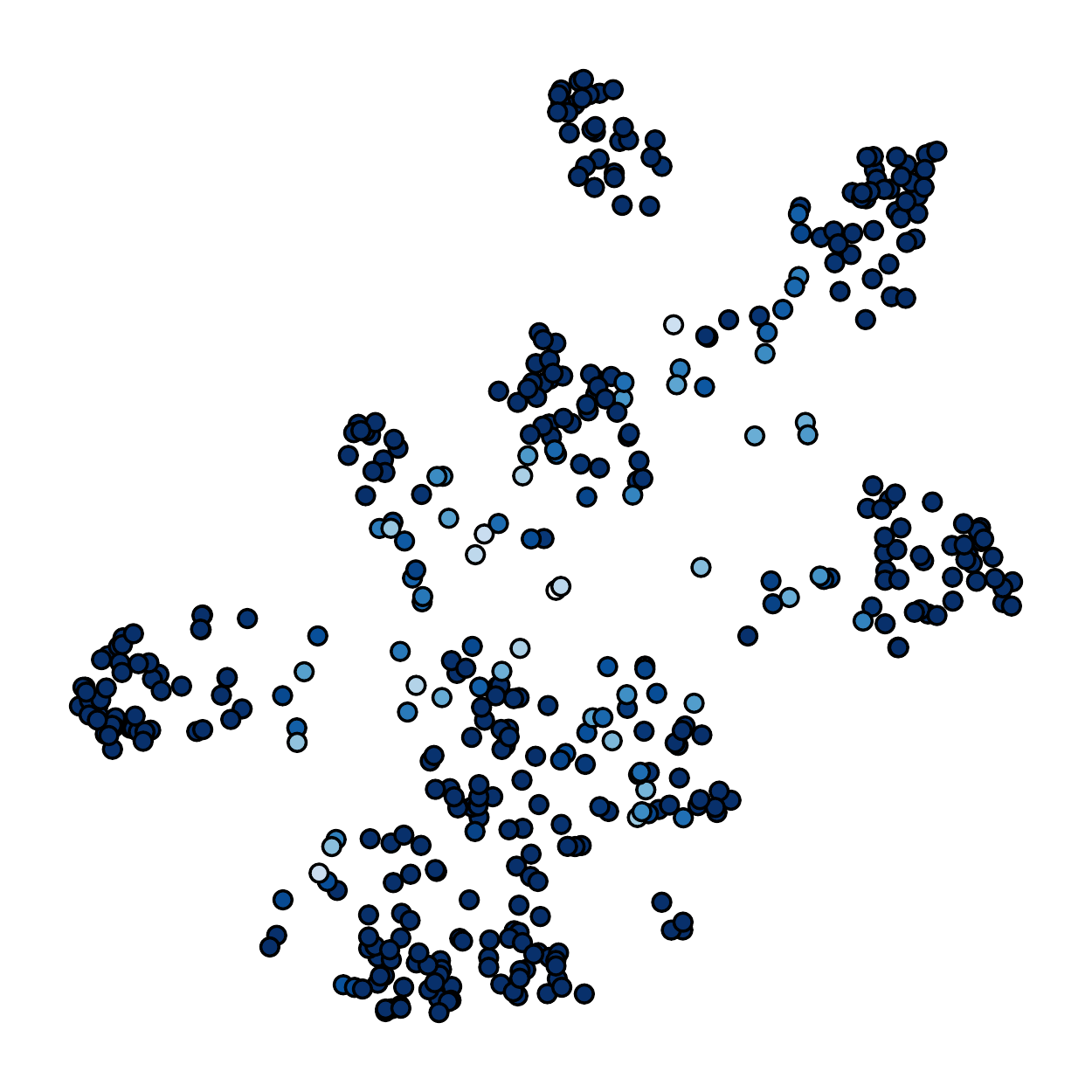}}\hfill & {\includegraphics[width=0.33\linewidth]{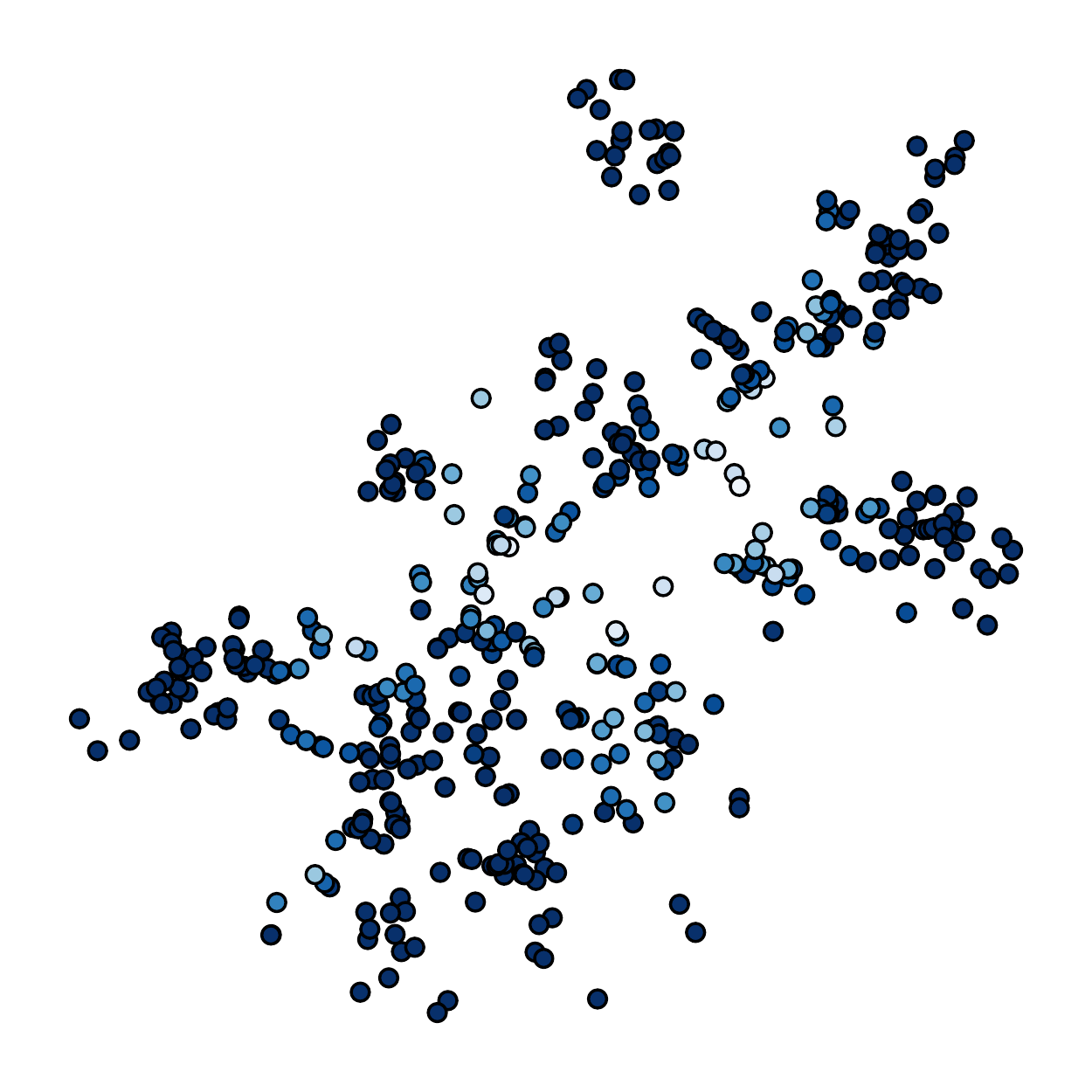}}\hfill  \\ \hline
90000 iter & {\includegraphics[width=0.33\linewidth]{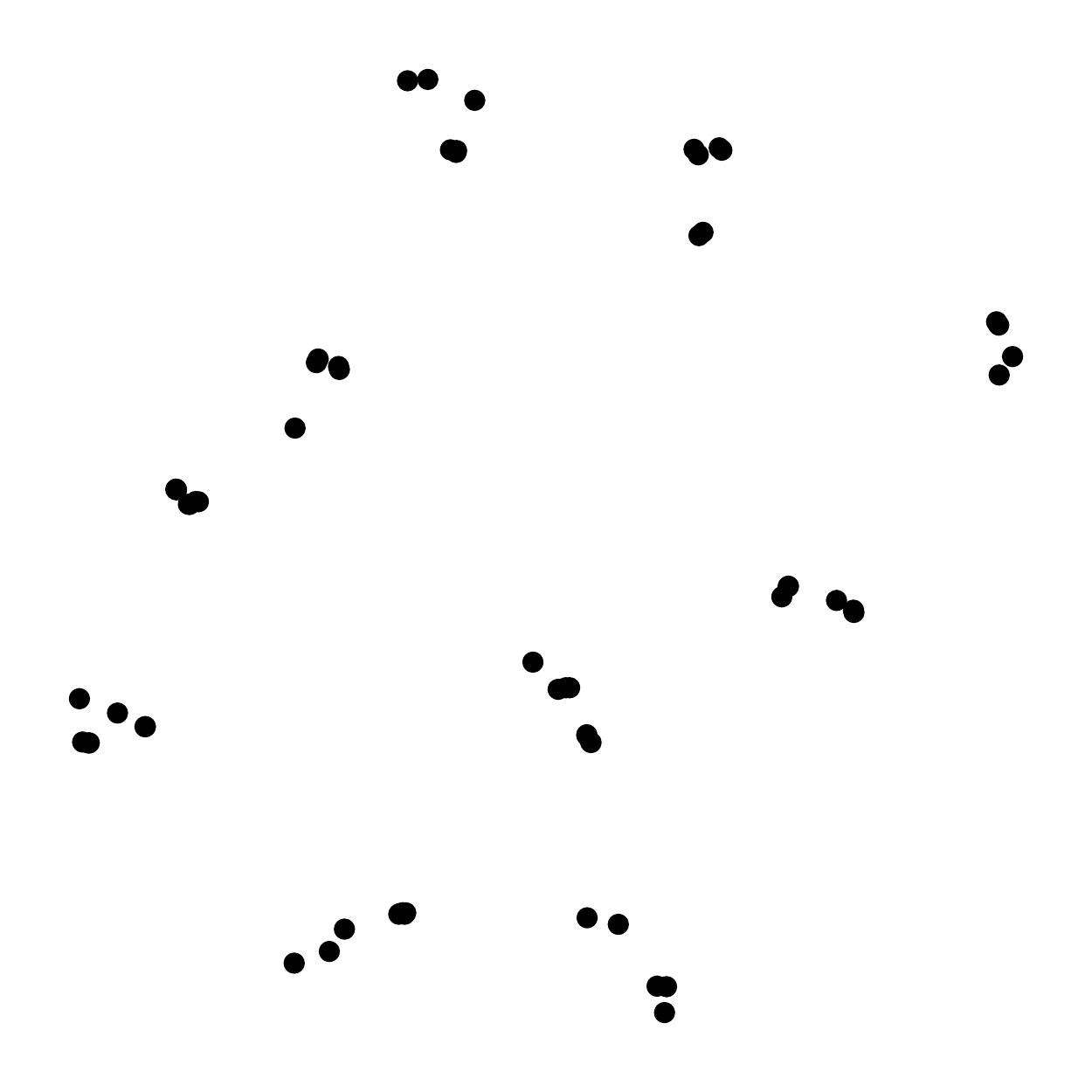}}\hfill &                     {\includegraphics[width=0.33\linewidth]{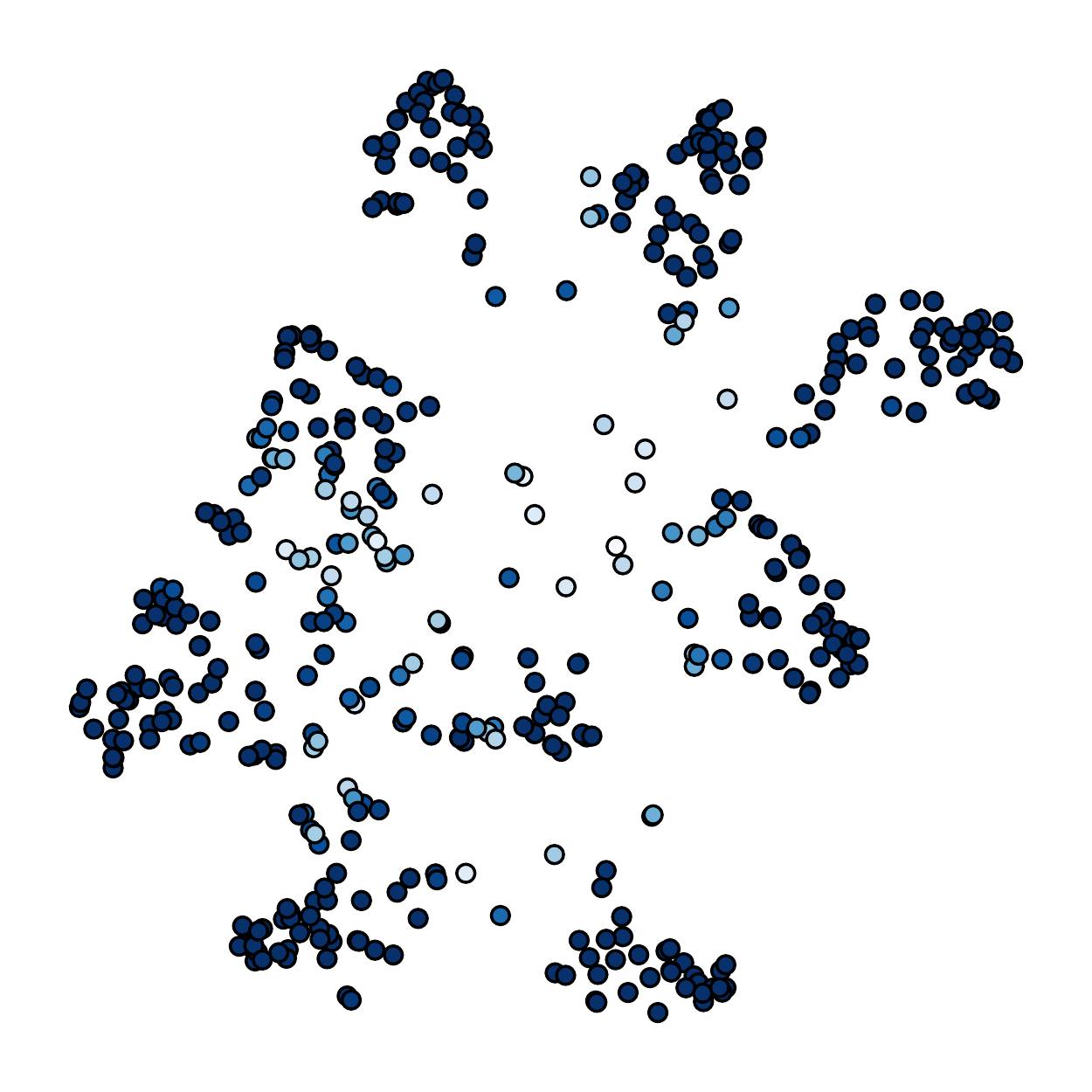}}\hfill & {\includegraphics[width=0.33\linewidth]{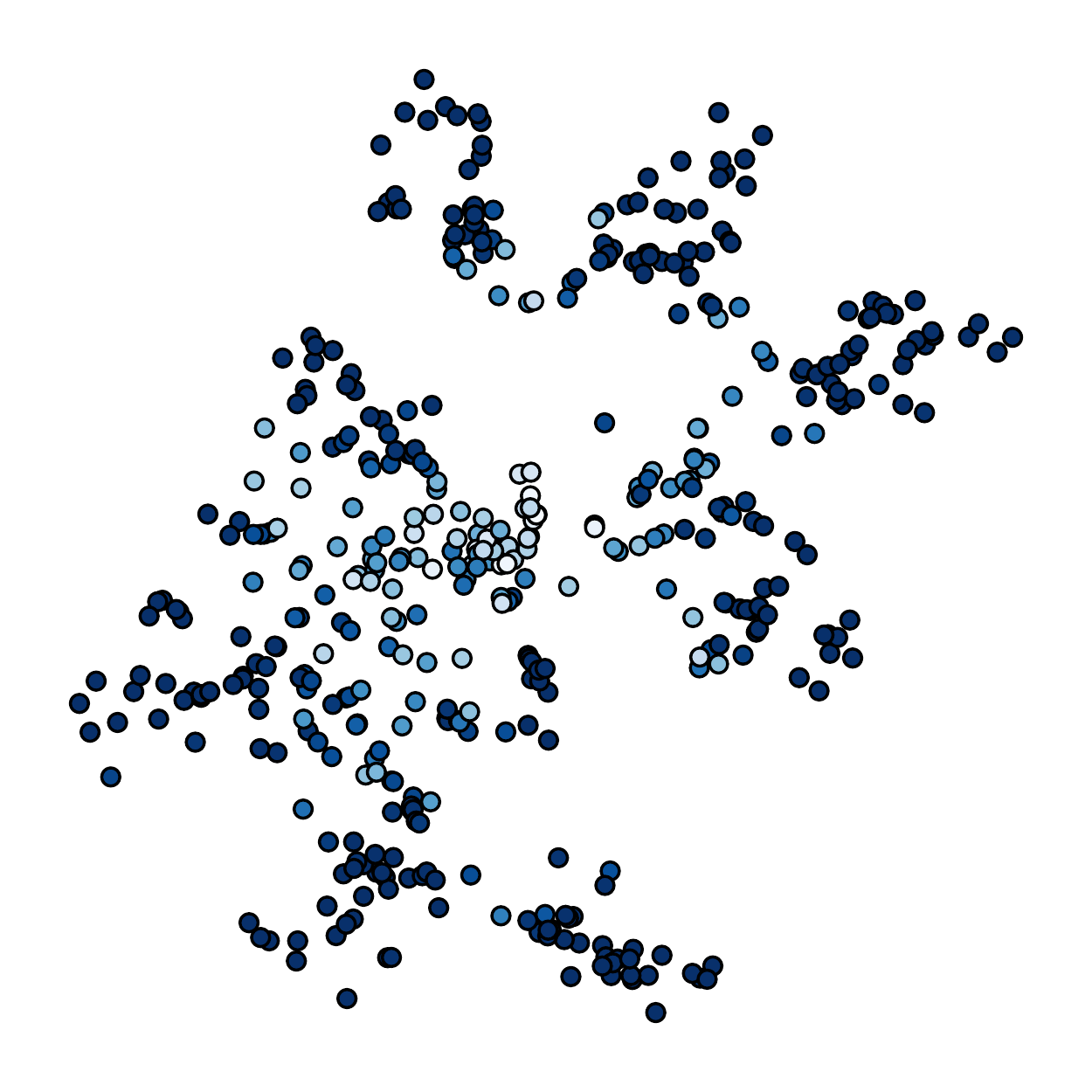}}\hfill  \\ \hline
\end{tabular}
}\vspace{-10pt}
\end{center}
\caption{Visualization on the transitions feature distribution with confidence in \textbf{FixMatch}. The details are same as above.}
\label{table:tsne-fix}
\end{table*}

\begin{table*}[]
\begin{center}
\scalebox{0.8}{
\begin{tabular}{c|c|c|c} 

\textbf{Con+Flex} & Labeled samples  &  Weakly augmented samples &  Strongly augmented samples\\ \hline
10000 iter & {\includegraphics[width=0.33\linewidth]{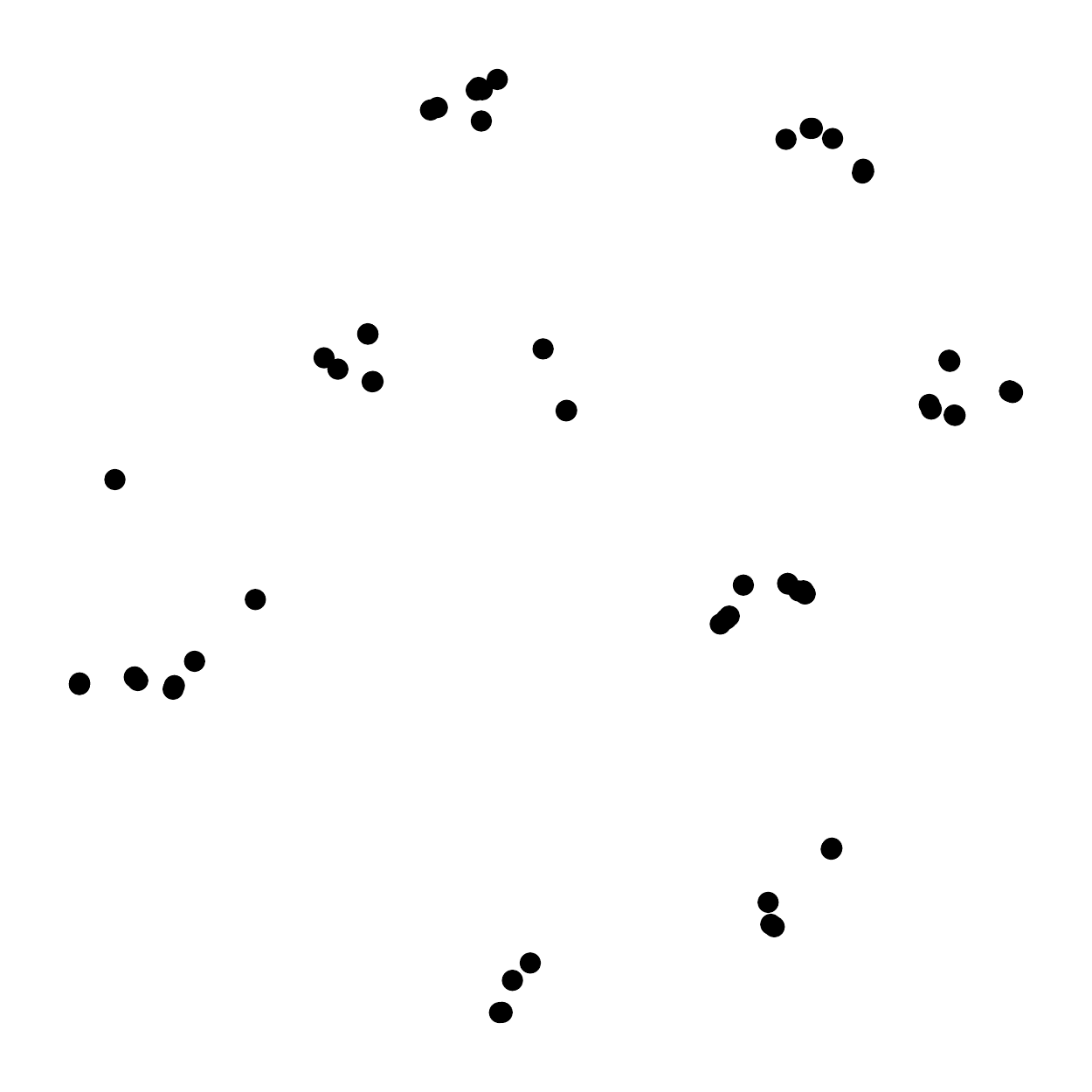}}\hfill &                     {\includegraphics[width=0.33\linewidth]{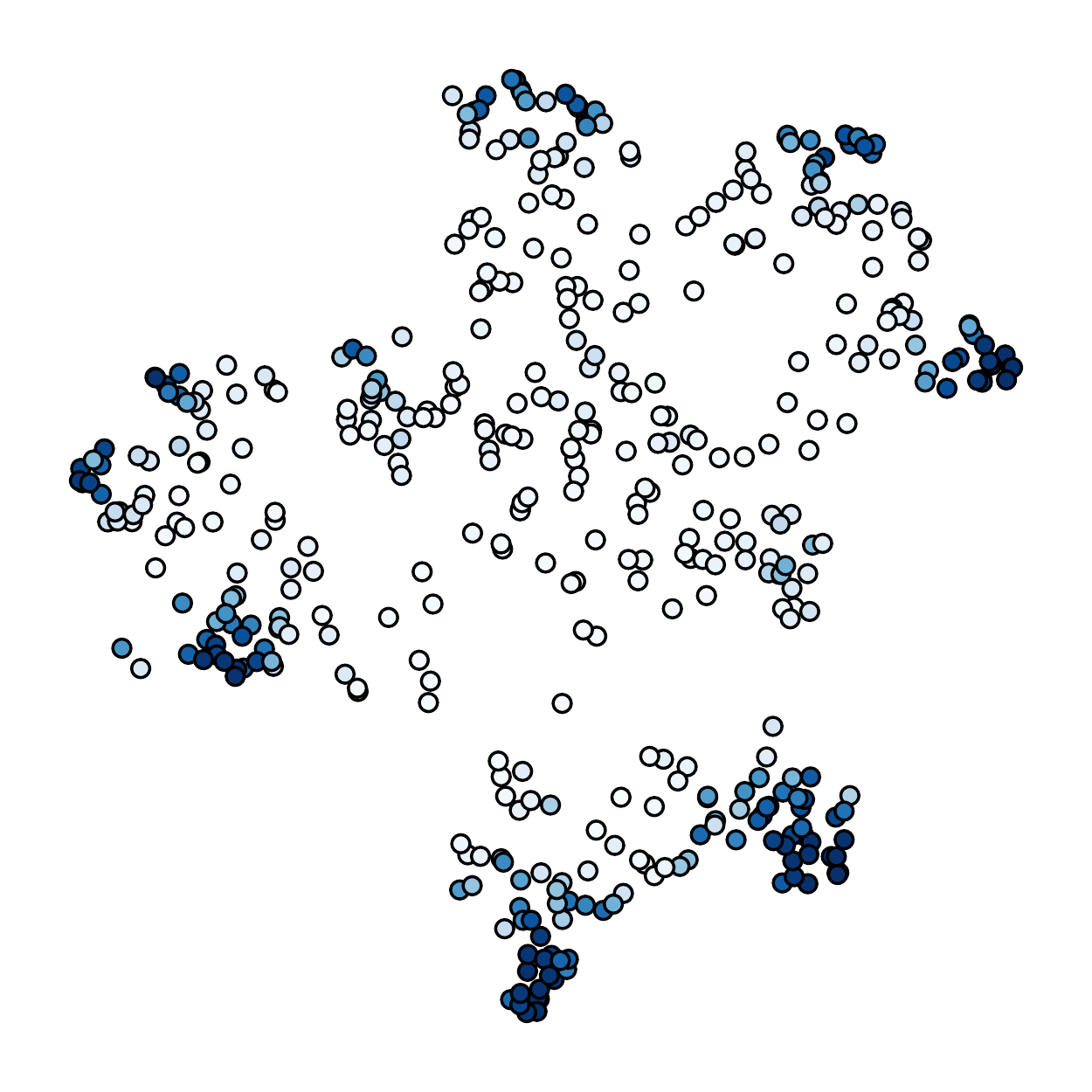}}\hfill & {\includegraphics[width=0.33\linewidth]{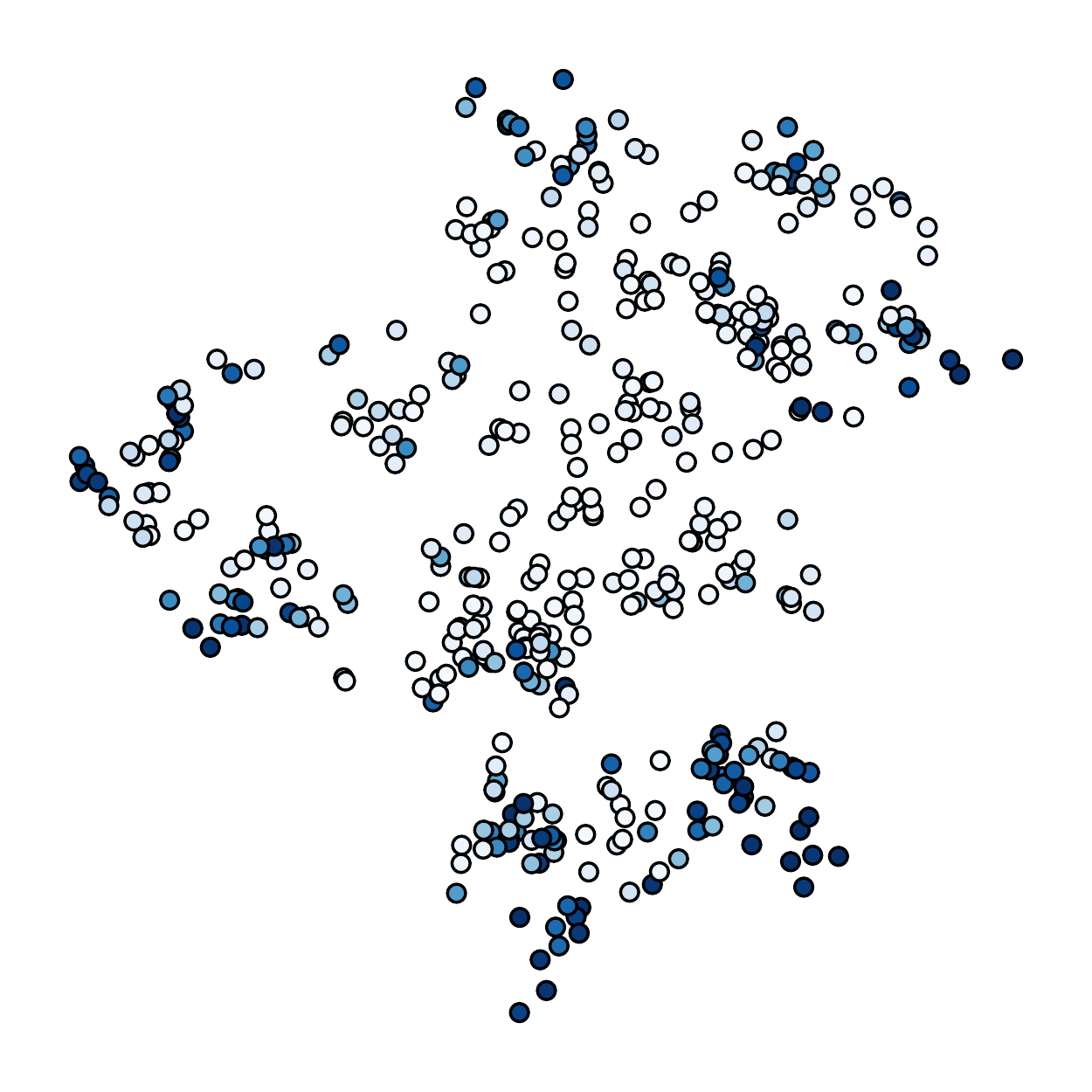}}\hfill  \\ \hline
30000 iter & {\includegraphics[width=0.33\linewidth]{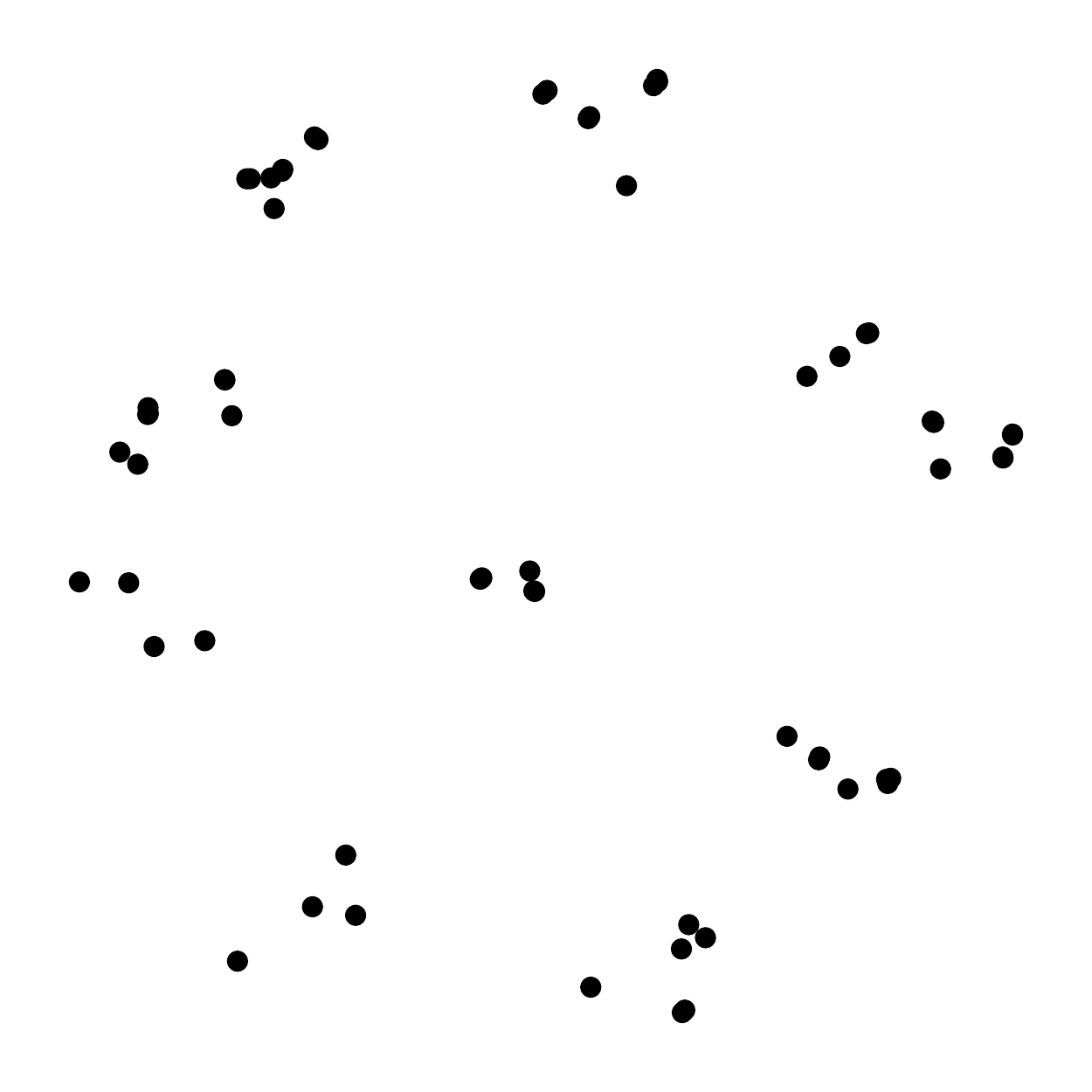}}\hfill &                     {\includegraphics[width=0.33\linewidth]{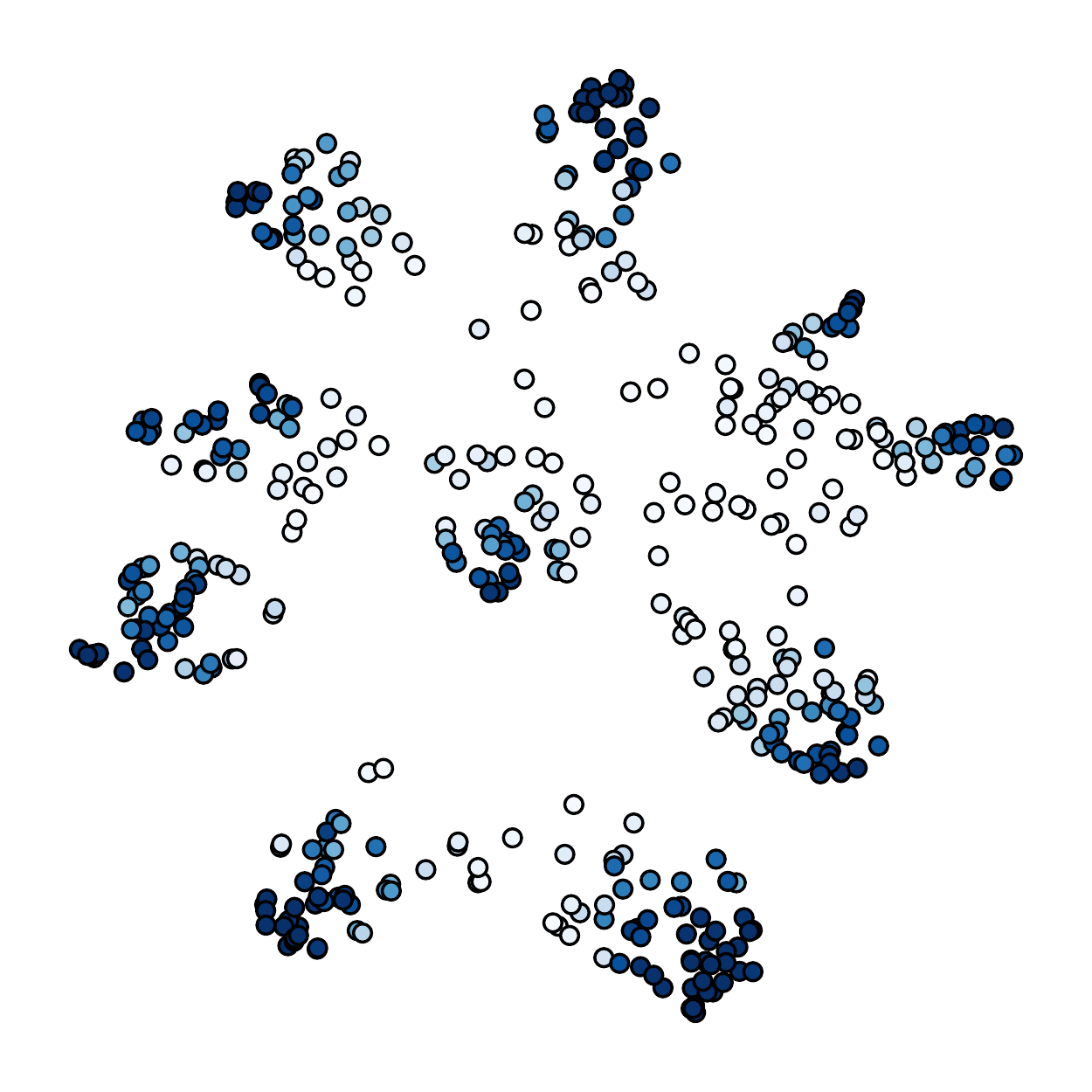}}\hfill & {\includegraphics[width=0.33\linewidth]{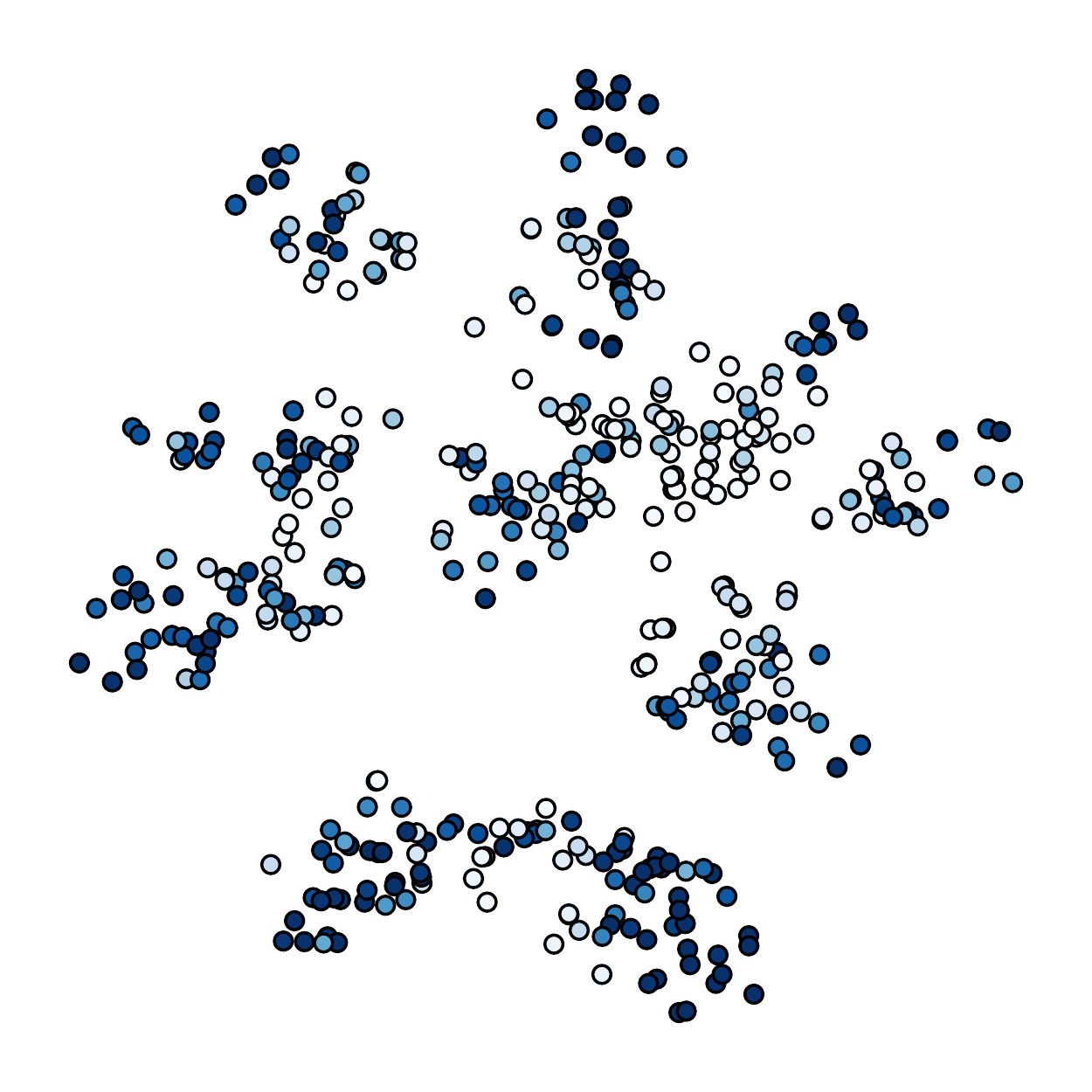}}\hfill  \\ \hline
50000 iter & {\includegraphics[width=0.33\linewidth]{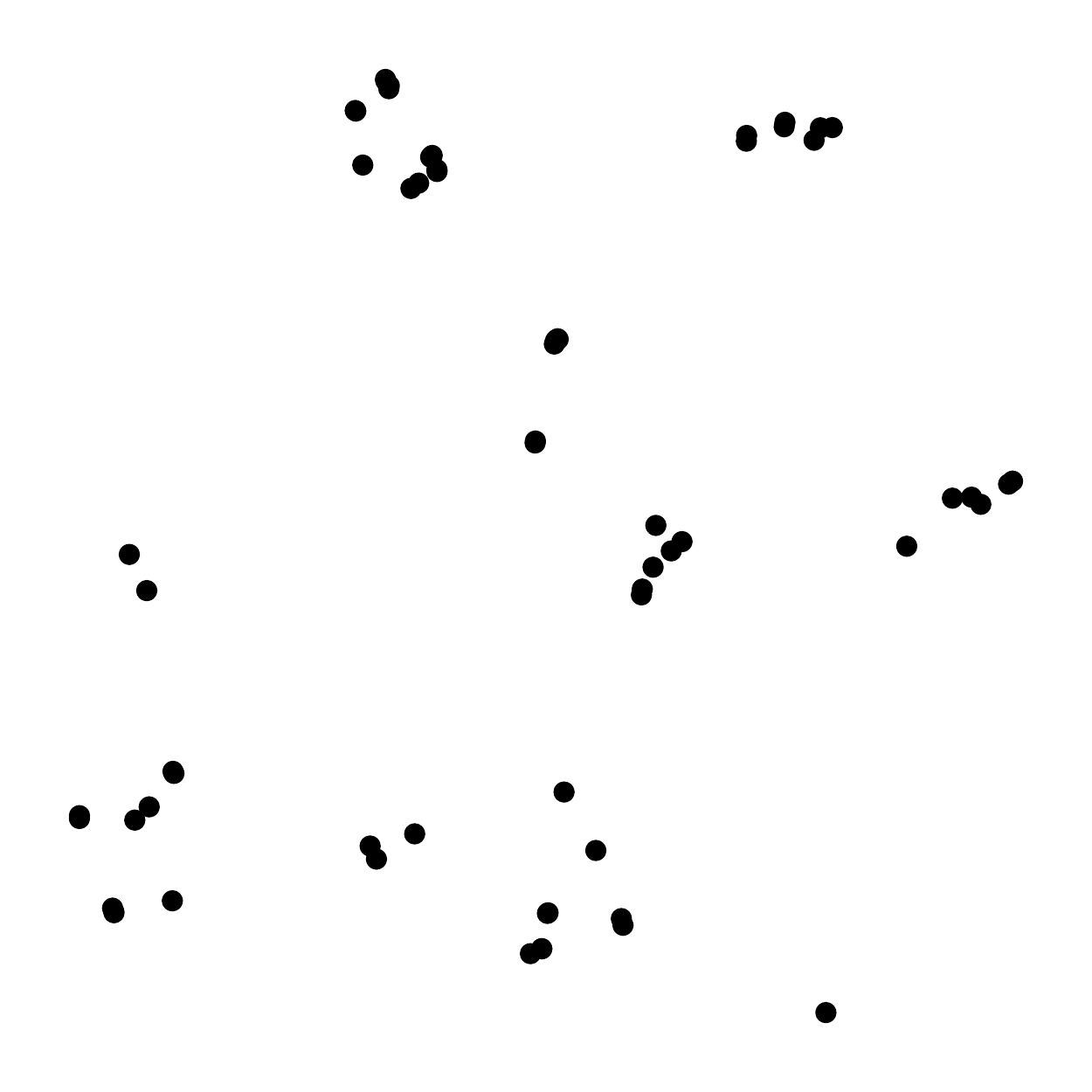}}\hfill &                     {\includegraphics[width=0.33\linewidth]{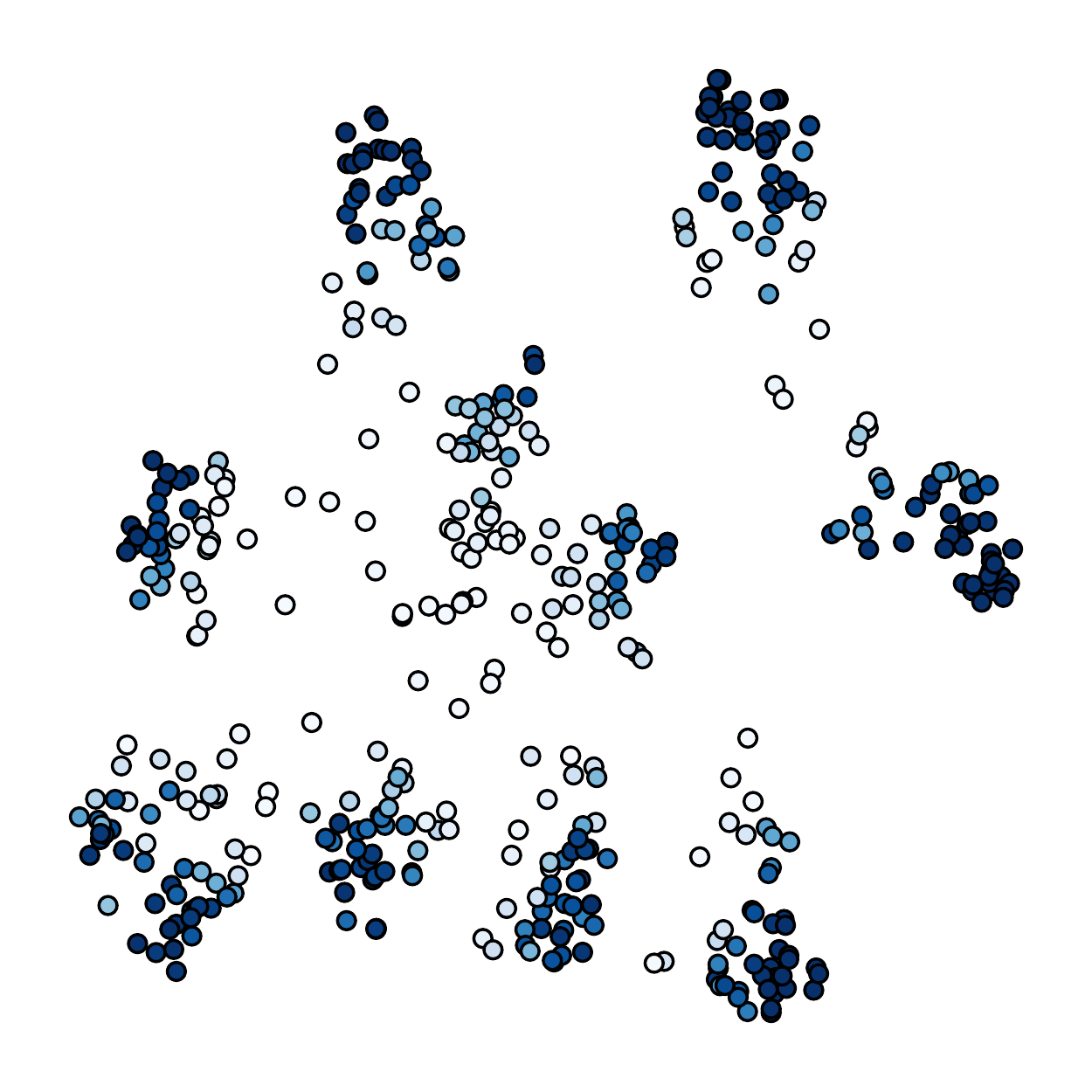}}\hfill & {\includegraphics[width=0.33\linewidth]{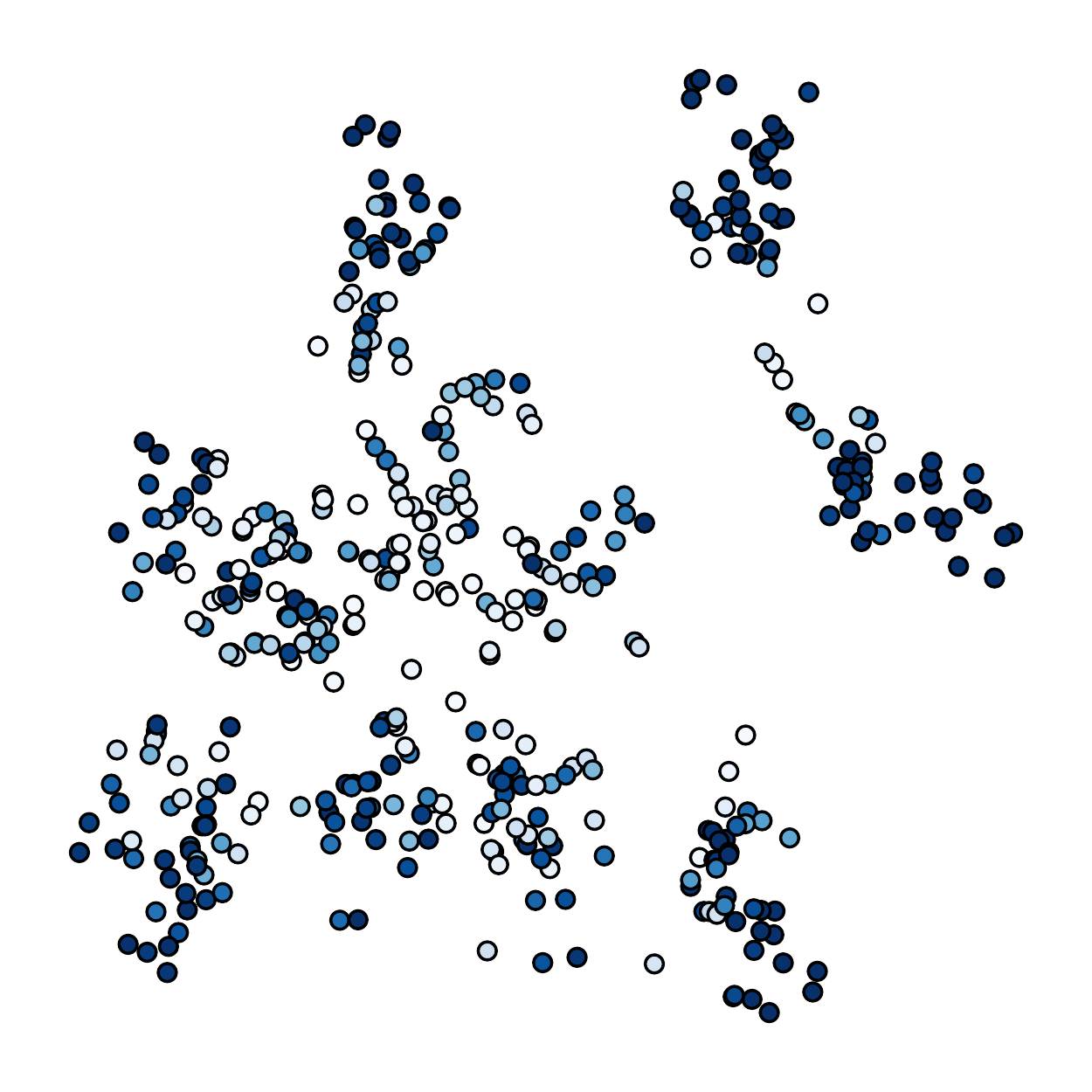}}\hfill  \\ \hline
70000 iter & {\includegraphics[width=0.33\linewidth]{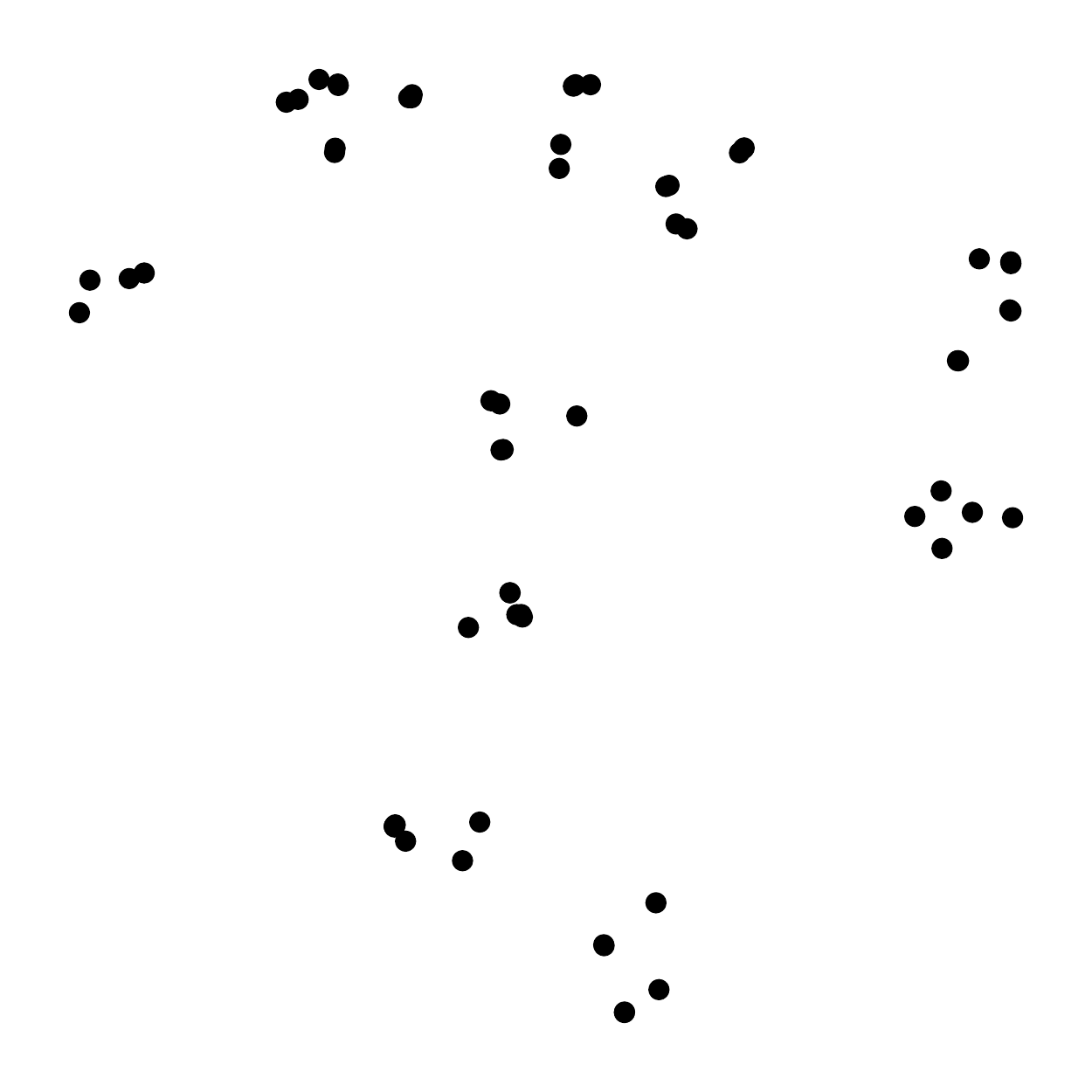}}\hfill &                     {\includegraphics[width=0.33\linewidth]{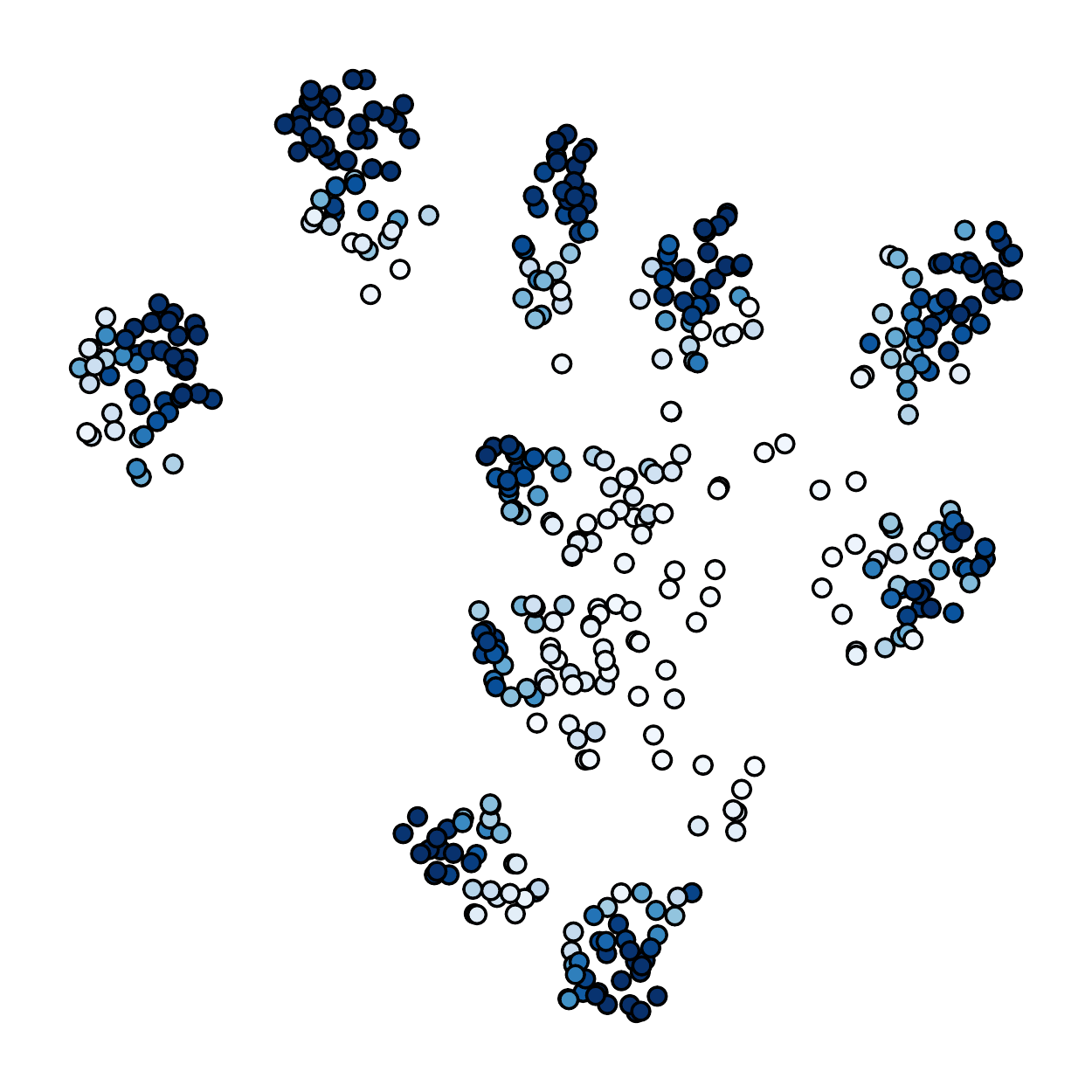}}\hfill & {\includegraphics[width=0.33\linewidth]{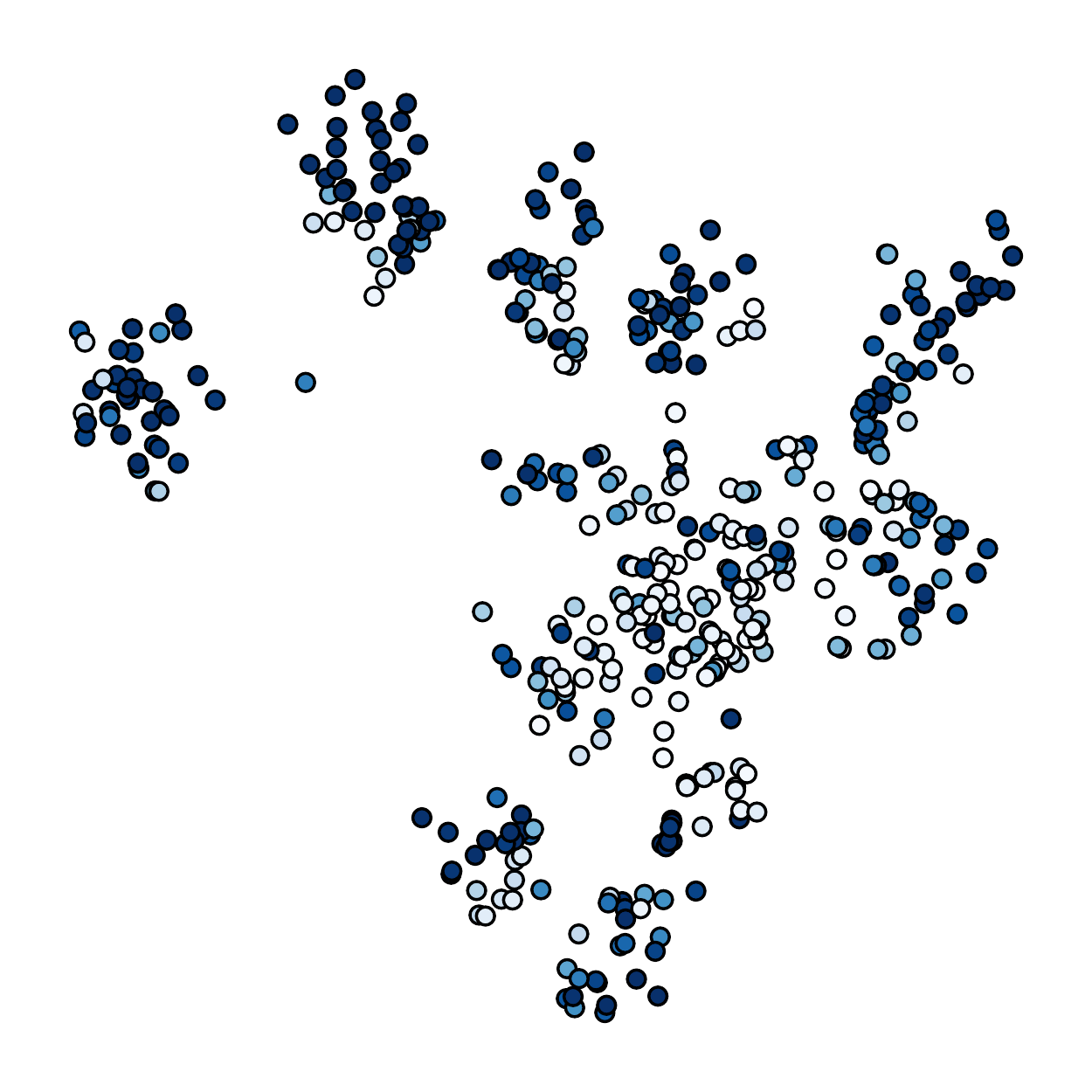}}\hfill  \\ \hline
90000 iter & {\includegraphics[width=0.33\linewidth]{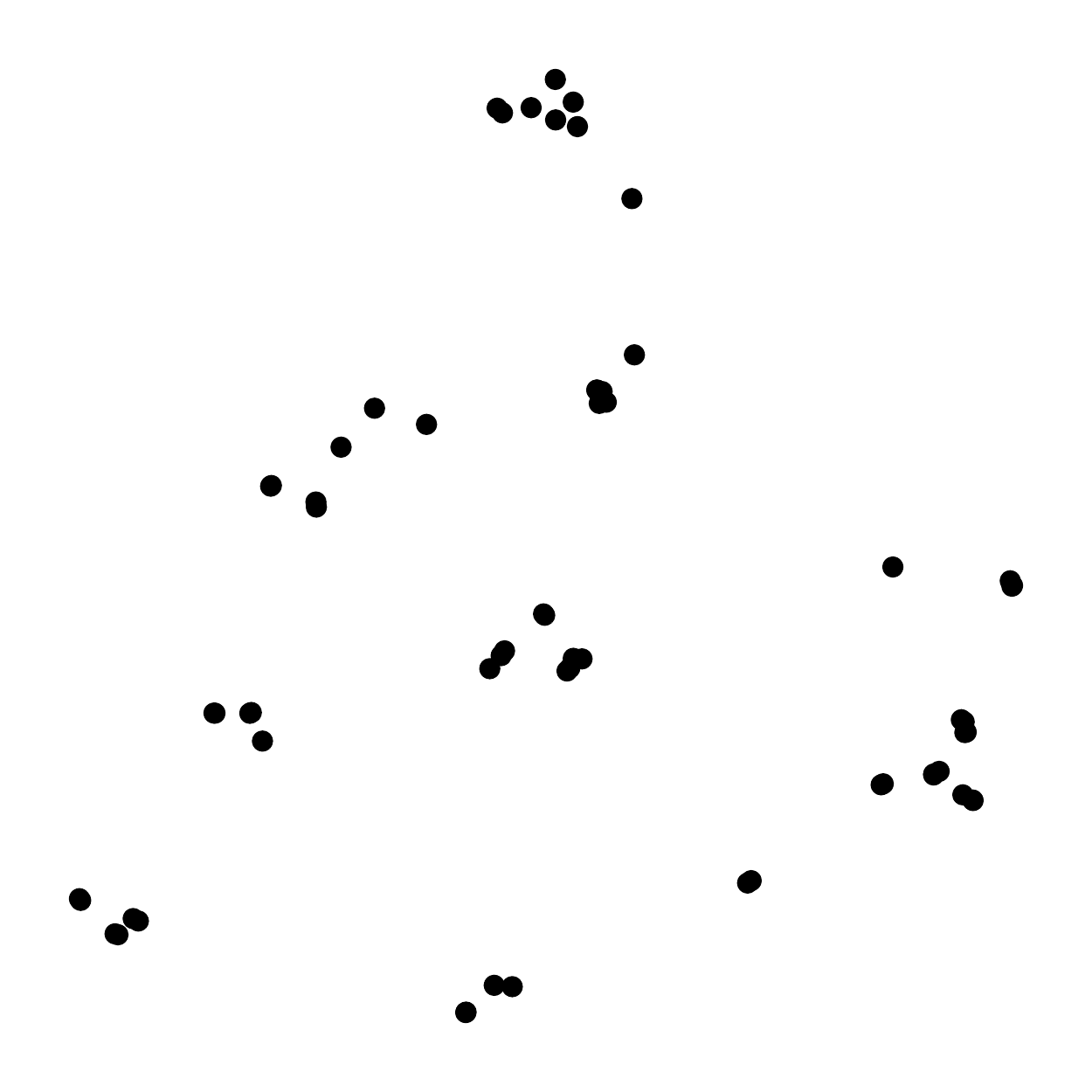}}\hfill &                     {\includegraphics[width=0.33\linewidth]{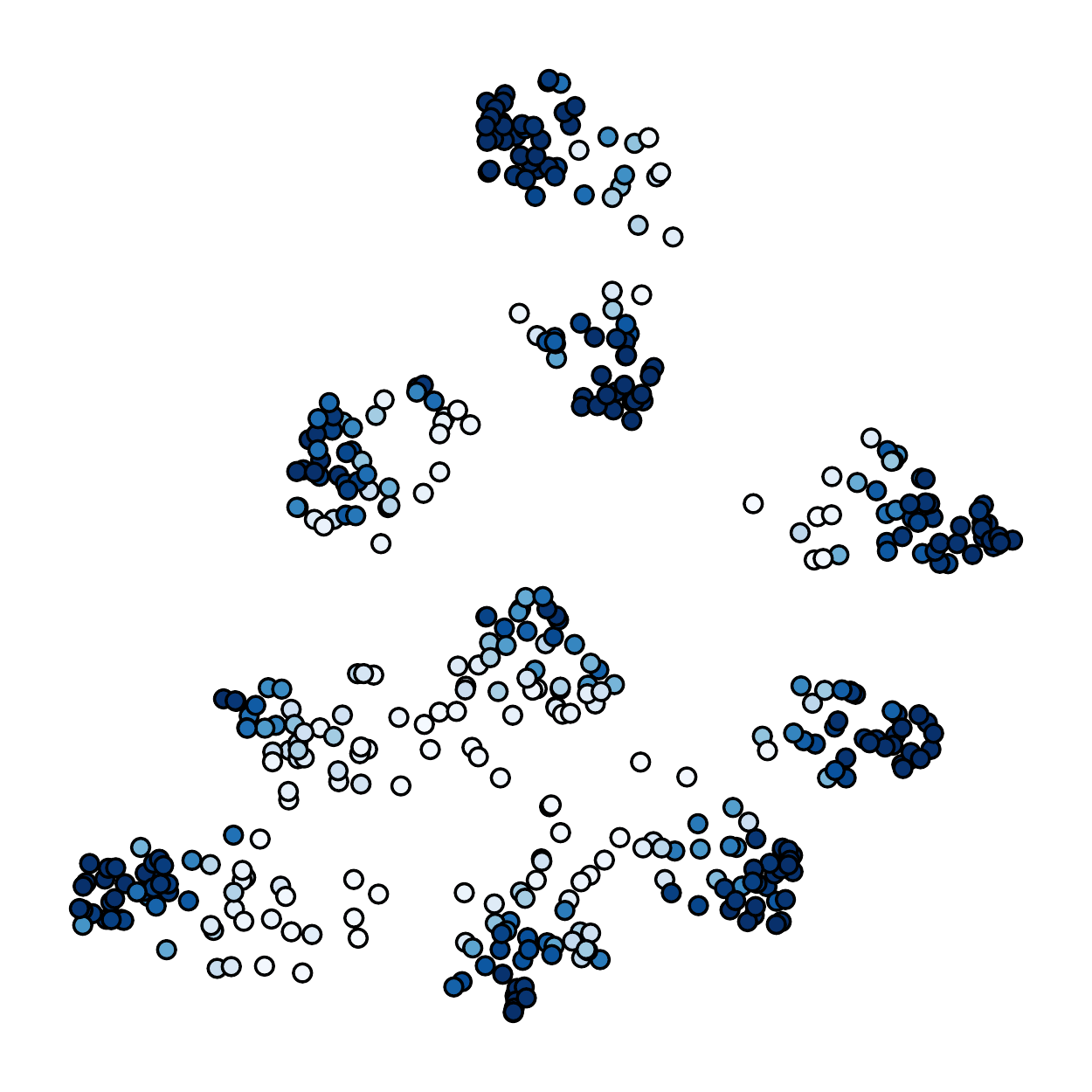}}\hfill & {\includegraphics[width=0.33\linewidth]{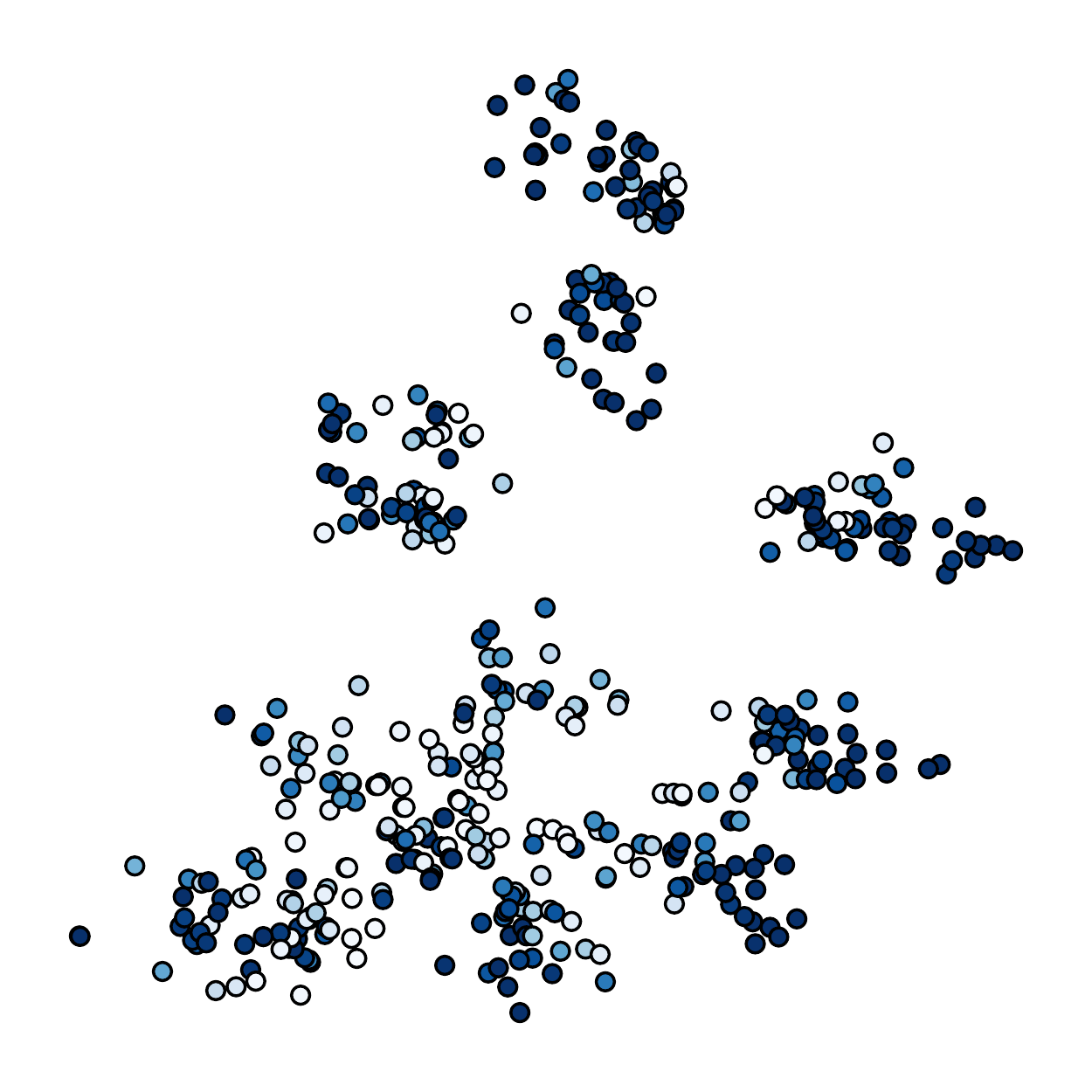}}\hfill  \\ \hline
\end{tabular}
}\vspace{-10pt}
\end{center}
\caption{Visualization on the transitions feature distribution with confidence in \textbf{ConMatch w/FlexMatch}. The details are same as above.}
\label{table:tsne-conflex}
\end{table*}

\begin{table*}[]
\begin{center}
\scalebox{0.8}{
\begin{tabular}{c|c|c|c} 

\textbf{FlexMatch} & Labeled samples  &  Weakly augmented samples &  Strongly augmented samples\\ \hline
10000 iter & {\includegraphics[width=0.33\linewidth]{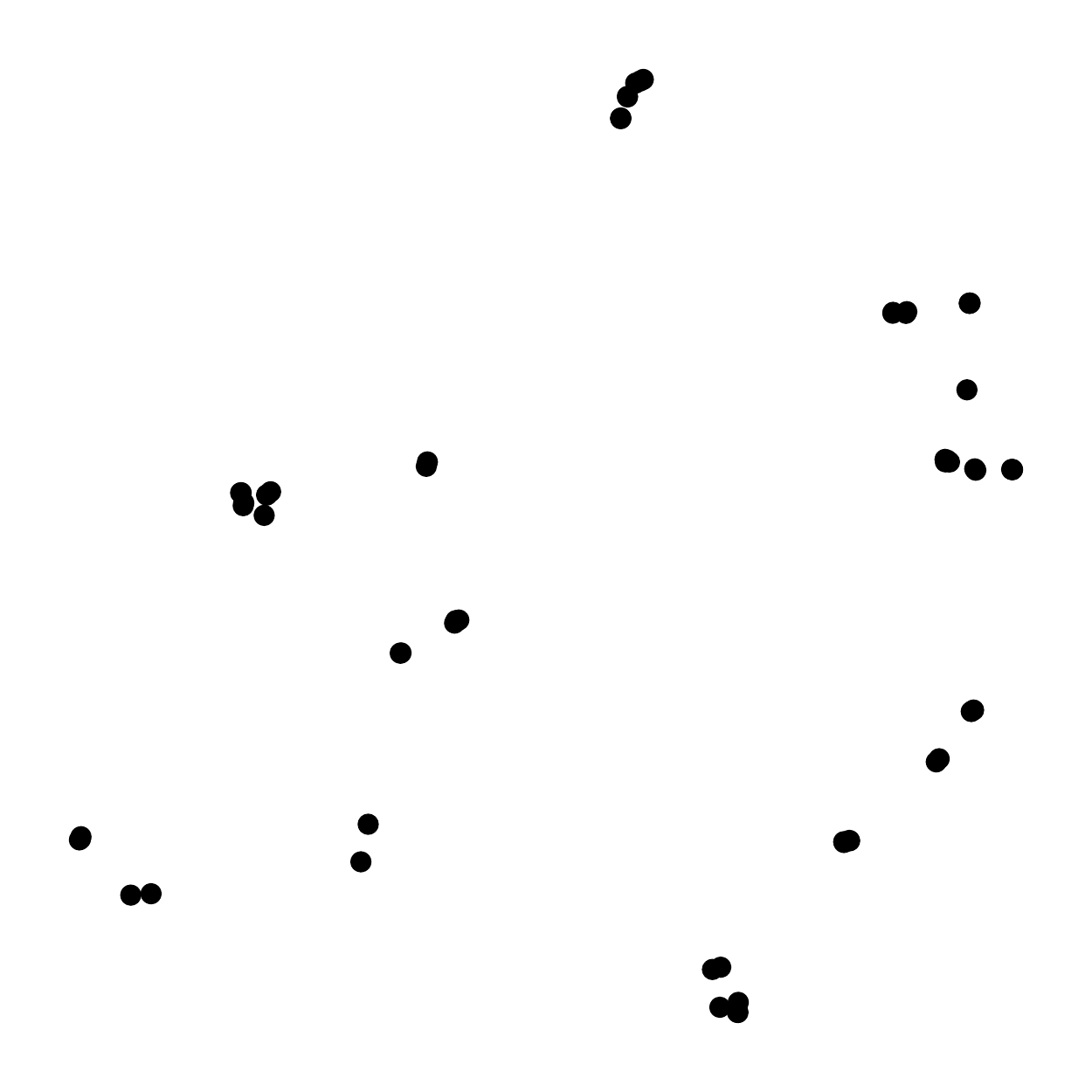}}\hfill &                     {\includegraphics[width=0.33\linewidth]{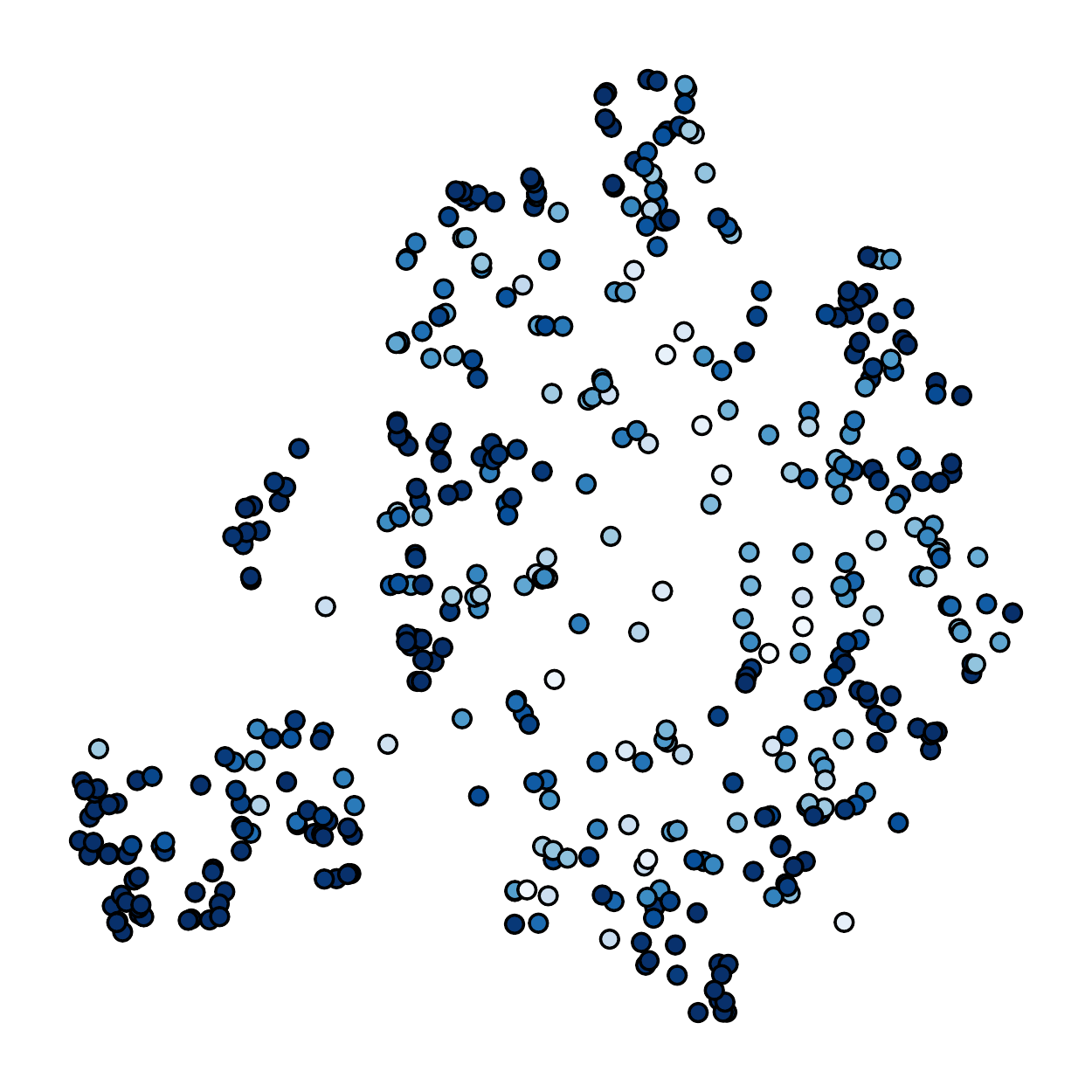}}\hfill & {\includegraphics[width=0.33\linewidth]{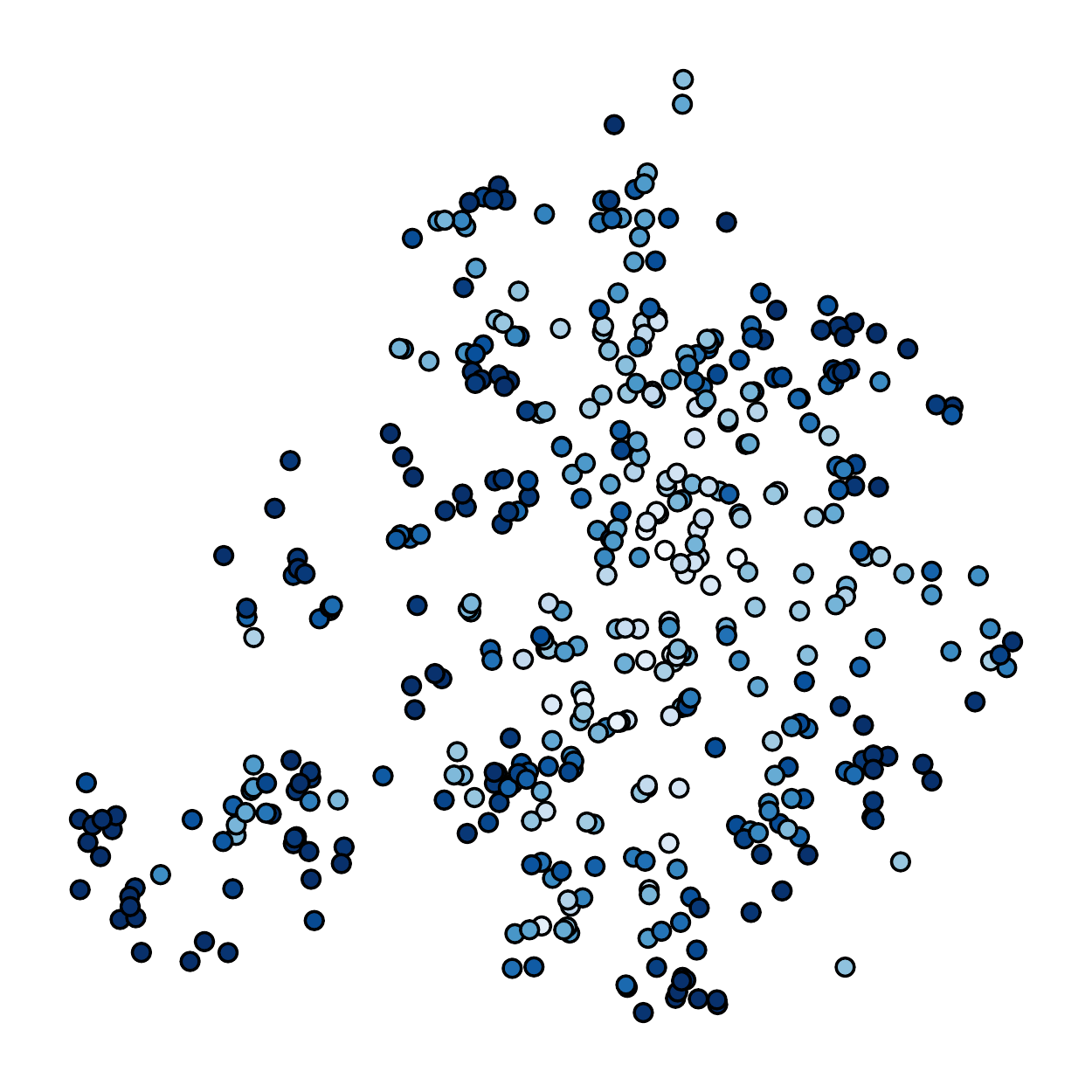}}\hfill  \\ \hline
30000 iter & {\includegraphics[width=0.33\linewidth]{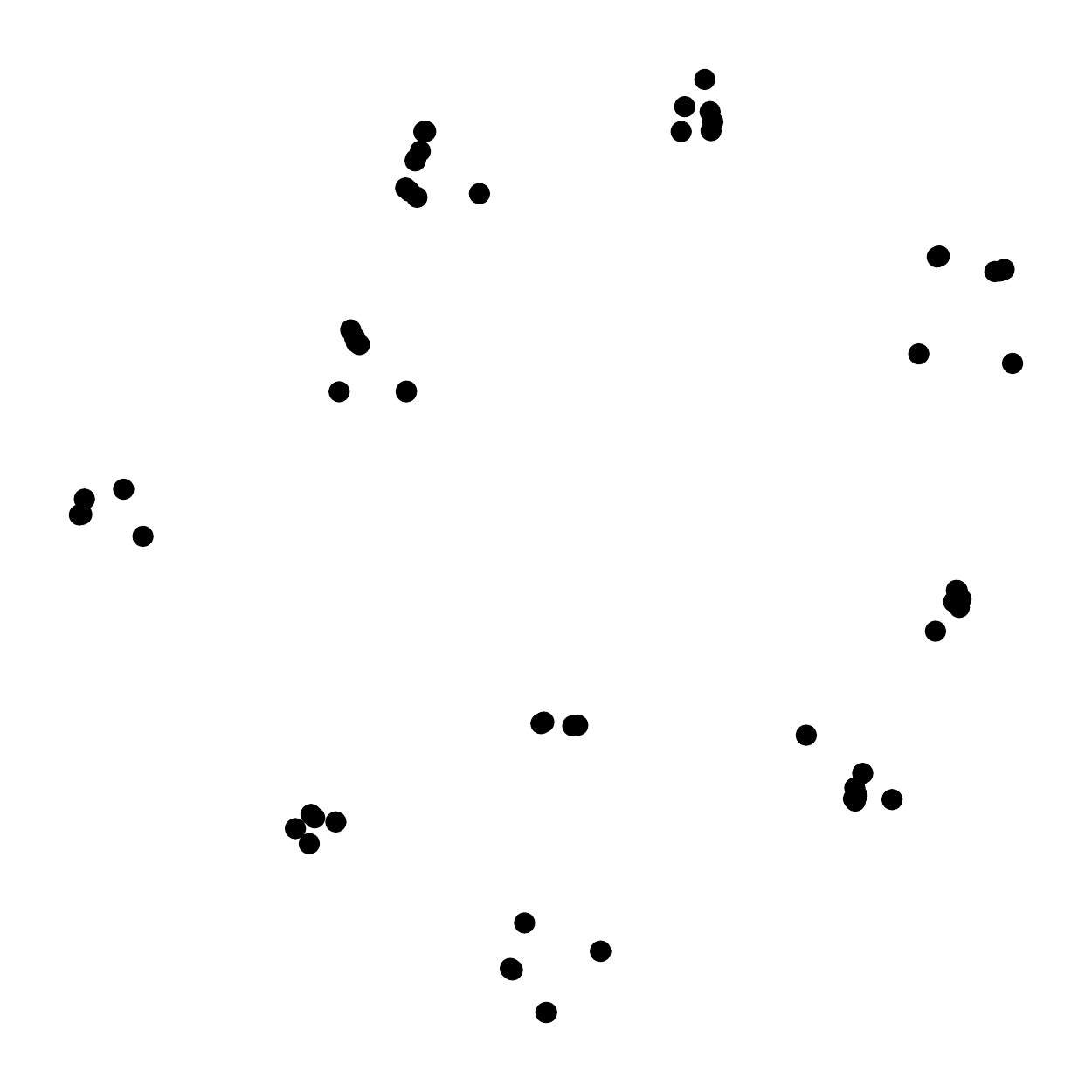}}\hfill &                     {\includegraphics[width=0.33\linewidth]{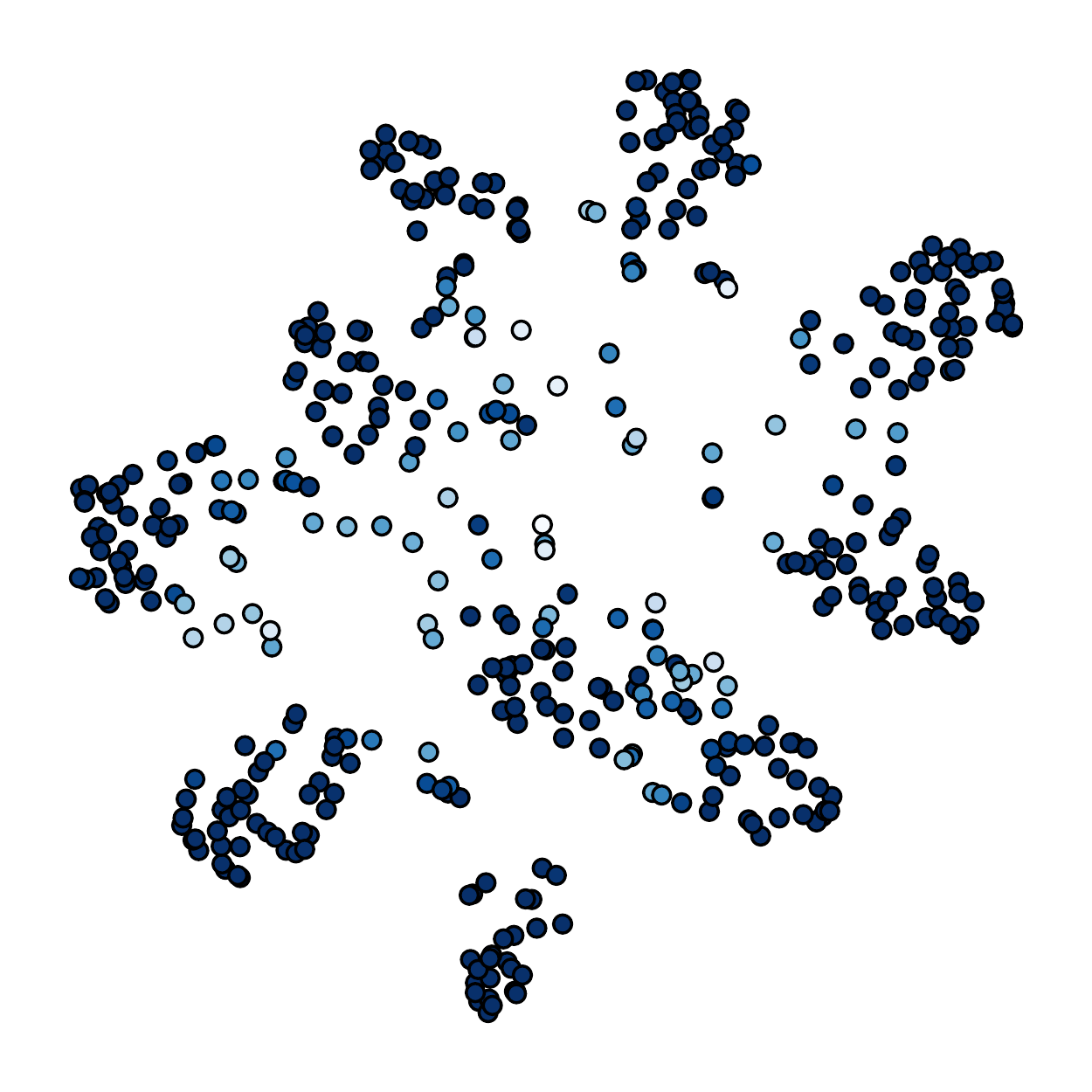}}\hfill & {\includegraphics[width=0.33\linewidth]{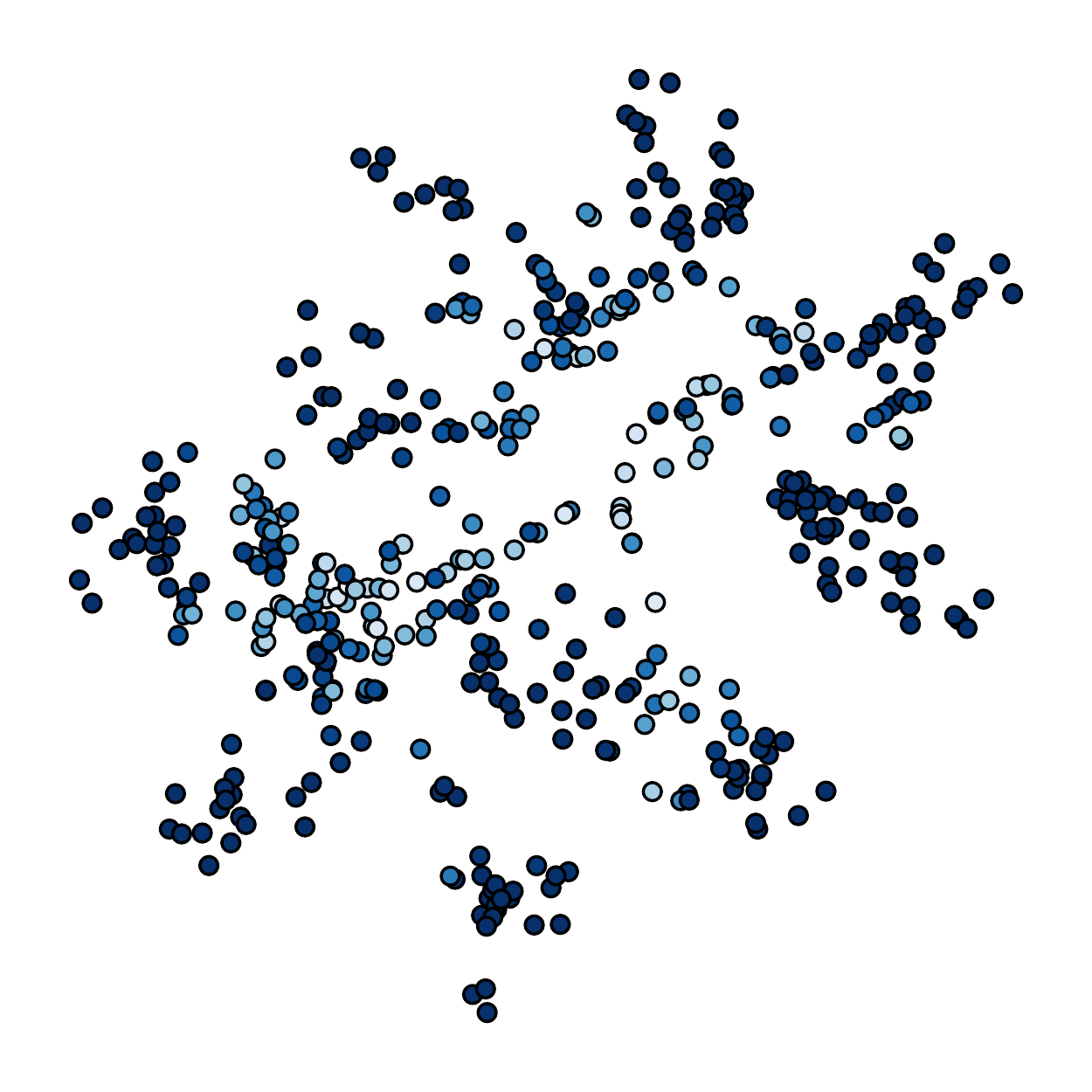}}\hfill  \\ \hline
50000 iter & {\includegraphics[width=0.33\linewidth]{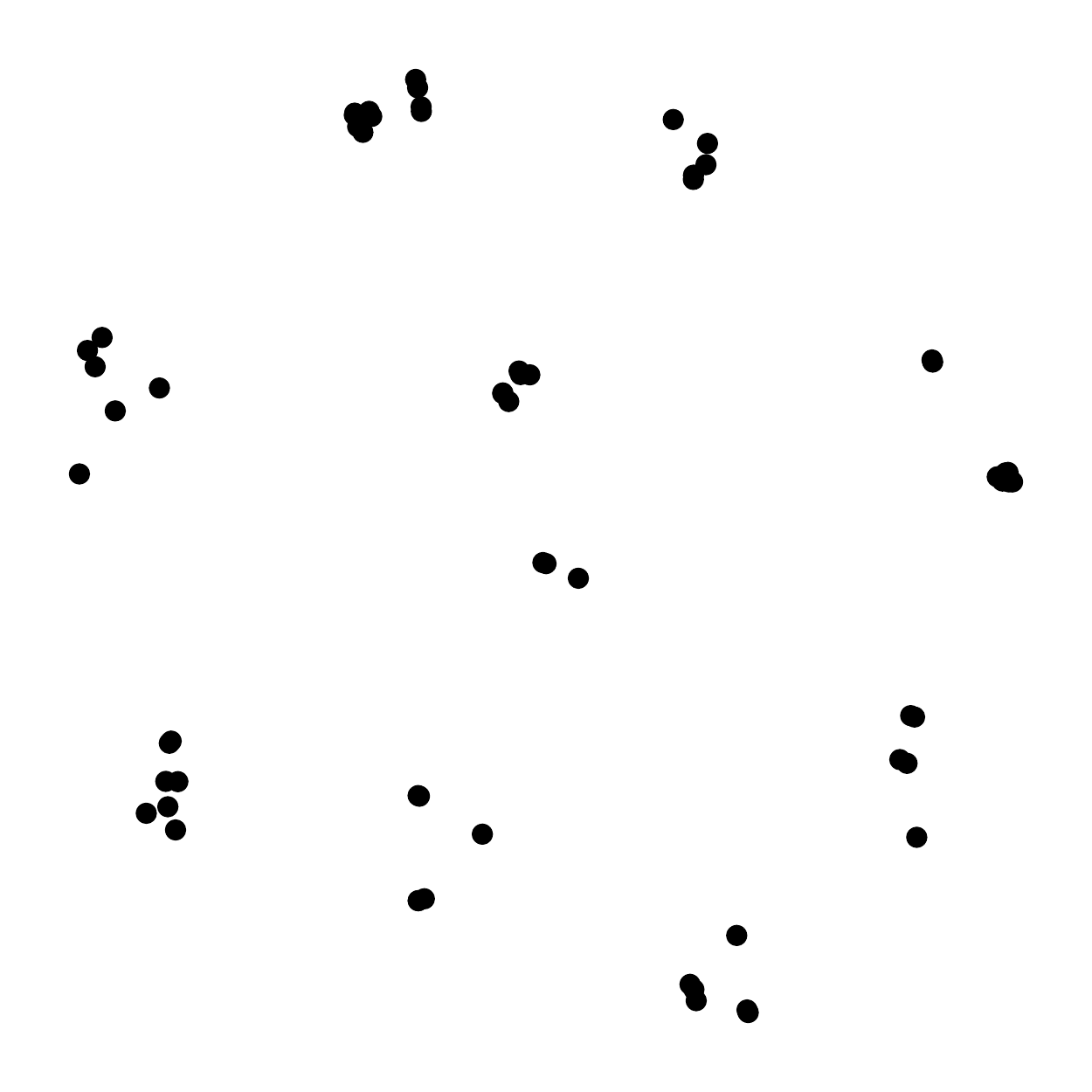}}\hfill &                     {\includegraphics[width=0.33\linewidth]{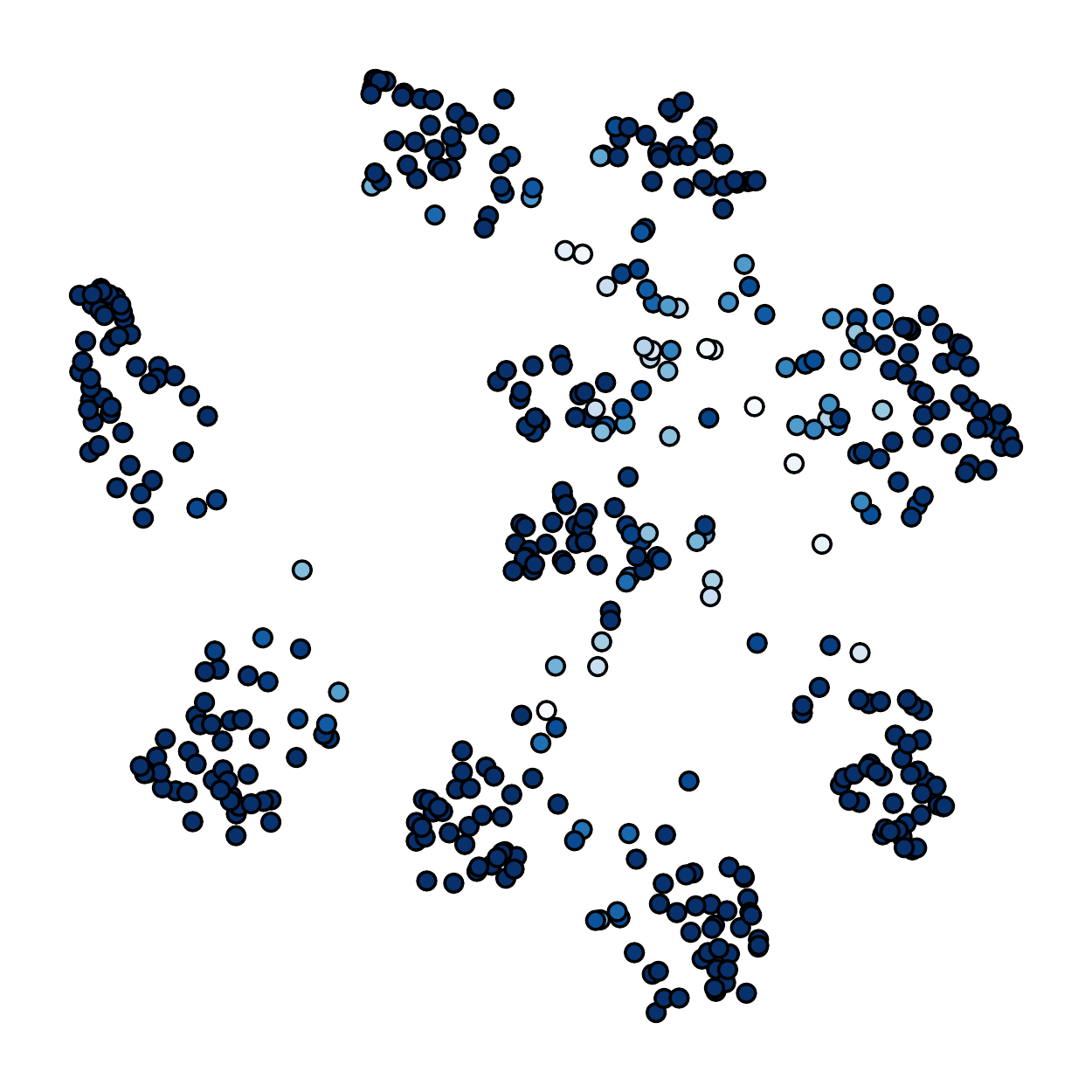}}\hfill & {\includegraphics[width=0.33\linewidth]{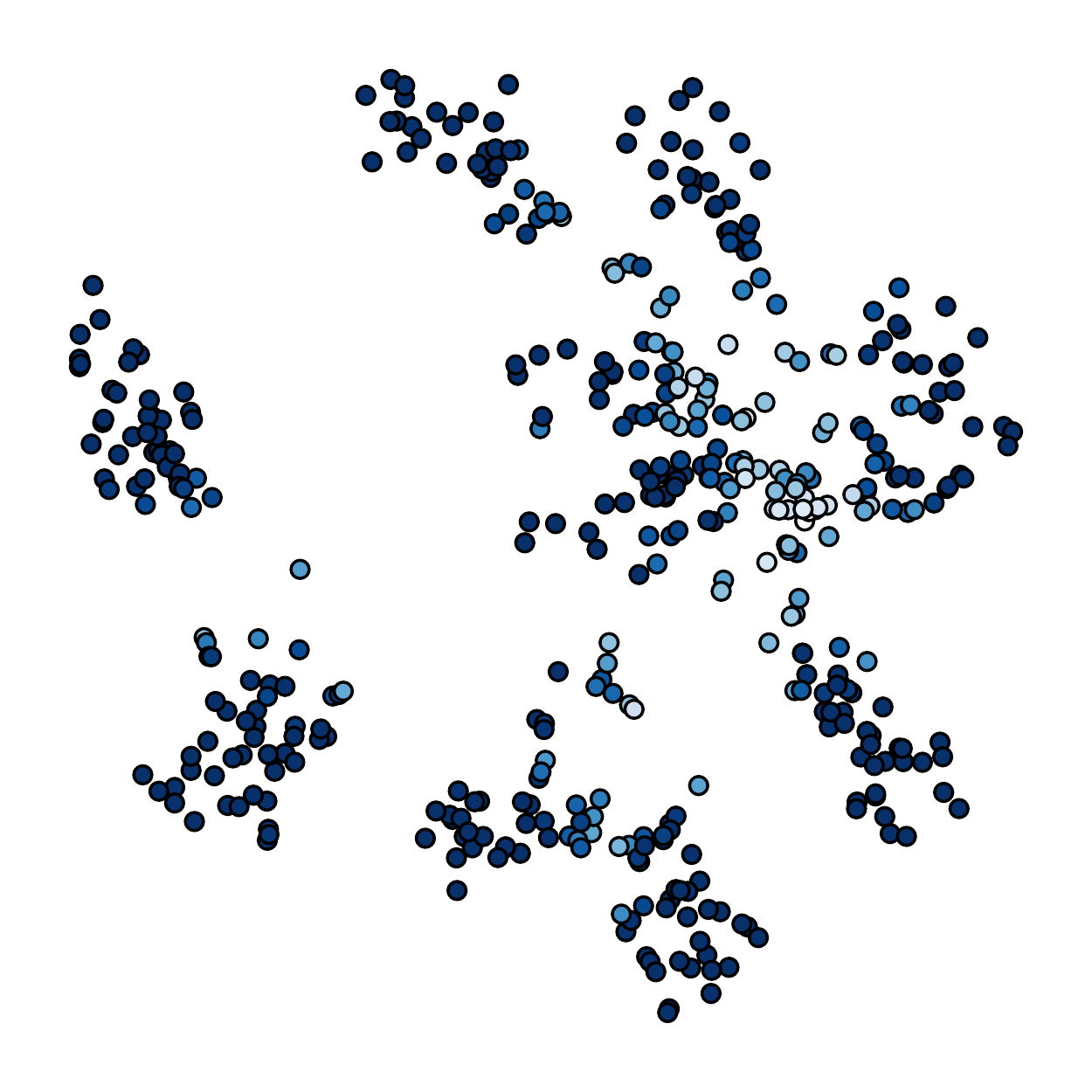}}\hfill  \\ \hline
70000 iter & {\includegraphics[width=0.33\linewidth]{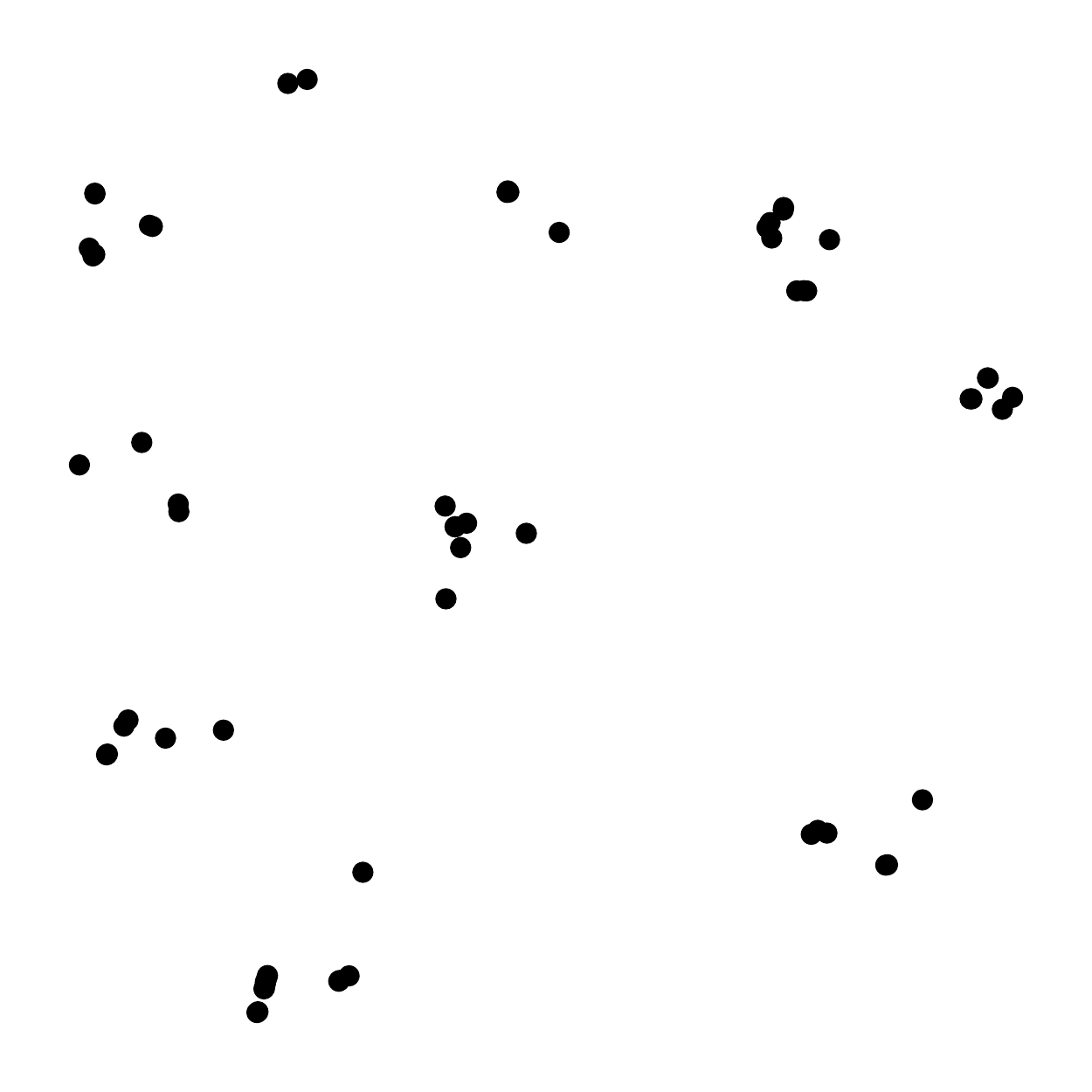}}\hfill &                     {\includegraphics[width=0.33\linewidth]{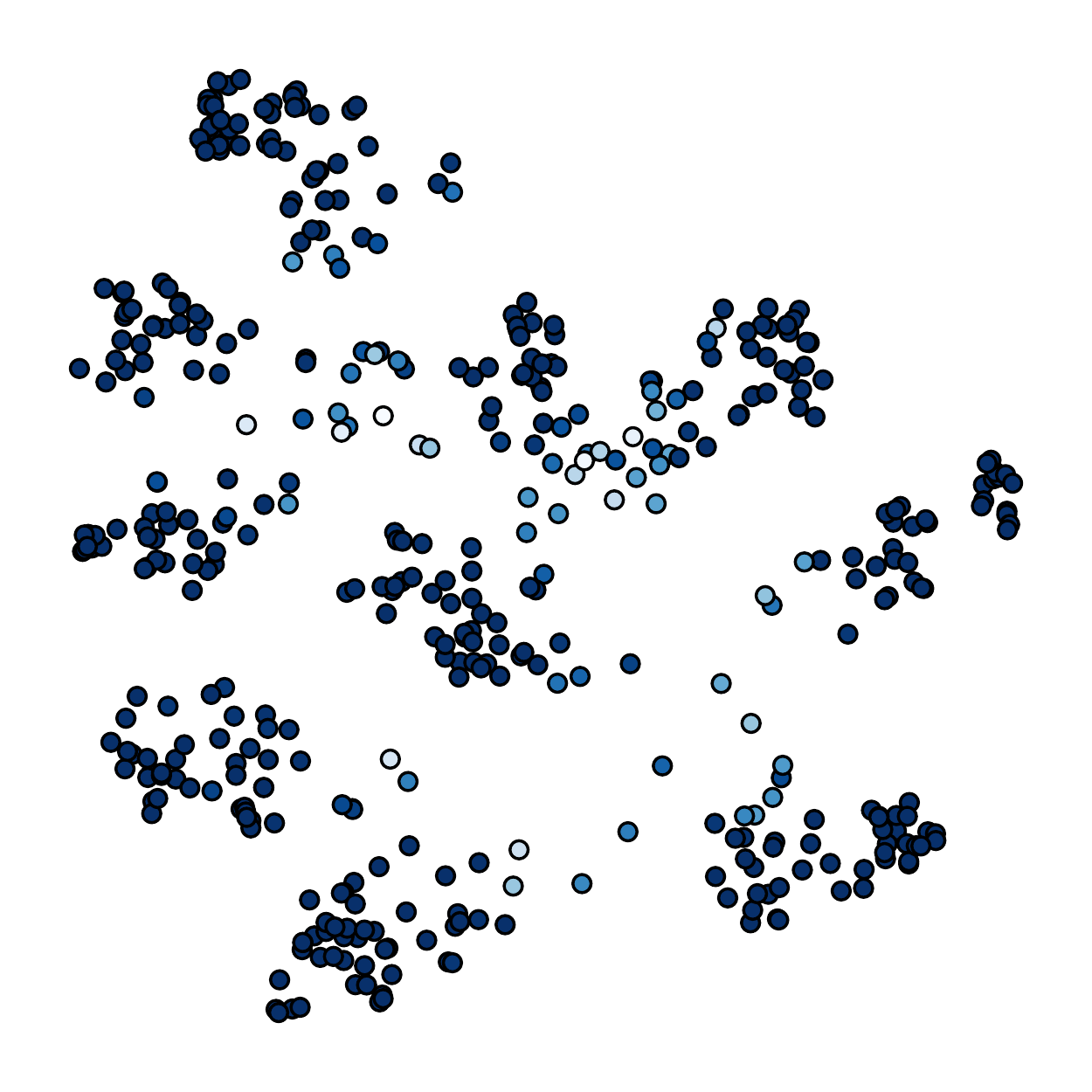}}\hfill & {\includegraphics[width=0.33\linewidth]{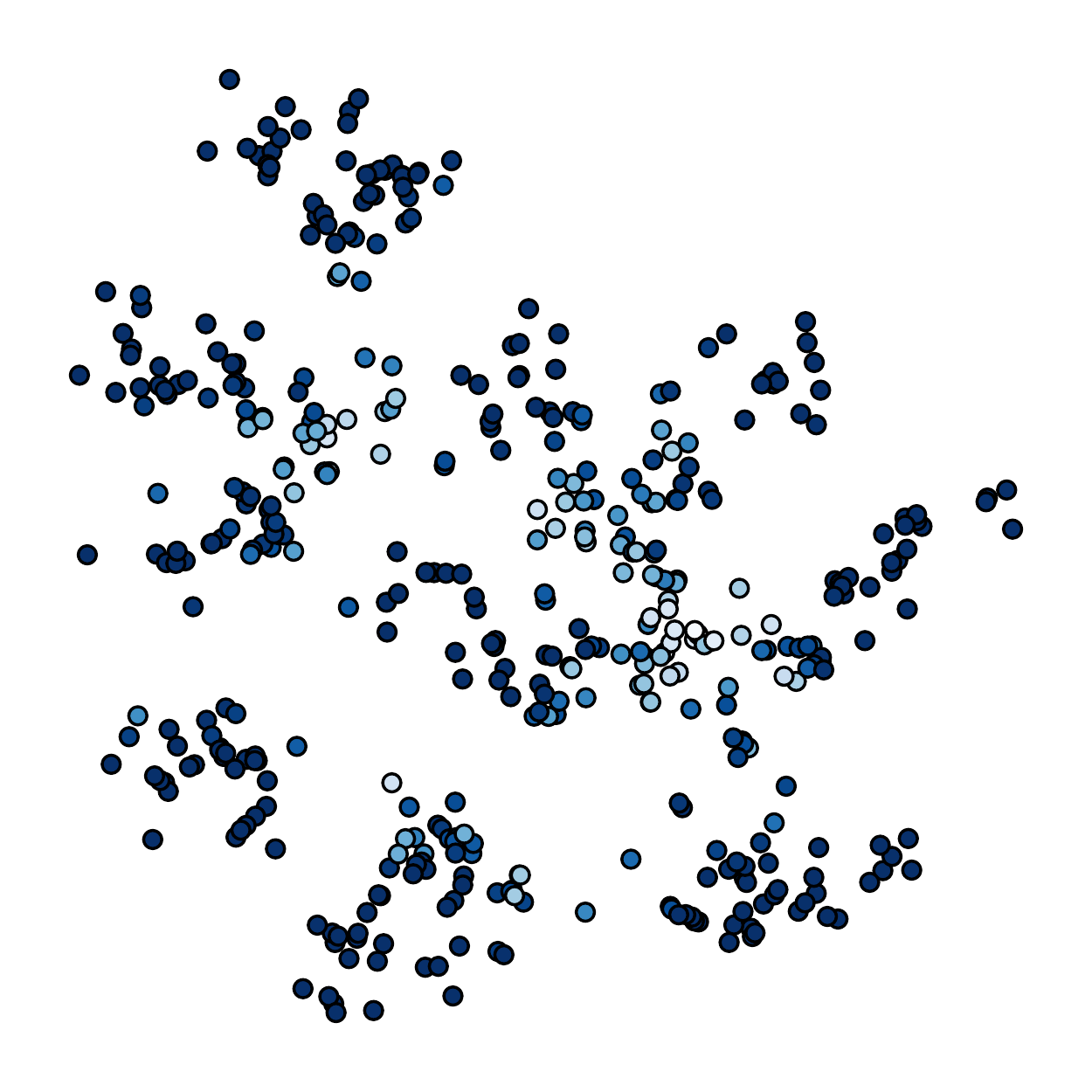}}\hfill  \\ \hline
90000 iter & {\includegraphics[width=0.33\linewidth]{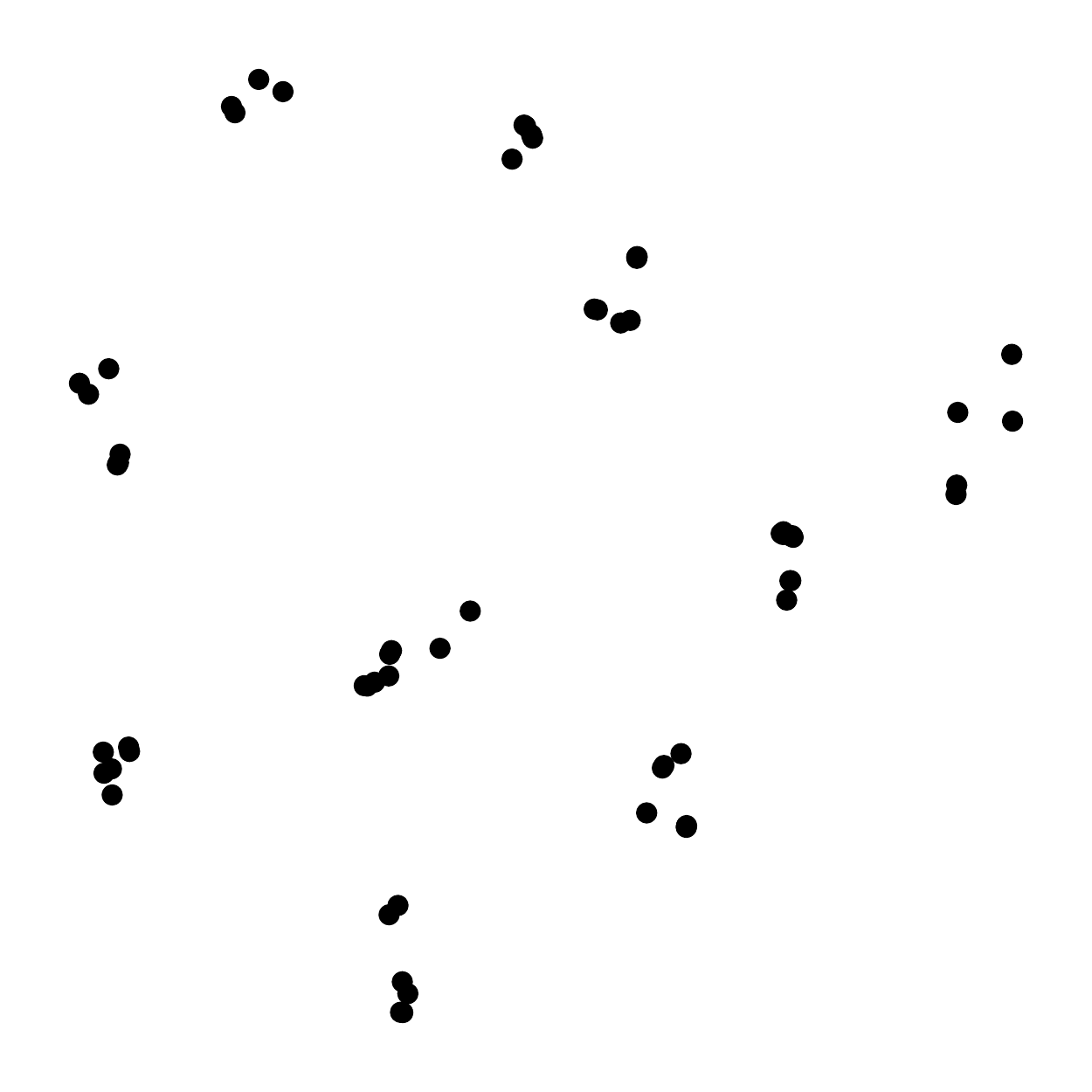}}\hfill &                     {\includegraphics[width=0.33\linewidth]{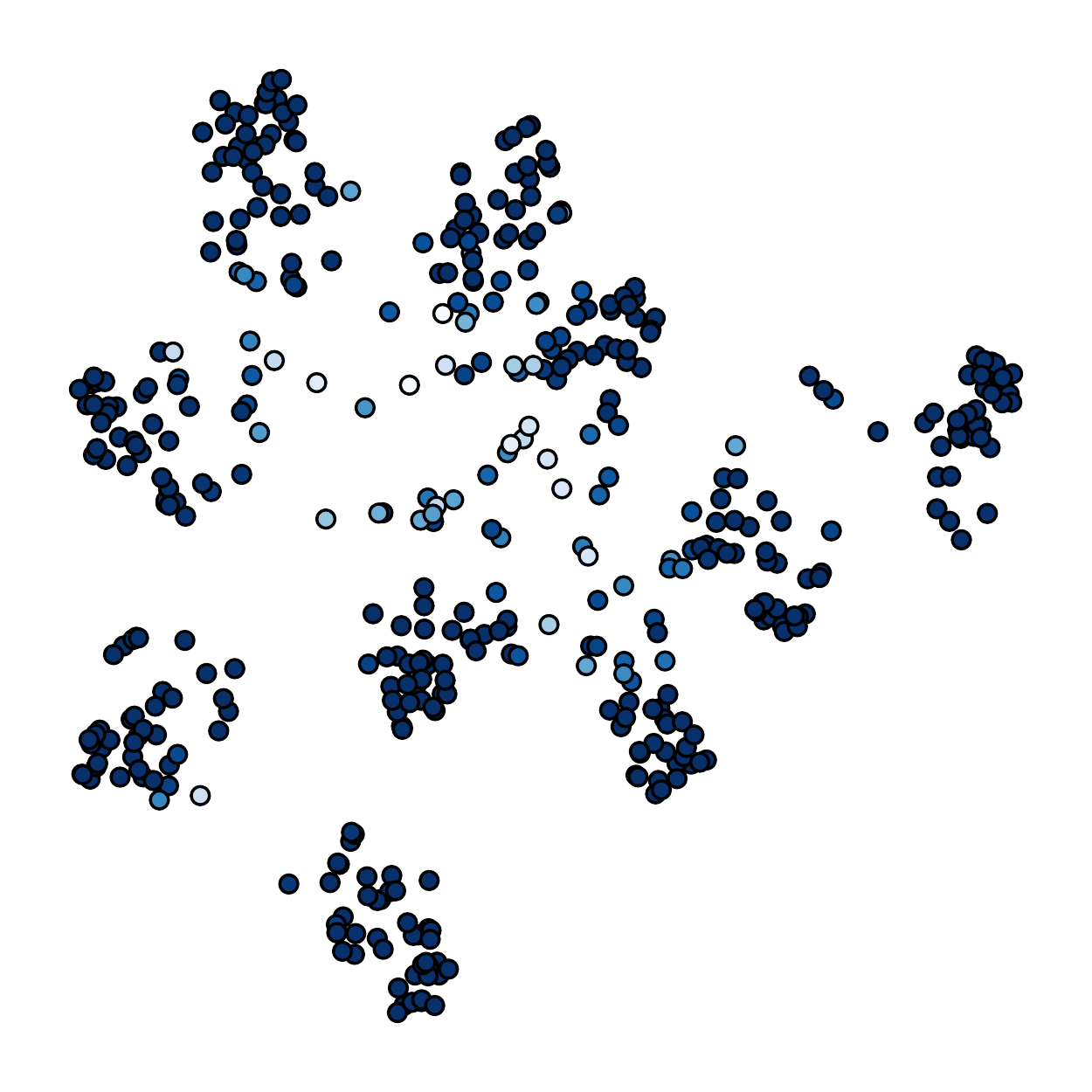}}\hfill & {\includegraphics[width=0.33\linewidth]{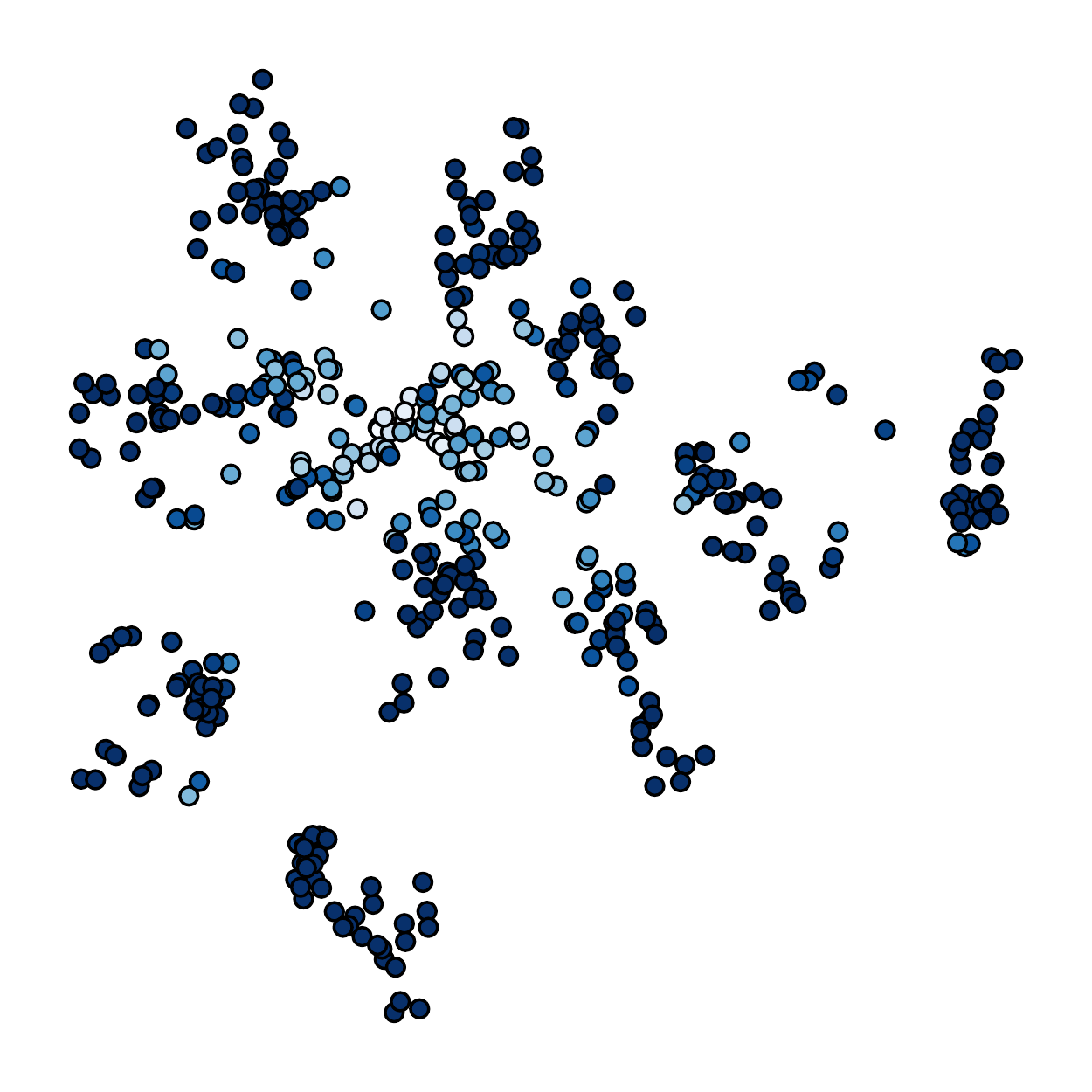}}\hfill  \\ \hline

\end{tabular}
}\vspace{-10pt}
\end{center}
\caption{Visualization on the transitions feature distribution with confidence in \textbf{FlexMatch}. The details are same as above.}
\label{table:tsne-flex}
\end{table*}

% \clearpage\mbox{}Page \thepage\ of the manuscript.
% \clearpage\mbox{}Page \thepage\ of the manuscript.

% This is the last page of the manuscript.
% Now we have reached the maximum size of the ECCV 2022 submission (excluding references).
% References should start immediately after the main text, but can continue on p.15 if needed.

% \clearpage
% % ---- Bibliography ----
% %
% % BibTeX users should specify bibliography style 'splncs04'.
% % References will then be sorted and formatted in the correct style.
% \bibliography{egbib}
% \end{document}

\bibliographystyle{splncs04}
\bibliography{egbib}
\end{document}